\documentclass[journal]{IEEEtran}
\usepackage{amsfonts}
\usepackage{graphicx}
\usepackage{subfigure}
\usepackage{multirow}
\usepackage{amssymb}
\usepackage{amsmath}
\usepackage{diagbox}
\usepackage{bbding}
\usepackage{makecell}
\usepackage{xcolor}
\usepackage{float}  
\usepackage[citecolor=blue, colorlinks]{hyperref}
\usepackage{subfiles}
\usepackage{booktabs}
\usepackage{soul}
\begin{document}

\title{Leveraging Single-View Images for \\ Unsupervised 3D Point Cloud Completion}

\author{Lintai Wu, Qijian Zhang, Junhui Hou,~\emph{Senior Member, IEEE}, and Yong Xu,~\emph{Senior Member, IEEE}
\thanks{This project was supported by the Hong Kong Research Grants Council under Grant 11219422, Grant 11202320, and Grant 11218121. \textit{(Corresponding author: Junhui Hou)}}
\thanks{L. Wu is with the Department of Computer Science, City University of Hong Kong, Hong Kong SAR, and also with the Bio-Computing Research Center, Harbin Institute of Technology, Shenzhen, Shenzhen 518055, Guangdong,
China. Email: lintaiwu2-c@my.cityu.edu.hk;}
\thanks{Q. Zhang and J. Hou are with the Department of Computer Science, City University of Hong Kong, Hong Kong SAR. Email: qijizhang3-c@my.cityu.edu.hk; jh.hou@cityu.edu.hk.}

\thanks{Y. Xu is with the Bio-Computing Research Center,
Harbin Institute of Technology, Shenzhen, Shenzhen 518055, Guangdong,
China, and also with the Shenzhen Key Laboratory of Visual Object Detection and Recognition, Shenzhen, Guangdong
518055, China. Email: yongxu@ymail.com.}}

\markboth{Revised Manuscript submitted to IEEE TMM}%
{Shell \MakeLowercase{\textit{et al.}}: Bare Demo of IEEEtran.cls for IEEE Journals}

\maketitle

\begin{abstract}
    Point clouds captured by scanning devices are often incomplete due to occlusion. To overcome this limitation, point cloud completion methods have been developed to predict the complete shape of an object based on its partial input. These methods can be broadly classified as supervised or unsupervised. However, both categories require a large number of 3D complete point clouds, which may be difficult to capture. In this paper, we propose Cross-PCC, an unsupervised point cloud completion method without requiring any 3D complete point clouds. We only utilize 2D images of the complete objects, which are easier to capture than 3D complete and clean point clouds. Specifically, to take advantage of the complementary information from 2D images, we use a single-view RGB image to extract 2D features and design a fusion module to fuse the 2D and 3D features extracted from the partial point cloud. To guide the shape of predicted point clouds,  we project the predicted points of the object to the 2D plane and use the foreground pixels of its silhouette maps to constrain the position of the projected points. To reduce the outliers of the predicted point clouds, we propose a view calibrator to move the points projected to the background into the foreground by the single-view silhouette image. To the best of our knowledge, our approach is the first point cloud completion method that does not require any 3D supervision. The experimental results of our method are superior to those of the state-of-the-art unsupervised methods by a large margin. Moreover, our method even achieves comparable performance to some supervised methods. 
    We will make the source code publicly available at \url{https://github.com/ltwu6/cross-pcc}.
\end{abstract}


\IEEEpeerreviewmaketitle

\section{Introduction}

\IEEEPARstart{T}{he} recent years have witnessed the thriving development of 3D sensing technology and the rapid popularization of 3D scanning equipment. As one of the most fundamental and common 3D representation modalities, point cloud data have attracted increasing interests from both industry and academia and played a key role in a variety of downstream applications, such as autonomous driving \cite{kato2018autoware,cui2021deep}, shape/scene understanding \cite{liu2020semantic,zhu2022vpfnet,zhang2023unleash,zhang2023pointmcd}, motion analysis \cite{wu2019unsupervised,liu2022geometrymotion}, and SLAM \cite{wang2022d}. Nevertheless, in practical scenarios, directly acquiring high-quality point clouds from the output end of various 3D sensors/scanners is far from straightforward. Due to self-occlusion as well as the existence of environmental obstacles, the captured raw point cloud data typically suffer from sparsity and incompleteness, which may significantly limit the subsequent processing pipelines and hurt downstream task performances.

To computationally address the above issues, driven by the recent advancements of 3D deep set architectures \cite{qi2017pointnet,qi2017pointnet++,li2018pointcnn,thomas2019kpconv,wang2019dynamic,zhang2022reggeonet,zhang2023flattening}, there have emerged numerous deep learning-based point cloud completion frameworks that consume partial point clouds as inputs and reconstruct complete point clouds of the corresponding objects. Generally, existing research on point cloud completion can be mainly classified into \textit{supervised} and \textit{unsupervised} learning paradigms, in terms of \textit{whether ground-truth partial-complete pairs are available and leveraged in the training stage.}

Architecturally, existing supervised point cloud completion frameworks \cite{12,14,15,16,7,6} typically adopt encoder-decoder architectures, which consume as input a partial point cloud and tend to recover its complete geometric structure at the output end. Thanks to the availability of partial-complete pairs, the training process can be directly driven by optimizing shape similarity between the reconstructed point cloud and its corresponding ground-truth complete point cloud. However, the practical applicability of these methods can be restricted due to the requirement of collecting a large amount of high-quality complete point cloud models as ground-truths, which can be demanding and time-consuming. Another representative family of unsupervised methods \cite{1,2,3,4, wu2020multimodal} introduce Generative Adversarial Networks (GANs) \cite{5} to perform point cloud completion without requiring the supervisions provided by the complete shapes paired with the partial inputs during the training stage. To acquire rich prior knowledge of 3D shapes, they first leverage a large complete and clean 3D point cloud dataset from other distributions to pre-train the model, and then they use the learned prior knowledge to guide the completion of the partial inputs with the help of GANs. However, these methods still heavily rely on a large complete and clean 3D point cloud dataset that is difficult to acquire in practice, and the training of GANs is widely known to be unstable.

In addition, most previous methods use only partial point clouds as inputs. However, a single partial point cloud of an object cannot always determine its missing part. For example, given a partial point cloud of a chair with only legs, it is difficult to infer whether this chair has a back. In this case, RGB images can help us predict the complete shape since they provide compact shape information from a specific view angle.

\begin{figure*}[htbp]	
	\centering
	\includegraphics[width=1.0\textwidth]{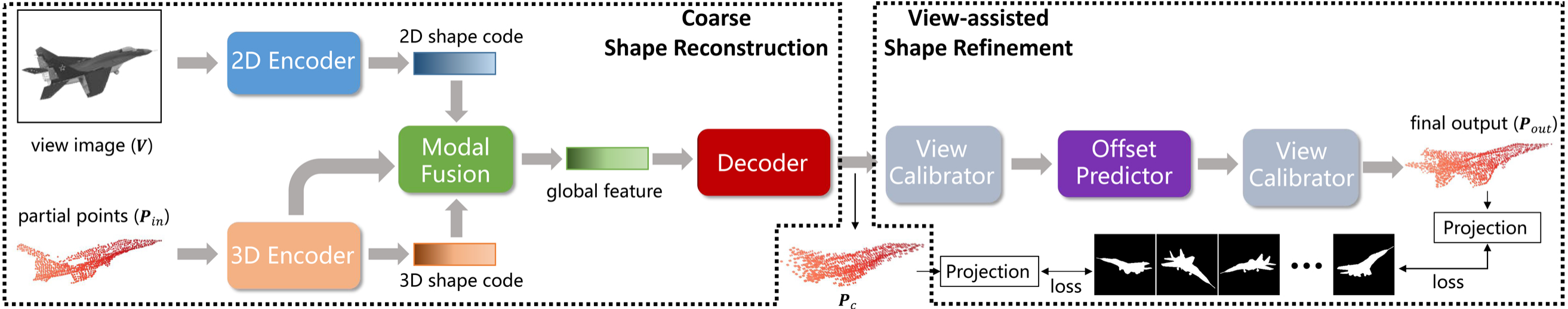}
	\caption{Flowchart of the proposed unsupervised 3D point cloud completion framework named Cross-PCC. Cross-PCC comprises two stages, i.e., Coarse Shape Reconstruction (CSR) and View-assisted Shape Refinement (VSR). The CSR stage consists of a 3D encoder, a 2D encoder, a modality fusion module and a decoder. The VSR stage is composed of two view calibrators and an offset predictor.}
	\label{overall}
\end{figure*}

In this paper, we use both view images and partial point clouds to recover the complete shapes without 3D complete point clouds as supervision. Our model consists of two stages, i.e., coarse shape reconstruction (CSR) and view-assisted shape refinement (VSR), as illustrated in Fig. \ref{overall}. Specifically, the CSR stage aims to generate a coarse yet complete point cloud based on the cross-modal features. In this stage, we use a 3D encoder to extract 3D features from a partial point cloud, and a 2D encoder to extract 2D features from a single-view image. The 3D and 2D features are then fused by a fusion module to get the global feature. In this way, the global feature takes full advantage of the information provided by both the partial point cloud and the view image. Our decoder aims to transform the global feature into a complete point cloud. Inspired by SnowflakeNet \cite{18}, we achieve this by first predicting a sparse point cloud and then splitting the points into multiple child points.
The point cloud produced by the CSR stage is coarse and inaccurate. It has many outliers and some details, such as small holes and parts, are lost. To address these issues, we refine the generated point cloud in the second stage.
In the VSR stage, in order to eliminate outliers, we propose a view calibrator (VC) to move the points projected within the background into the foreground of the silhouette map binarized from the input view image. This calibrator eliminates outliers and recovers some details from a single view, but outliers from other views still exist. To further refine the results, we adopt DGCNN \cite{19} to predict the offset of each point and use the VC to calibrate the result again.

Besides, different from previous supervised and unsupervised methods, both requiring complete and clean point clouds as supervision\footnote{The previous unsupervised methods also require a large number of complete point clouds as supervision to pre-train models for capturing prior shape information.}, in this paper, we only use 2D images as supervision. Specifically, 
inspired by the unsupervised single image-based 3D reconstruction method 2DPM \cite{20},  in the data pre-processing stage, we uniformly sample as ground truth 2D points in the foreground pixels from different views of \textit{silhouette images}. In the training stage, we project the predicted point cloud to the 2D plane and compute the Chamfer Distance (CD) between the projected points and sampled points. Extensive experiments demonstrate that our method outperforms the state-of-the-art unsupervised methods by a large margin. Moreover, our method even yields comparable performance to some supervised methods.

Conclusively, the core contributions of this paper are three-fold:

\begin{itemize}

	\item We propose \textit{the first} unsupervised point cloud completion method completely without requiring additional 3D supervision. Compared to the previous unsupervised methods, our method requires only 2D images, which are easier and inexpensive to capture.
	
	\item We propose to use image features to enhance the features of 3D point clouds. We design a feature extraction and a fusion module to fuse these features of two modalities. Our model takes full advantage of both 2D images and 3D point clouds, which largely boosts the performance.
	
	\item We propose a view-assisted refinement stage to move outliers to more correct positions and refine them with 2D images. The view calibrator in the refinement stage can effectively reduce outliers and recover details.

\end{itemize}

The remainder of this paper is organized as follows. Section \ref{related_work} reviews the previous point cloud completion, reconstruction and single image-based 3D reconstruction approaches. Section \ref{proposed_method} introduces our methods in detail. Section \ref{experiments} introduces our experimental settings, results and analyses. Section \ref{conclusion} summarizes our work.

\section{Related Work}
\label{related_work}

\subsection{3D Point Cloud Completion and Reconstruction}
\subsubsection{Supervised methods}

The supervised methods can be generally divided into two categories according to the intermediate data representations. 
One type is the voxel-based methods \cite{7,8,9,10}. These methods first voxelize point clouds as occupancy grids and then employ 3D convolutional neural networks (3D 
CNNs) architecture to process them. Dai et al. \cite{8} used a 3D-Encoder Predictor Network to predict a coarse shape and then designed a 3D shape synthesis method to refine the coarse output with the shape prior from a complete shape database. Xie et al. \cite{7} proposed a novel Gridding and Gridding Reverse module for converting between point clouds and voxels to reduce structural information loss. Besides, they designed a Cubic Feature Sampling module to extract context features of neighbor points. 
Although the voxel-based methods could take advantage of the strong ability of CNNs, the 3D CNNs require large computational costs. Also, the voxelization operation may lose the accuracy of point coordinates. 

Another type is point cloud-based methods. These methods directly use point clouds as inputs and intermediate representations and employ Multi-Layer Perceptron (MLP) to extract features. There are two mainstream methods to generate missing parts of point clouds. 

The first one assumes that the 3D surfaces of point clouds can be computed from the 2D manifold by consecutive mapping(folding-based methods). For example, FoldingNet \cite{11} employed an encoder to extract the global feature of the partial point cloud and then used a 2D grid together with the global feature to reconstruct the complete point cloud by a decoder. However, the FoldingNet cannot fold a 2D grid into a complex shape. Yuan et al. \cite{12} extended this method by first using fully-connected (FC) layers to predict a coarse point cloud and then employing the Folding operation for each coarse point. This method takes full advantage of 
both the flexibility of FC layers and the surface-smoothness of Folding operation.
Groueix et al. \cite{13} used multiple MLPs to deform 2D grids into different local
patches. Although this method can generate more complex shapes than FoldingNet, there are some overlapped regions between patches. In order to solve this problem, Liu et al. \cite{14} proposed an expansion penalty to reduce the overlap areas. Tang et al. \cite{21} used multiple grids to generate a point cloud with shared MLP. They designed a Stitching loss to preserve the uniformity of points.
Wen et al. \cite{16} proposed a skip-attention module to capture the geometric information of local regions and used the Folding operation to generate the complete point cloud hierarchically.
Rather than utilizing the fixed 2D grid as seeds, Yang et al. \cite{23} used a network to learn a high-dimension seed to generate point clouds. Similarly, Pang et al. \cite{24} argued that the fixed 2D grid cannot capture the complex topologies of point clouds. They used a network to learn the point cloud topology and tore the grid into patches to fit the object shape.
Yu et al. \cite{15} considered the point cloud completion as a translation task and 
employed the Transformer to infer the missing parts based on the given parts. 
They first generated coarse missing points and then used the Folding operation to 
generate their neighbor points. 

The main drawback of the folding-based methods is that they predefine a topology for the target object, which constrains the generated complete shape. Therefore, some approaches directly generated points in 3D space(point-based methods) instead of by 2D-manifold. Tchapmi et al. \cite{17} designed a rooted-tree decoder to generate points hierarchically as the tree grows. Inspired by the generation process of snowflakes, Xiang et al. \cite{18} proposed to split each parent point into several child points in a coarse-to-fine manner. Besides, they proposed a skip-transformer block to capture the relation between the features of child points and parent points. 

Recently, the multimodal fusion of point clouds and images has been proven effective in many tasks. For example, in the 3D object detection task, the performance of Lidar and image fusion-based methods usually surpass that of single modal methods \cite{25,26,27}. Some researchers have begun to make use of 2D images to help the completion of point clouds.

Intuitively, the RGB image can provide information 
about occupancy and emptiness. To utilize the emptiness information, Gong et 
al. \cite{28} used MLPs to encode both partial input and emptiness ray generated from 
depth images and then refine the predicted complete point cloud under the 
guidance of the emptiness. This method can distinguish the boundary of the point 
cloud well. Zhang et al. \cite{6} adopted a single view image to predict a coarsely complete point cloud and proposed a Dynamic Offset Predictor to refine the coarse output. However, directly predicting a complete point cloud from an image is difficult and inaccurate. 
Zhu et al. \cite{zhu2023csdn} proposed a  cross-modal shape-transfer dual-refinement network to leverage the image features for point cloud completion. To obtain a coarse prediction aided by images, they introduced a module named IPAdaIN transferring global shape information from images to the point cloud domain. Furthermore, they designed a dual-refinement module consisting of local refinement and global constraint units to refine the coarse output.
Nevertheless, \cite{6} and \cite{zhu2023csdn} are supervised learning frameworks, and it is non-trivial to adapt them to our unsupervised setting that lacks complete point clouds as ground truth. To better fuse the 3D and 2D modalities, Aiello et al. \cite{aiello2022cross} proposed XMFnet, a model that uses cross-attention and self-attention modules to merge the point and pixel features. Additionally, they proposed a strategy to supervise the training with 2D images. However, our method differs significantly from XMFnet in terms of the model architecture and supervision approach. Firstly, XMFnet focuses on utilizing attention mechanisms to fuse 3D and 2D features, while we adopt a cascaded strategy to merge global and local features from point clouds and images. Additionally, we place significant emphasis on the calibration of reconstructed results. Secondly, XMFnet employs differentiable rendering based on alpha compositing for supervision, whereas we utilize a projection-based approach that enables faster transformation of results onto a 2D plane compared to XMFnet's method.

\begin{figure*}[ht]	
	\centering
	\includegraphics[width=1.0\textwidth]{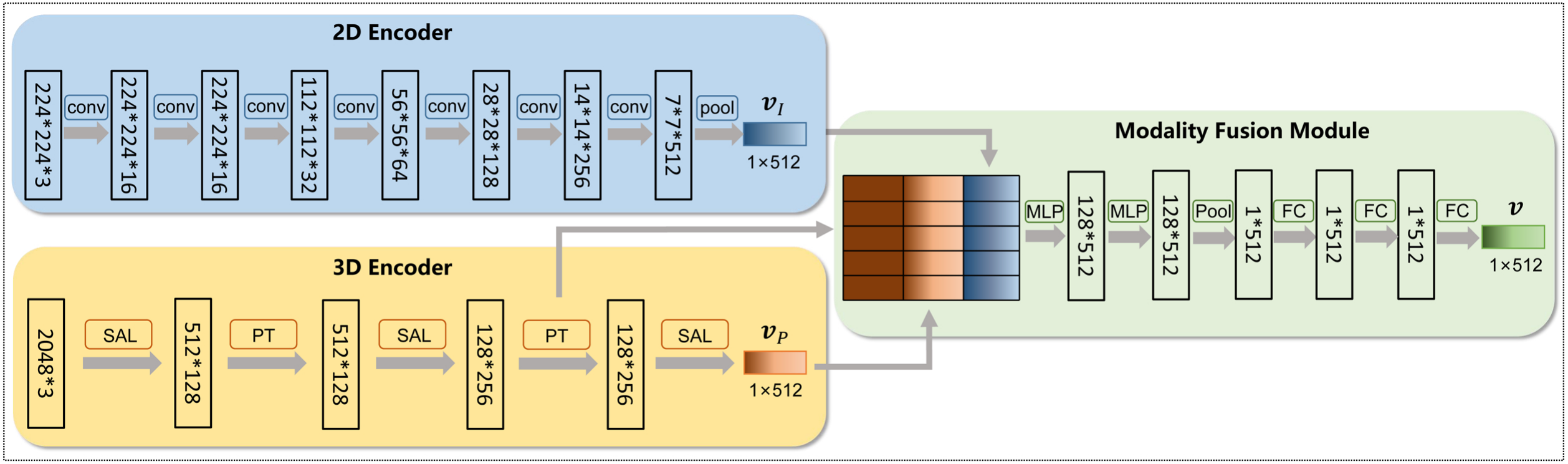}
	\caption{Architecture of 2D encoder, 3D encoder and modality fusion module. ``SAL" and ``PT" denote Set Abstraction Layer in PointNet++ \cite{qi2017pointnet++} and Point Transformer \cite{18, zhao2021point,pan20213d,guo2021pct}. }
	\label{encoder}
\end{figure*}

\subsubsection{Unsupervised methods}
The existing unsupervised point cloud completion methods aim to estimate the complete point cloud of a partial point cloud in an unpaired manner. They do not require the supervision provided by the complete shapes paired with the partial inputs during the training stage. Instead, they require a large dataset of complete and clean point clouds to train a model to capture the prior 3D shape information. Then they leveraged the pre-trained model to guide the completion of the partial point cloud. 

Chen et al. \cite{1} proposed the first unsupervised point cloud method. They first pre-trained two Auto-encoders by reconstructing partial inputs and complete point clouds, respectively. The partial inputs and complete point clouds are from different data distributions. Then they used a generator to transform the global feature of the partial input and a discriminator to differentiate between the transformed global feature and that of the complete point clouds. By making the global feature of partial input close to the complete one, the completion results are more reasonable. Wen et al. \cite{2} extended this method by simultaneously learning the transformation from partial inputs to complete point clouds and its inverse transformation to enhance the 3D shape understanding ability of the model. Zhang et al. \cite{3} introduced the GAN Inversion to the point cloud completion task. They first pre-trained a GAN with a complete and clean dataset. Then they searched for a global feature(latent code) that was best matched with the partial input in the latent space of GAN and optimized it. Cai et al. \cite{4} proposed to learn a unified latent space for partial and complete point clouds by modeling the incompleteness as the occlusion degree of the global feature of complete point clouds. The model is capable of well understanding the complete shape by being trained on the unified latent space.
The advantage of these methods is the strong generation ability. Since the pre-trained Auto-encoders and latent codes contain information of many 3D shapes, the model can generate many reasonable shapes when inputting a new partial point cloud from different data distributions.
However, they require a large complete 3D point cloud dataset. Although these data are unpaired with the input partial point cloud, capturing 3D data is expensive and inconvenient. Besides, the training of GAN is difficult and time-consuming. 

\subsection{Single Image-based 3D Reconstruction}
Single image-based 3D reconstruction targets to predict the 3D shape from the single-view image\cite{29,30,31,fan2017point,mescheder2019occupancy,wen20223d}. Here, we briefly review the unsupervised methods for point cloud representation since they are more related to our work.

Without 3D supervision, the unsupervised methods usually use differentiable renderers to render the generated point cloud into a silhouette or RGB image and then compare the similarity of the rendered and ground truth image \cite{29,30,31,32,33,34,37,38}. They first project the 3D points to the image plane under the given camera parameters and then compute the influence of each point on the pixel's color of the target image. The color of each pixel is determined by the influences and depth values of all projected points.

The rendering process consumes a lot of computational time. To reduce the computation cost while keeping the performance, some researchers only project 3D points to the image plane, and then utilize silhouette images to constrain the positions of the projected points. We refer to these methods as projection-based methods. For example, Han et al. \cite{35} designed a smooth silhouette loss to pull the projected points inside the foreground of the silhouette image, and a repulsion loss to make the projected points cover the entire foreground. To recover detailed structures of 3D shapes, Chen et al. \cite{20} proposed to uniformly sample a fixed number of 2D points within the silhouettes regardless of the image resolution, and then computed the Chamfer Distance between the projected and sampled points to supervise the generated shapes. Extensive experimental results demonstrate that the projection-based methods are effective and require less computational cost than the rendering-based methods. In this paper, we utilize the supervision method of \cite{20} to guide the training of our framework. However, our framework and data differ significantly from \cite{20}. That is, \cite{20} is specifically designed for the task of single-view image reconstruction, which leverages the 2D shape information of objects in RGB images to perceive their corresponding 3D shapes. In contrast, ours focuses on point cloud completion, aiming to infer the complete shape based on the existing 3D information of the partial point cloud.

\section{Proposed Method}
\label{proposed_method}

Denote by \(\mathbf{P}_{in}\in\mathbb{R}^{N\times3}\) the partial point cloud of an object and \(\mathbf{V}\in\mathbb{R}^{W\times H\times3}\) an RGB image captured from a random view angle, where \(N\) is the number of points, and $W$ and $H$ are the width and height of the image, respectively. Our goal is to find a mapping \(f(\mathbf{P}_{in}, \mathbf{V})=\mathbf{P}_{out}\), where \(\mathbf{P}_{out}\) denotes the predicted complete point cloud of $\mathbf{P}_{in}$.

As depicted in Figure \ref{overall}, our approach, namely Cross-PCC, consists of two stages, i.e., coarse shape reconstruction (CSR) and view-assisted shape refinement (VSR). In the CSR stage, we predict a coarse complete point cloud based on the partial point cloud and a view image. In the VSR stage, we propose a view calibrator and an offset predictor to refine the coarse point cloud. In what follows, we will detail our method.

\subsection{Coarse Shape Reconstruction}

The CSR stage consists of a 3D encoder, a 2D encoder, a modality fusion module, and a decoder, as depicted in Figure \ref{encoder}. The 3D encoder is responsible for extracting a 3D shape code that represents the global shape of the partial point cloud. It incorporates three set abstraction layers \cite{qi2017pointnet++} and two point Transformer layers \cite{18, zhao2021point,pan20213d,guo2021pct}, which aim to capture the correlation between point features. On the other hand, the 2D encoder extracts a 2D shape code that represents the global shape from the view image. It comprises eight convolution layers and a global pooling layer.
To fuse the features from both modalities and capture both local and global shape information, we introduce a modality fusion module, which first concatenates the intermediate 3D local feature, and 2D and 3D shape codes, and then employ MLPs to encode them into a global feature. To further enhance this global feature, we utilize three FC layers for embedding.

\begin{figure}[t]	
	\centering
	\includegraphics[width=0.48\textwidth]{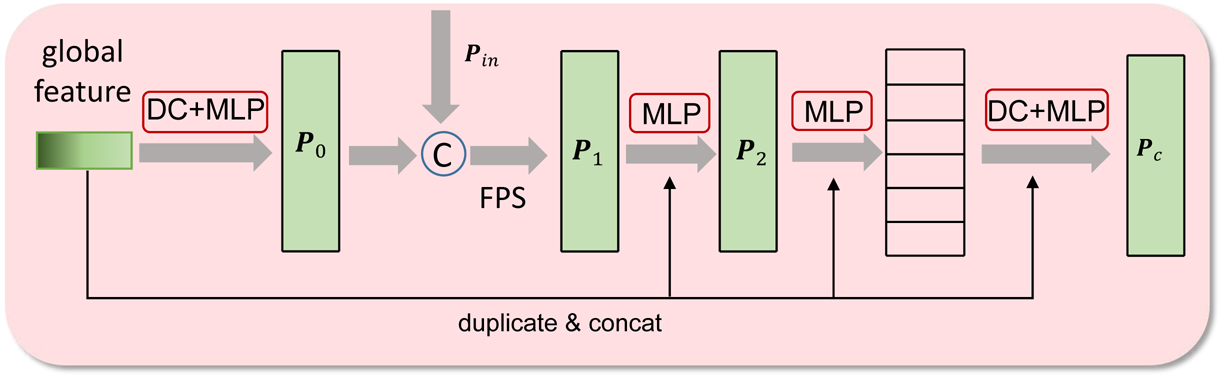}
	\caption{Architecture of the decoder. ``DC" denotes the ``Deconvolution" operation. ``C" in the circle means concatenation operation.}
	\label{decoder}
\end{figure}

\begin{figure*}[htbp]	
	\centering
	\includegraphics[scale=0.25]{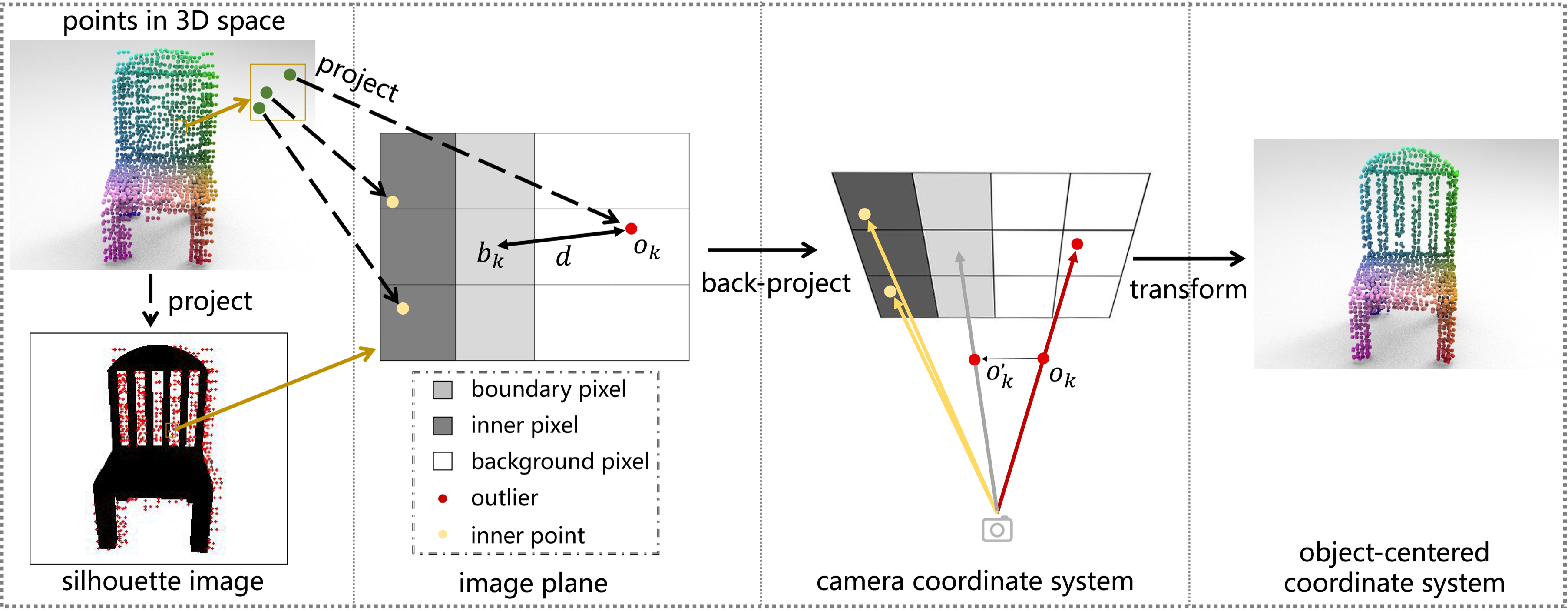}
	\caption{Illustration of the calibration operation. We project the predicted points in 3D space into the image plane and then replace the \(x\) and \(y\) coordinates of the outlier with those of its nearest boundary pixel. In the camera coordinate system in which all the points in the image plane are rays starting from the camera, this procedure means moving the outlier to the ray of its nearest boundary pixel.}
	\label{calibrator}
\end{figure*}

The decoder plays a crucial role in predicting the complete shape while preserving the partial inputs. Motivated by \cite{18}, our approach involves a coarse-to-fine process: first, we predict a sparse yet complete point cloud, and then we generate detailed parts by splitting each point into several child points using MLPs. The architecture of the decoder is illustrated in Figure \ref{decoder}.
Specifically, we employ a deconvolution layer followed by four layers of MLPs to transform the global feature \(\mathbf{v}\) into the sparse point cloud \(\mathbf{P}_{0}\in\mathbb{R}^{N_0\times 3} \). In order to better preserve the partial input, we concatenate the partial input \(\mathbf{P}_{in}\) with \(\mathbf{P}_{0}\) and adopt the Farthest Point Sampling method to sample \(N_1\) points, resulting in \(\mathbf{P}_{1}\in\mathbb{R}^{N_1\times 3}\).   
Considering that \(\mathbf{P}_{1}\) may exhibit inconsistencies in terms of density and shape, we feed it 
along with the global feature into MLPs to obtain a refined point cloud \(\mathbf{P}_{2}\in\mathbb{R}^{N_1\times 3}\).
Next, we embed \(\mathbf{P}_{2}\) into point features, and using a deconvolution layer, we split each point feature into \(r\) child features. This split operation aims to infer \(r\) neighbors' positions for each point in \(\mathbf{P}_{2}\), thereby obtaining a finer point cloud. Finally, we use MLPs to decode these child features to obtain the coarse output \(\mathbf{P}_{c}\in\mathbb{R}^{N\times 3}\).

\subsection{View-assisted Shape Refinement}

The coarse point cloud predicted by the CSR stage often includes points lying outside the boundaries of the object. As Figure \ref{calibrator} shows, these points lie in the background of the silhouette images when projected into the image plane (we refer to these out-of-boundary points as outliers and others as inner points). To address this issue, we propose a view calibrator (VC) to calibrate these outliers with the help of single-view silhouette image, as illustrated in Figure \ref{calibrator}. Specifically, given a point cloud \(\mathbf{P}\in\mathbb{R}^{N\times3}\) and an RGB view image from a random view angle, we first binarize the input view image to obtain a silhouette image \(\mathbf{S}\). And then we find the coordinates of all the boundary pixels \(\mathbf{B}\in\mathbb{R}^{L\times2}\) of the silhouette image by \cite{suzuki1985topological}, where \(L\) is the number of boundary pixels. Next, we project \(\mathbf{P}\) into the image plane by employing the intrinsic and extrinsic camera parameters \(\mathbf{T}\in\mathbb{R}^{4\times 4}\). Let the projected point set be \(\mathbf{P}'=(\mathbf{x'}, \mathbf{y'}, \mathbf{z'})\), and the projection can be formulated as
\begin{equation}
	(\mathbf{x'}, \mathbf{y'}, \mathbf{z'}, \mathbf{1})^\texttt{T} = \mathbf{T}*(\mathbf{x}, \mathbf{y}, \mathbf{z}, \mathbf{1})^\texttt{T}.
\end{equation}

Then we determine outliers and inner points according to the silhouette image. Let \(\mathbf{O} \in \mathbb{R}^{K\times3}\) be the outlier set in \(\mathbf{P}'\), i.e., \(\mathbf{O} \subset \mathbf{P}'\), where \(K\) is the number of outliers. Since the pixel value for background is 0 and foreground is 1 in the silhouette images, the \(j\)-th project point \(\mathbf{p}'_{j}=(x'_{j},y'_{j},z'_{j}) \in \mathbf{O}\) if \(\mathbf{S}(\lfloor x'_{j},y'_{j}\rfloor)=0\).

In the next step, for each outlier \(\mathbf{o}_{k}=(x_{k}, y_{k}, z_{k}) \in \mathbf{O}\), we compute the Euclidean distance between it and all boundary pixels in the image plane without \(z\)-axis and select the nearest boundary pixel \(\mathbf{b}_{k}=(x_{b_{k}}, y_{b_{k}})\), i.e., 
\begin{equation}
	\mathbf{b}_{k} = \texttt{argmin}\, d(\mathbf{o}_{k},\mathbf{B}),
\end{equation} 
where \(\mathbf{b}_{k} \in \mathbf{B}\), and $d(\cdot, \cdot)$ means the Euclidean distance of two points. Then we replace the \(x\)-coordinate and \(y\)-coordinate of \(\mathbf{o}_{k}\) with the counterparts of \(\mathbf{b}_{k}\). The new coordinate of \(\mathbf{o}_{k}\) is \((x_{b_{k}}, y_{b_{k}}, z_{k})\). That means we move the outliers inside the foreground in the silhouette image while keeping the \(z\)-coordinate unchanged.
Finally, we back-project the new point set into the objected-center coordinate by:
\begin{equation}
(\mathbf{x}_{cal}, \mathbf{y}_{cal}, \mathbf{z}_{cal}, \mathbf{1})^\texttt{T} = \mathbf{T}^{-1}*(\mathbf{x}'_{cal}, \mathbf{y}'_{cal}, \mathbf{z}'_{cal}, \mathbf{1})^\texttt{T},
\end{equation} 
where \((\mathbf{x}_{cal}, \mathbf{y}_{cal}, \mathbf{z}_{cal})=\mathbf{P}_{cal}\) represents the calibrated 3D point set and \((\mathbf{x}'_{cal}, \mathbf{y}'_{cal}, \mathbf{z}'_{cal})\) denotes the calibrated points in image plane.

We leverage VC to eliminate outliers and recover some details of \(\mathbf {P}_{c}\) from a single view. However, outliers from other views still exist and the calibration may affect the density of the points. To further adjust the positions of the calibrated points, we use DGCNN \cite{19} as the offset predictor to infer the offset for each point of the calibrated shape. 
Specifically, the offset predictor is composed of five layers of Edge Convolutions \cite{19}. It takes the calibrated point cloud as input and generates offset values for all points. The result \(\mathbf {P}_{op}\) is obtained by adding these offsets to the calibrated point cloud.
Due to the possibility of incorrect offset predictions by DGCNN, there may be outliers in \(\mathbf {P}_{op}\). To address this issue, we utilize the VC to calibrate \(\mathbf {P}_{op}\) again and obtain the final output \(\mathbf {P}_{out}\). Compared to \(\mathbf {P}_{op}\), \(\mathbf {P}_{out}\) has fewer outliers.

\subsection{Loss Function}
Considering both the effectiveness and computational cost, we employ the projection-based method proposed by Chen et al. \cite{20} to train our Cross-PCC. We assume that each object has $I$ RGB view images captured from different view angles. In the data-preprocessing stage,
we binarize these RGB images to obtain silhouette images.
And then we uniformly sample \(M\) 2D points \(\mathbf{G}^{i} \in\mathbb{R}^{M\times2} \) in the foreground pixels of each silhouette image, where \(i=1,2,...,I\). During the training stage, 
we project the predicted point cloud \(\mathbf {P}_{c}\) into the $i$-th image plane 
and preserve only the \(x\)-coordinate and \(y\)-coordinate, resulting in a 2D point set \(\mathbf{Q}^{i}_{c} \in\mathbb{R}^{N\times2}\). 
Then we compute the 2D Chamfer Distance (CD) between the sampled points \(\mathbf{G}^{i}\) and \(\mathbf{Q}^{i}_{c}\):
\begin{equation}
\begin{split}
	{\rm CD}(\mathbf{G}^{i},\mathbf{Q}^{i}_{c})=\frac{1}{|\mathbf{G}^{i}|}\sum_{\mathbf{g}^{i}\in \mathbf{G}^{i}} \mathop{\texttt{min}}\limits_{\mathbf{q}^{i}\in \mathbf{Q}^{i}_{c}}||\mathbf{g}^{i}-\mathbf{q}^{i}|| \\ 
	+ \frac{1}{|\mathbf{Q}^{i}_{c}|}\sum_{\mathbf{q}^{i}\in \mathbf{Q}^{i}_{c}} \mathop{\texttt{min}}\limits_{\mathbf{g}^{i}\in \mathbf{G}^{i}}||\mathbf{q}^{i}-\mathbf{g}^{i}||.
\end{split}
\end{equation}

For each object, we compute the CD loss for all views and aggregate them. Our projection loss is defined as
\begin{equation}
\begin{split}
	L_{proj}(\mathbf{G},\mathbf{Q}_{c})=\frac{1}{I}\sum_{i=1}^I {\rm CD}(\mathbf{G}^{i},\mathbf{Q}^{i}_{c}).
\end{split}
\end{equation}

In addition, to preserve the partial input, we also employ the partial matching loss \cite{2}, which computes the unidirectional CD from output to input: 
\begin{equation}
\begin{split}
	L_{part}(\mathbf{P}_{in}, \mathbf{P}_{c})=\frac{1}{|\mathbf{P}_{in}|}\sum_{\mathbf{p}_{in}\in \mathbf{P}_{in}} \mathop{\texttt{min}}\limits_{\mathbf{p}_{c}\in \mathbf{P}_{c}}||\mathbf{p}_{in}-\mathbf{p}_{c}||.
\end{split}
\end{equation}

In the CSR stage, in addition to utilizing the coarse output \(\mathbf {P}_{c}\), we also incorporate intermediate results $\mathbf{P}_0$, and $\mathbf{P}_2$ to calculate the loss. In the VSR stage, 
since the VC module is non-differentiable, computing loss using \(\mathbf {P}_{out}\) prevents gradient backpropagation. Moreover, the VC module does not have any trainable parameters. Therefore, we use the output of the offset predictor \(\mathbf {P}_{op}\) instead of \(\mathbf {P}_{out}\) to compute the loss.
The total loss functions of the CSR stage and VSR stage are 
\begin{equation}
\begin{split}
	L_{CSR} = L_{proj}(\mathbf{G},\mathbf{Q}_{0})+L_{part}(\mathbf{P}_{in}, \mathbf{P}_0) \\ +L_{proj}(\mathbf{G},\mathbf{Q}_{2})+L_{part}(\mathbf{P}_{in}, \mathbf{P}_2) \\ +L_{proj}(\mathbf{G},\mathbf{Q}_{c})+L_{part}(\mathbf{P}_{in}, \mathbf{P}_{c}),
\end{split}
\end{equation}

\begin{equation}
\begin{split}
	L_{VSR} = L_{proj}(\mathbf{G},\mathbf{Q}_{op})+L_{part}(\mathbf{P}_{in}, \mathbf{P}_{op}),
\end{split}
\end{equation}
where $\mathbf{Q}_{0}$, $\mathbf{Q}_{2}$, $\mathbf{Q}_{c}$ and $\mathbf{Q}_{op}$ are the 2D points projected by $\mathbf{P}_0$, $\mathbf{P}_2$, $\mathbf{P}_{c}$ and $\mathbf{P}_{op}$.

\begin{figure}[htbp]
	\centering  
	\subfigbottomskip=1pt 
	\subfigure{
		\includegraphics[width=0.3\linewidth]{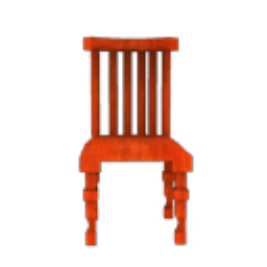}}
	\subfigure{
		\includegraphics[width=0.3\linewidth]{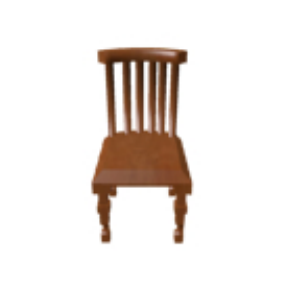}}
	\subfigure{
		\includegraphics[width=0.3\linewidth]{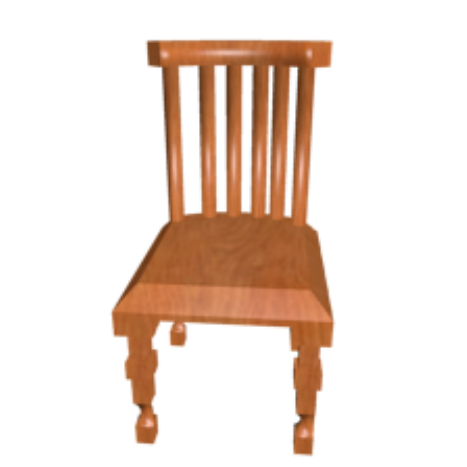}}
	\\
	\setcounter{subfigure}{0}
	\subfigure[2DPM \cite{20}]{
		\includegraphics[width=0.3\linewidth]{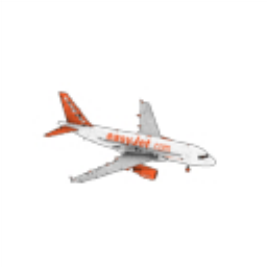}}
	\subfigure[ViPC \cite{6}]{
		\includegraphics[width=0.3\linewidth]{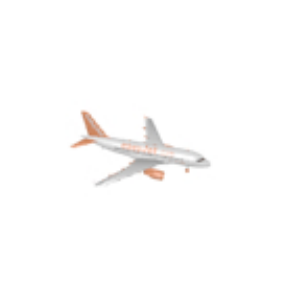}}
	\subfigure[Ours]{
		\includegraphics[width=0.3\linewidth]{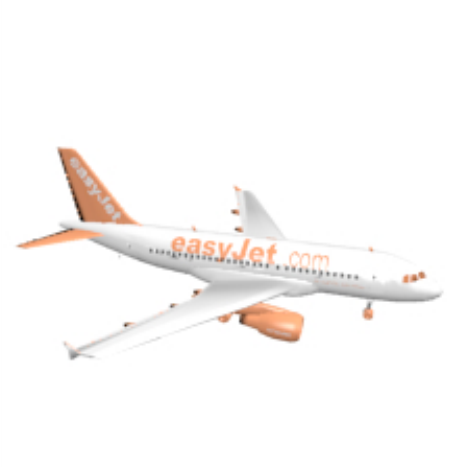}}
	\caption{Comparison between our rendered images and the existing rendered images. Note that all the images are shown at the same size. The images provided by 2DPM \cite{20} are too unreal, and the rendered objects in the images of ViPC \cite{6} are too small. In contrast, our renderer images are more real and the objects are larger so as to keep more details and reduce the background pixels.}
	\label{render}
\end{figure}

\section{Experiments} \label{experiments}

\begin{table*}[htbp]
	\centering
    \renewcommand\arraystretch{1.25}
	\caption{Results on 3D-EPN benchmark in terms of L2 CD\(\downarrow\) (scaled by \(10^{4}\)).  Note that the reported values to the left and right of the slash ``/" are the ${\rm CD}_{min}$ and ${\rm CD}_{avg}$, respectively. The \textbf{bold} numbers represent the best results of the unsupervised methods.}
	\label{epn_results}
	\begin{tabular}{l|c|c|c|c|c|c|c|c|c|c}
		\toprule[1.2pt]
		Methods             & Supervised & Average & Plane & Cabinet & Car & Chair & Lamp	& Couch	& Table	& Watercraft \\ \hline  
		3D-EPN \cite{8}& yes    & 29.1 & 60.0 & 27.0 & 24.0    & 16.0    & 38.0	& 45.0	& 14.0	& 9.0    \\
		FoldingNet \cite{11}& yes    & 8.2 & 2.7 & 8.7 & 5.0    & 9.8    & 14.5	& 8.7	& 9.7	& 7.1    \\
		PCN \cite{12}      & yes & 7.6 & 2.0 & 8.0 & 5.0    & 9.0    & 13.0	& 8.0	& 10.0	& 6.0 \\
		TopNet \cite{17}      & yes & 7.8 & 2.5 & 8.5 & 4.6    & 9.1    & 13.9	& 8.8	& 9.1	& 6.5    \\ 
		SnowflakeNet \cite{18}      & yes & 5.0 & 1.4 & 6.3 & 4.0    & 6.1    & 6.2	& 6.3	& 6.5	& 3.4    \\ \hline
		PCL2PCL \cite{1}      & no & 17.4 & 4.0 & 19.0 & 10.0    & 20.0    & 23.0	& 26.0	& 26.0	& 11.0    \\
		Cycle \cite{2}   & no & 13 & 3.58 & \textbf{12.0} & \textbf{7.4} & 14.6    & 23.5	& 14.3	& 20.9	& 8.1    \\
		Inversion \cite{3}    & no & 18.3 & 5.1 & 21.9    & 12.2 & 21.6 & 24.2	& 20.0	& 31.7	& 9.8 \\ \hline
		Cross-PCC (ours)          & no	& \textbf{8.4/9.1} & \textbf{2.3/2.4} & 13.0/14.0 & \textbf{7.4}/7.6 & \textbf{9.8/10.6} & \textbf{7.8/9.3} & \textbf{11.5/12.1} & \textbf{9.6/10.7}	& \textbf{6.1/6.8}	 \\ \bottomrule[1.2pt]
	\end{tabular}
\end{table*}

\begin{table*}[htbp]
	\centering
	\renewcommand\arraystretch{1.25}
	\caption{Results on PCN benchmark in terms of L2 CD\(\downarrow\) (scaled by \(10^{4}\)).  Note that the reported values to the left and right of the slash ``/" are the ${\rm CD}_{min}$ and ${\rm CD}_{avg}$, respectively. The \textbf{bold} numbers represent the best results of the unsupervised methods.}
	\label{pcn_results}
	\begin{tabular}{l|c|c|c|c|c|c|c|c|c|c}
		\toprule[1.2pt]
		Methods             & Supervised & Average & Plane & Cabinet & Car & Chair & Lamp	& Couch	& Table	& Watercraft \\ \hline 
		FoldingNet \cite{11}& yes    & 7.7 & 2.4 & 8.4 & 4.9    & 9.2    & 11.5	& 9.6	& 8.4	& 7.4    \\
		PCN \cite{12}      & yes & 7.5 & 3.0 & 7.5 & 5.7    & 9.7    & 9.2	& 9.5	& 9.2	& 6.2 \\
		TopNet \cite{17}      & yes & 7.3 & 2.3 & 8.2 & 4.7    & 8.6    & 11.0	& 9.3	& 8.0	& 6.3    \\ 
		SnowflakeNet \cite{18}      & yes & 4.5 & 1.3 & 6.7 & 4.3    & 5    & 4.3	& 6.8	& 4.4	& 3.6    \\ \hline
		Cycle \cite{2}   & no & 14.4 & 4.1 & \textbf{14.2} & \textbf{9.9} & 14.6    & 19.2	& 27.8	& 16.8	& 9.0    \\
		Inversion \cite{3}    & no & 14.1 & 3.9 & 17.4    & 11.0 & 13.8 & 14.2	& 23.0	& 20.3	& 9.7 \\ \hline
		Cross-PCC (ours)          & no	& \textbf{9.2/9.9}	&  \textbf{2.5/2.8} & \textbf{13.9}/14.7 & 10.6/11.1 & \textbf{11.7/12.4} & \textbf{6.7/7.9} & \textbf{13.5/14.3} & \textbf{8.7/9.8} & \textbf{6.1/6.9}	 \\ \bottomrule[1.2pt]
	\end{tabular}
\end{table*}

\subsection{Dataset and Evaluation Metric}
To evaluate our method comprehensively, we experiment on both synthetic data and real-world data. For the synthetic point clouds, we use the PCN dataset \cite{12} and the 3D-EPN dataset \cite{8} provided by PCL2PCL \cite{1}.
In both datasets, each object contains eight partial point clouds, which are generated by back-projecting depth images of eight view angles into 3D space. 
We set the number of partial points to 2048 by upsampling or downsampling. The point numbers of complete shapes in 3D-EPN and PCN are 2048 and 16384, respectively.

For the view images, as Figure \ref{render} shows, the existing rendered RGB images are either too unreal \cite{20,35,insafutdinov2018unsupervised} or the objects in the images are too small \cite{6,xu2019disn}, which may lose some details and waste many convolutional parameters to compute the background pixels. To acquire high-quality image data, we re-render RGB images from ShapeNet \cite{chang2015shapenet} by Blender\footnote{\url{https://www.blender.org/}}.
The image size is $224 \times 224 \times 3$. We set the object to cover the largest space in the image while keeping all foreground pixels within the image boundary. For each object, we render eight RGB images from different view angles. Specifically, we sample 4 angles uniformly along the \(y\)-axis, and one angle each along the \(x\)-axis in the positive and negative directions.

The evaluation metric we employ for synthetic scans is CD. Considering that each object contains eight view images captured from different view angles, we report two kinds of metrics named ${\rm CD}_{min}$  and ${\rm CD}_{avg}$. Specifically, ${\rm CD}_{min}$ computes the minimum CD from all view images of each object. It measures which view image of an object can generate the best point cloud. ${\rm CD}_{avg}$ computes the average CD among all view images of each object. The formulas of the ${\rm CD}_{min}$ and ${\rm CD}_{avg}$ are listed as follows.
\begin{equation}
\begin{split}
	{\rm CD}_{min}=\frac{1}{J}\sum_{j=1}^{J}{\mathop{\texttt{min}}\limits_{i=1,...,I}{\rm CD}(\mathbf{P}_{out}(j,i), \mathbf{P}_{comp}(j))},
	\\	
	{\rm CD}_{avg}=\frac{1}{I\times J}\sum_{j=1}^{J}{\sum_{i=1}^{I}{\rm CD}(\mathbf{P}_{out}(j,i), \mathbf{P}_{comp}(j))},
\end{split}
\end{equation} 
where $J$ denotes the number of objects, $I$ represents the number of view images for each object, and $\mathbf{P}_{comp}$ means the complete and clean point cloud.

For the real-world dataset, we use KITTI \cite{geiger2012we}, which contains both the raw point clouds and RGB scene images.
Following previous works \cite{1,3}, we only evaluate our model on the car category. The partial point clouds we used are extracted from the car objects with more than 100 points in the Velodyne data. As for the view images, we crop and segment the cars in the images provided by KINS \cite{QiKINS}. These car images are captured from various view angles. In the raw images of KITTI, the cars vary in size and occlusion degree.  
To acquire high-quality view images, considering both the completeness of the objects and image resolutions, we only select the cars whose lengths of bounding boxes are larger than 100 pixels and are not occluded by other objects. Totally, we get 156 car images and partial point clouds. 
Since there are no complete point clouds in KITTI, following the previous papers \cite{12,1,3,15}, we employ the Minimal Matching Distance (MMD) as our evaluation metric when evaluating our model on KITTI. The MMD is the minimum CD between each predicted result and the complete car point clouds from the ShapeNet. It measures the similarity between the predicted shape and typical cars.

\begin{figure*}[htbp]
	\centering  
	\subfigbottomskip=1pt 
	\subfigure{
		\includegraphics[width=0.08\linewidth]{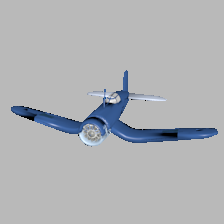}}
	\subfigure{
		\includegraphics[width=0.1\linewidth]{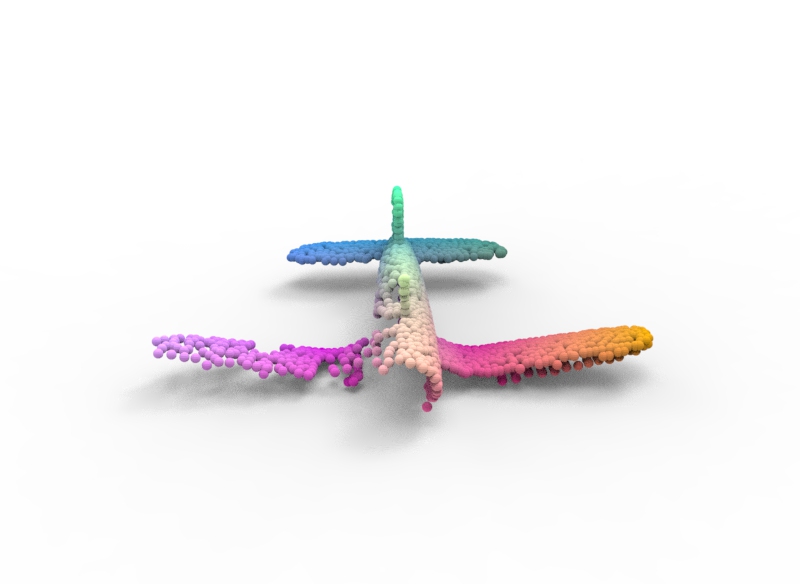}}
	\subfigure{
		\includegraphics[width=0.1\linewidth]{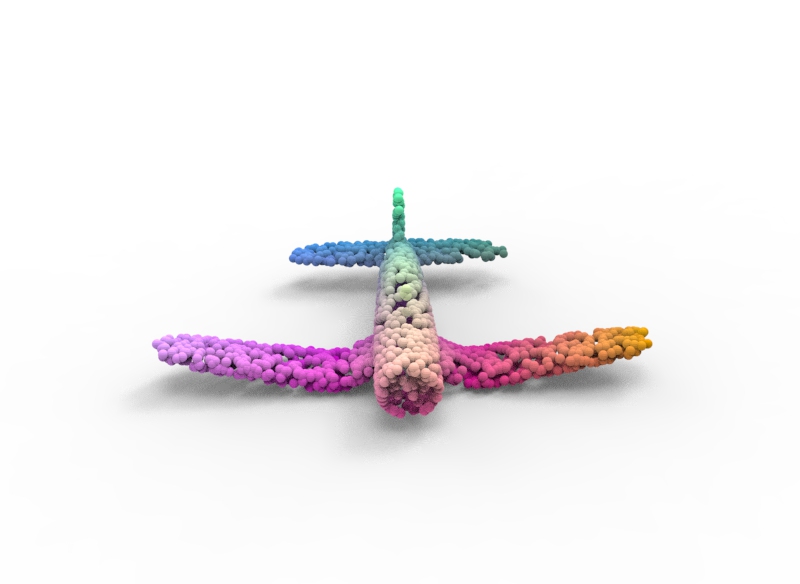}}
	\subfigure{
		\includegraphics[width=0.1\linewidth]{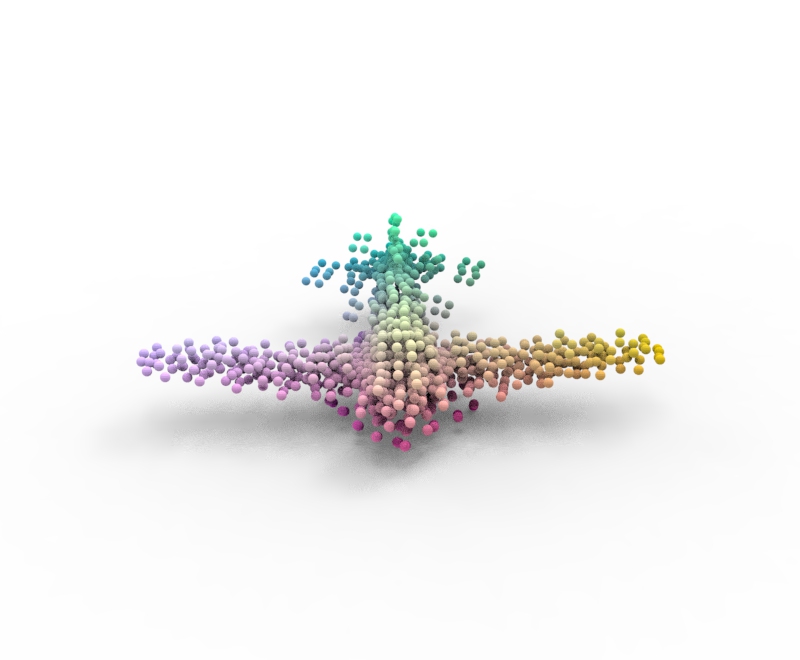}}
	\subfigure{
		\includegraphics[width=0.1\linewidth]{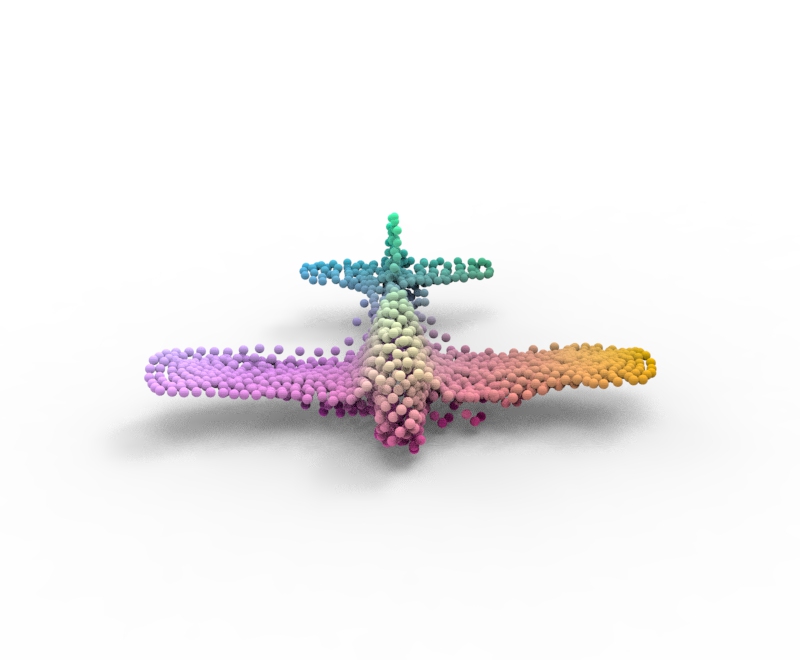}}
	\subfigure{
		\includegraphics[width=0.1\linewidth]{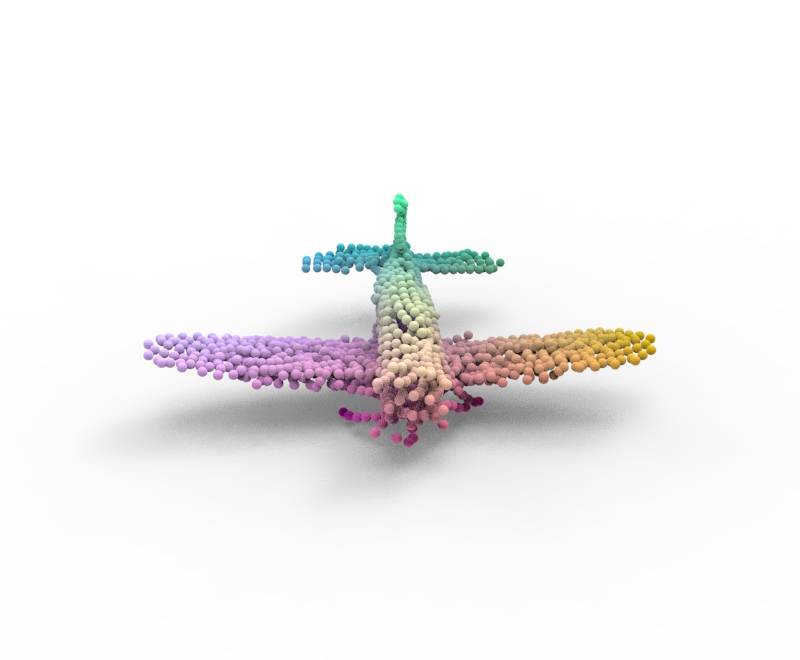}}
	\subfigure{
		\includegraphics[width=0.1\linewidth]{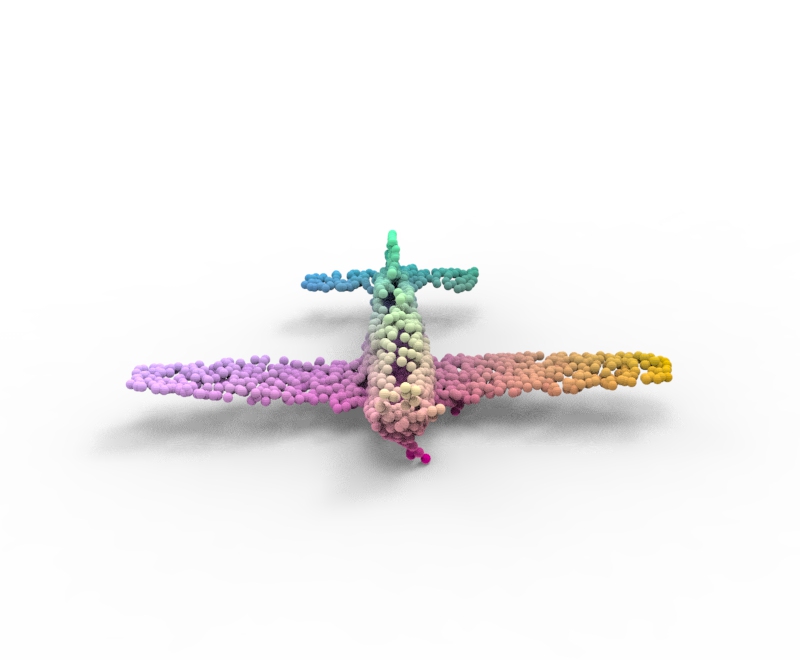}}
	\subfigure{
		\includegraphics[width=0.1\linewidth]{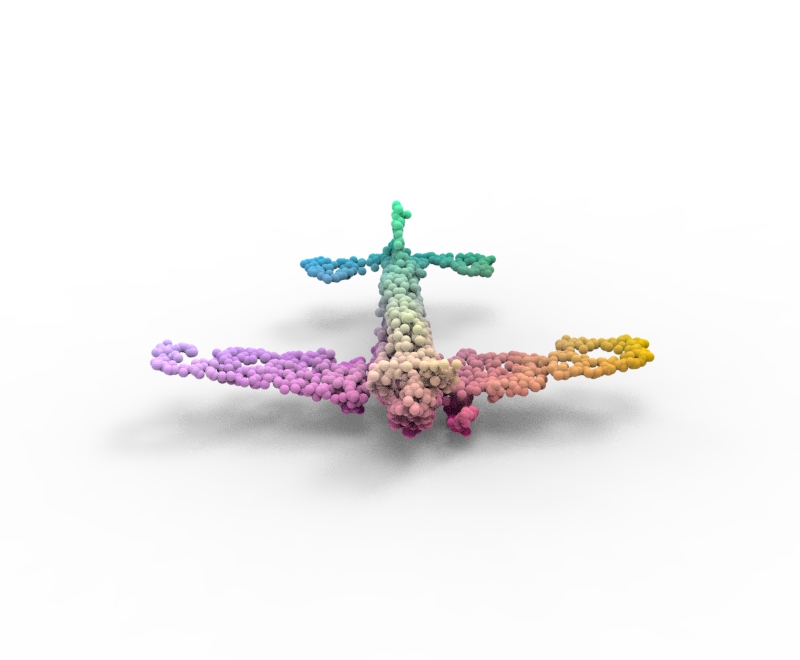}}
	\subfigure{
		\includegraphics[width=0.1\linewidth]{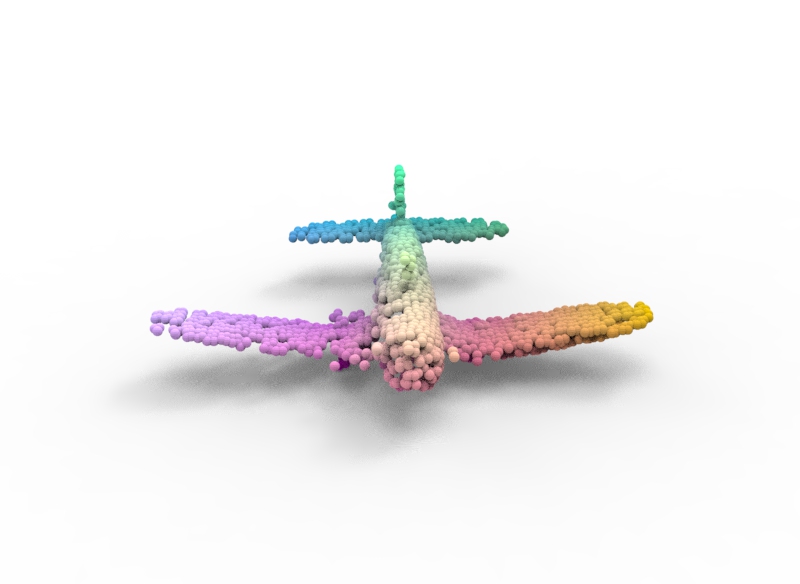}}
	\\
	\subfigure{
		\includegraphics[width=0.08\linewidth]{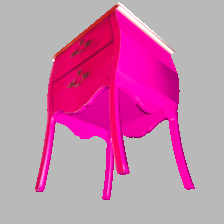}}
	\subfigure{
		\includegraphics[width=0.1\linewidth]{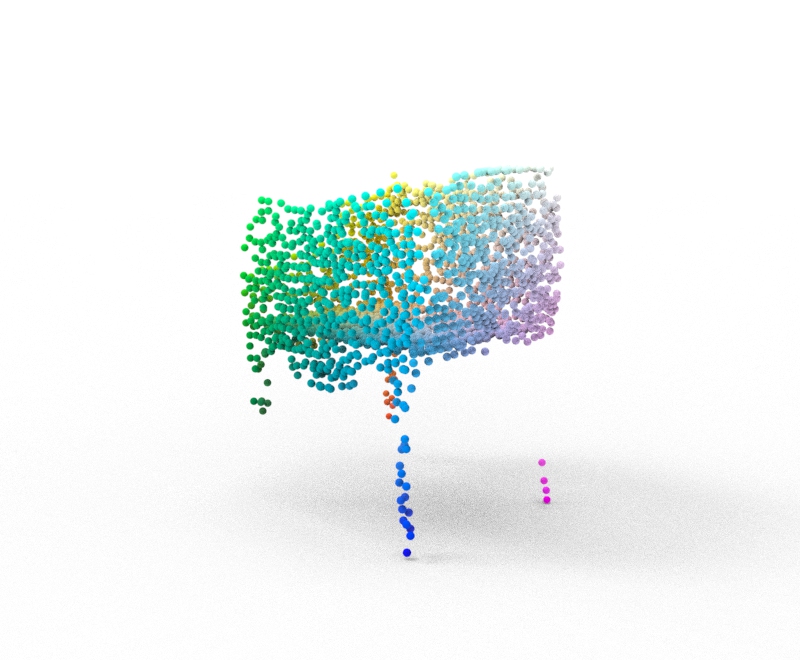}}
	\subfigure{
		\includegraphics[width=0.1\linewidth]{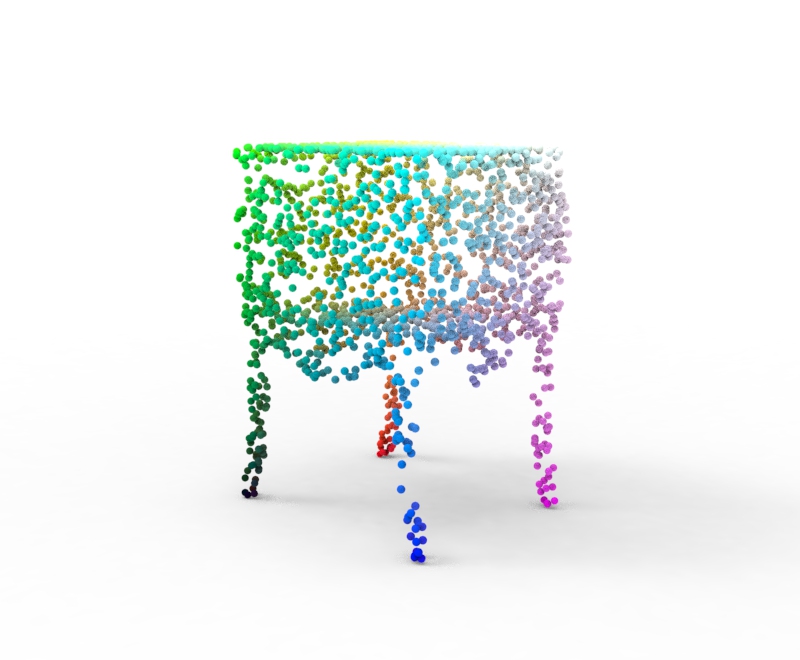}}
	\subfigure{
		\includegraphics[width=0.1\linewidth]{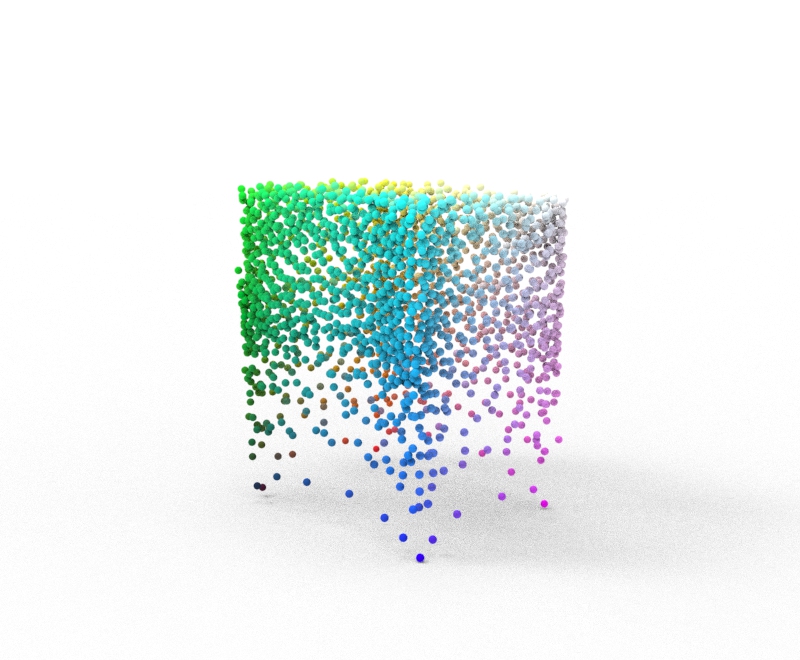}}
	\subfigure{
		\includegraphics[width=0.1\linewidth]{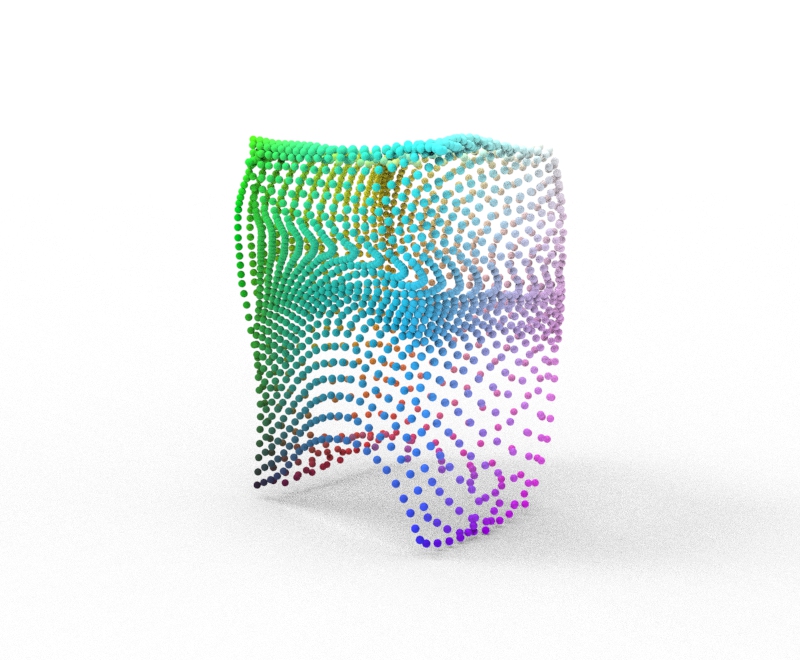}}
	\subfigure{
		\includegraphics[width=0.1\linewidth]{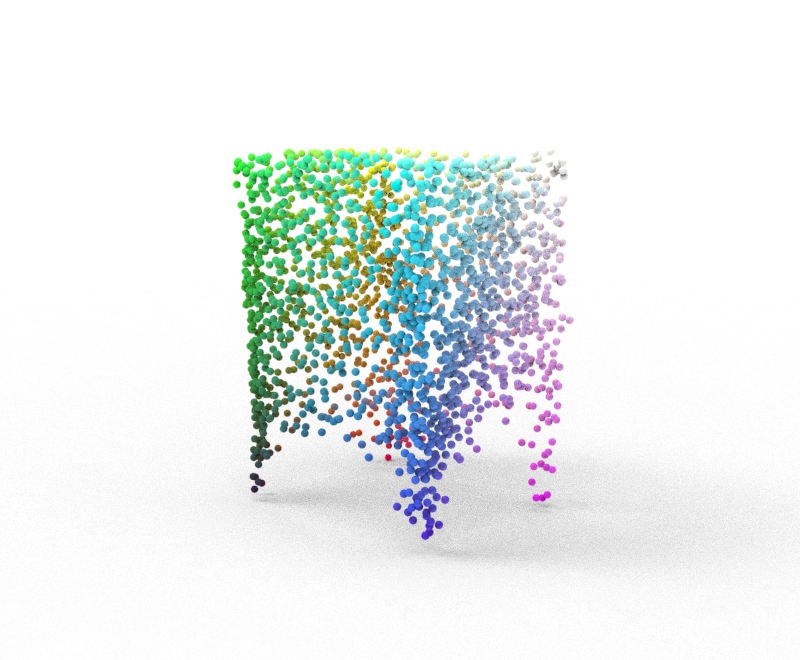}}
	\subfigure{
		\includegraphics[width=0.1\linewidth]{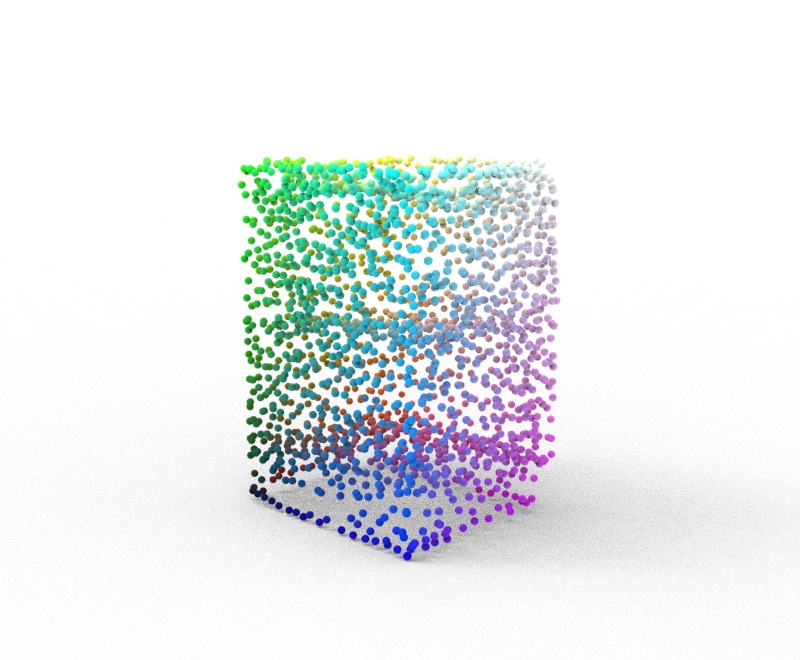}}
	\subfigure{
		\includegraphics[width=0.1\linewidth]{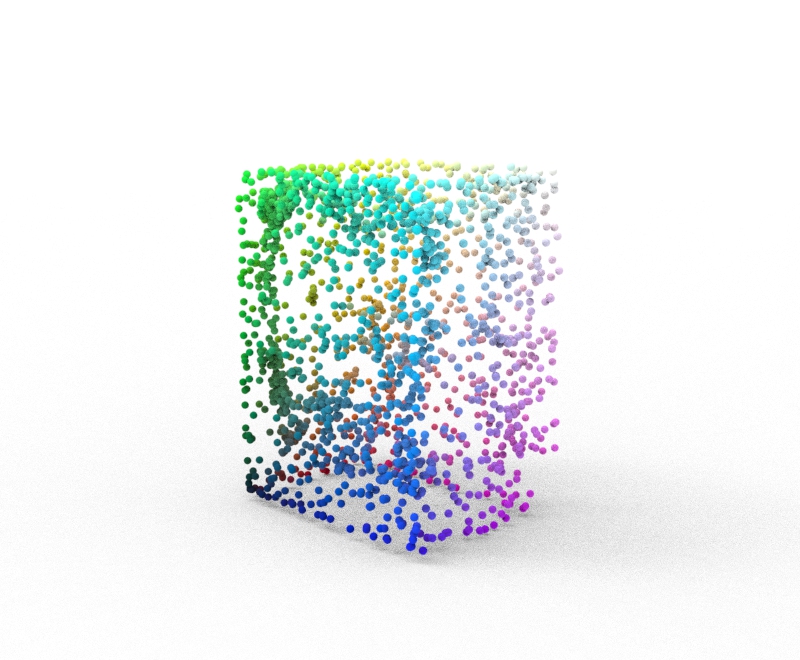}}
	\subfigure{
		\includegraphics[width=0.1\linewidth]{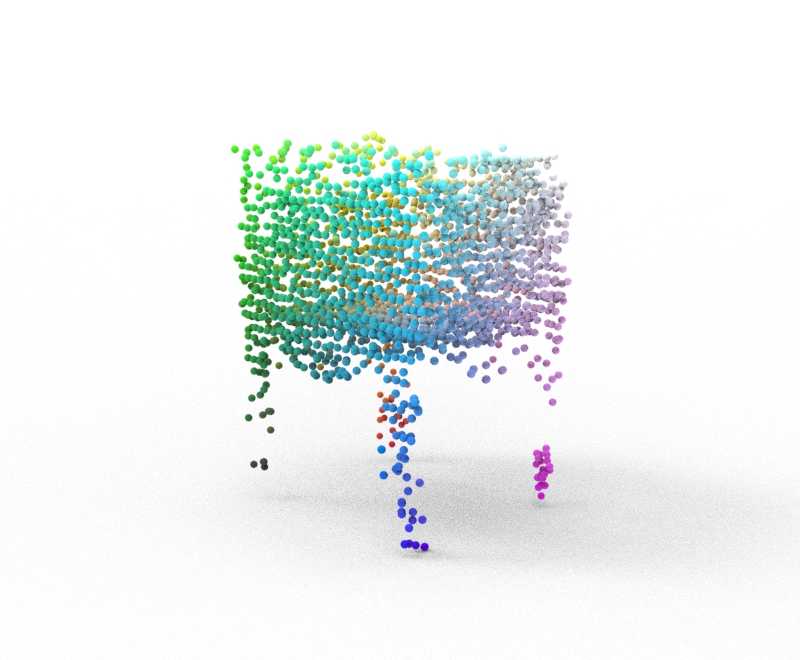}}
	\\
	\subfigure{
		\includegraphics[width=0.08\linewidth]{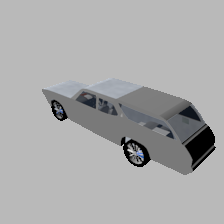}}
	\subfigure{
		\includegraphics[width=0.1\linewidth]{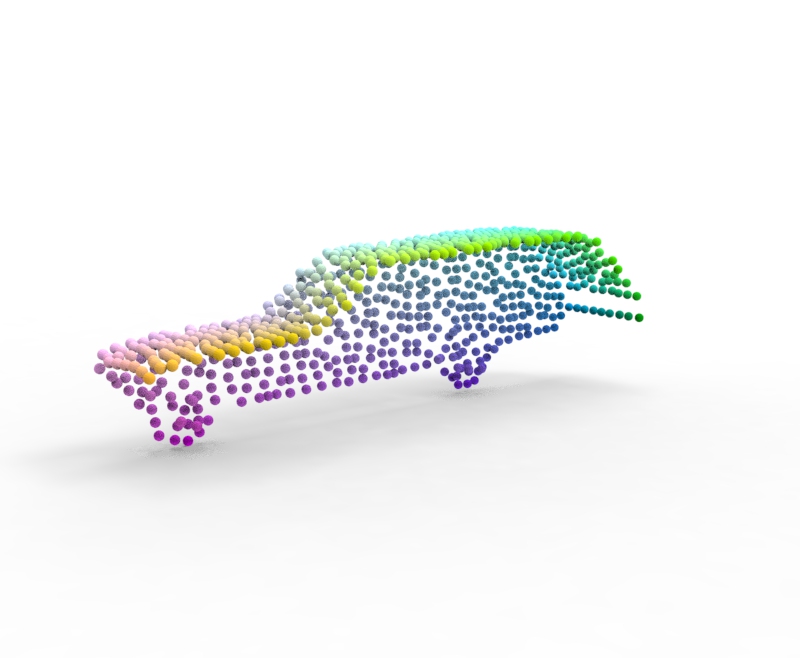}}
	\subfigure{
		\includegraphics[width=0.1\linewidth]{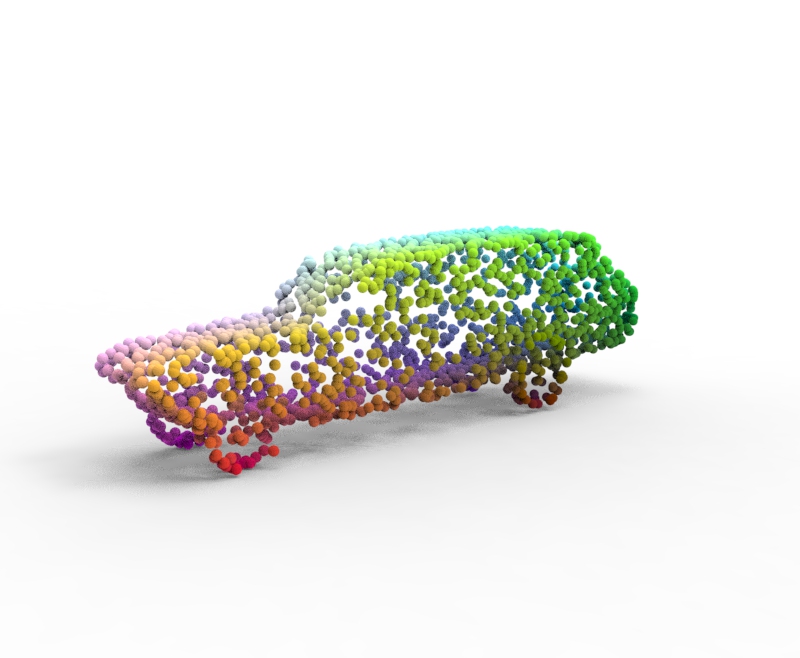}}
	\subfigure{
		\includegraphics[width=0.1\linewidth]{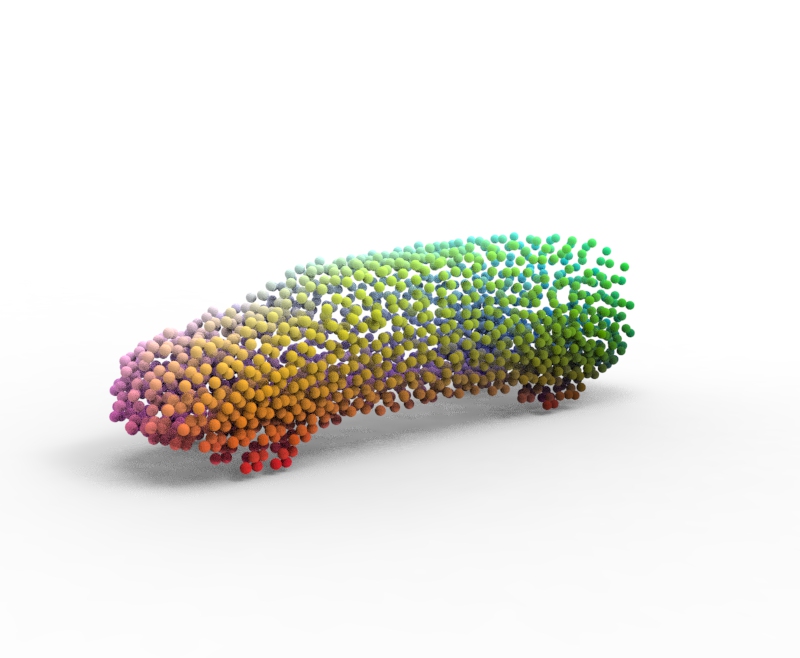}}
	\subfigure{
		\includegraphics[width=0.1\linewidth]{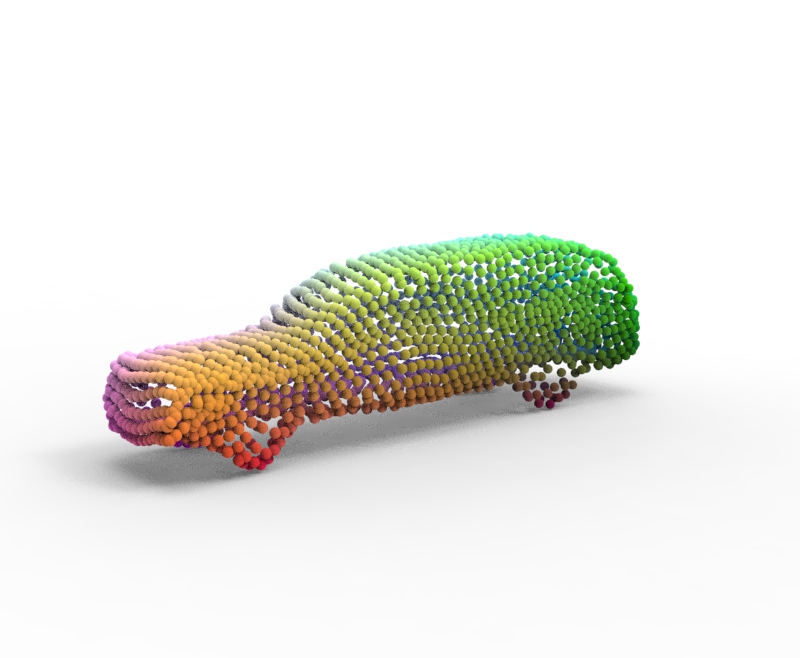}}
	\subfigure{
		\includegraphics[width=0.1\linewidth]{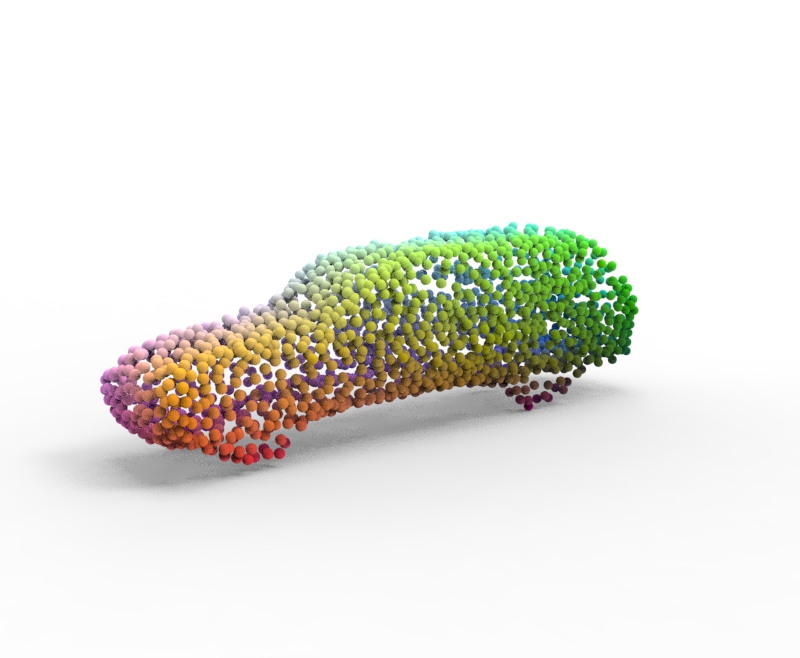}}
	\subfigure{
		\includegraphics[width=0.1\linewidth]{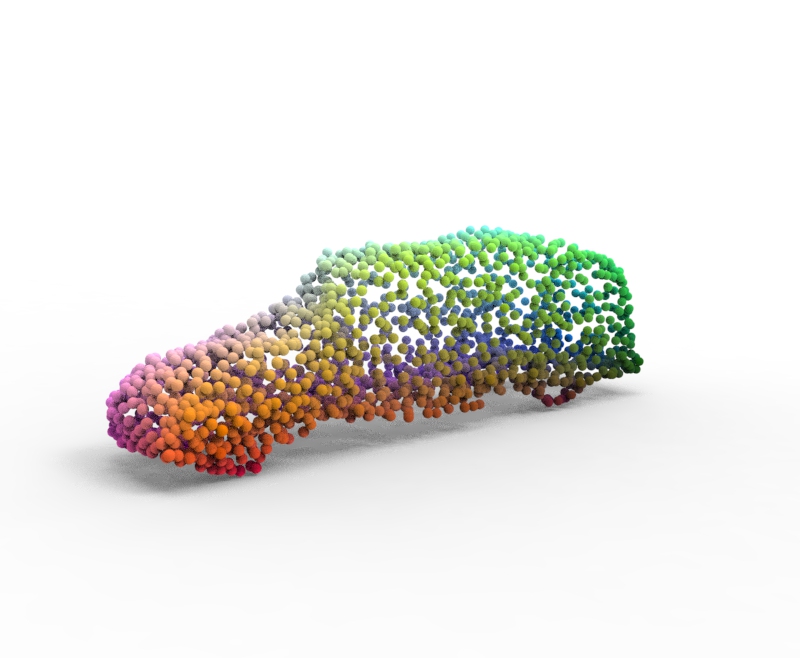}}
	\subfigure{
		\includegraphics[width=0.1\linewidth]{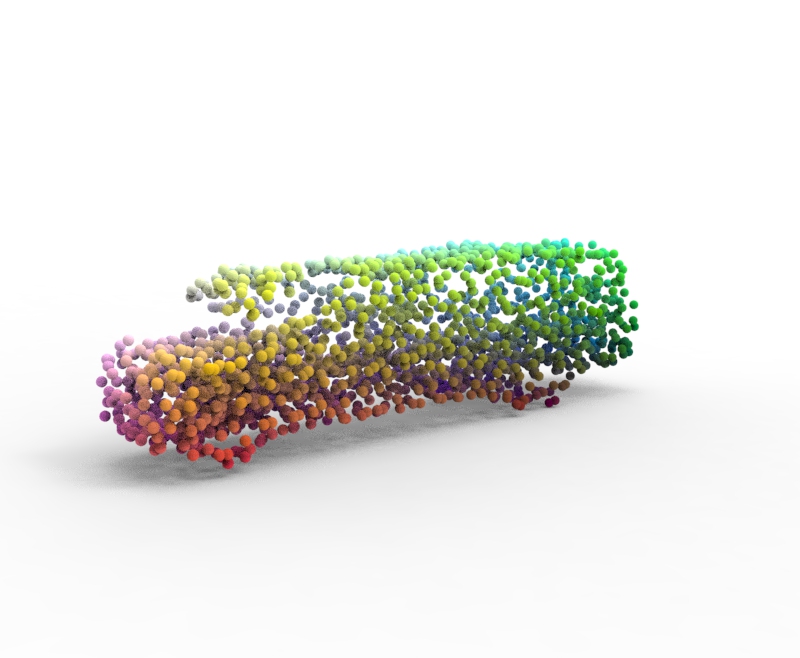}}
	\subfigure{
		\includegraphics[width=0.1\linewidth]{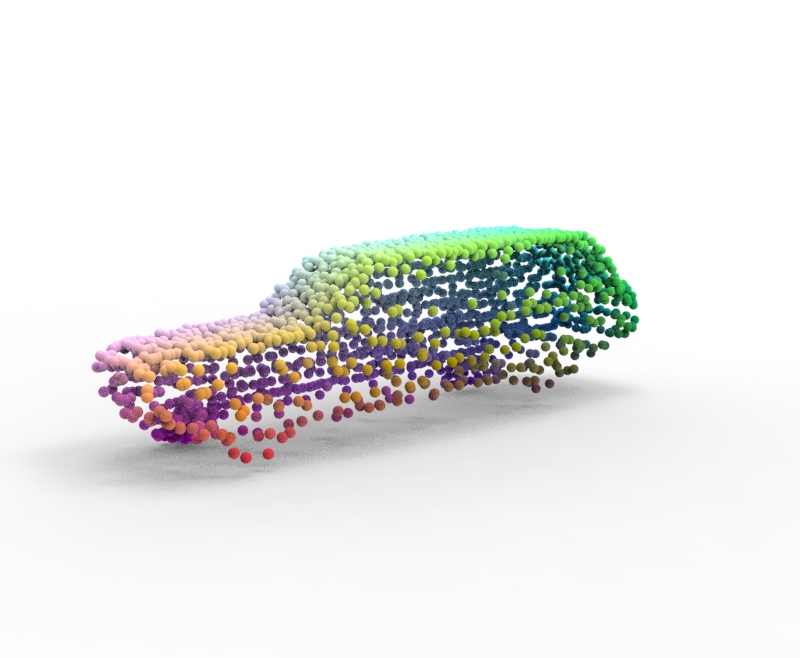}}
	\\
	\subfigure{
		\includegraphics[width=0.08\linewidth]{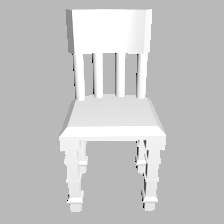}}
	\subfigure{
		\includegraphics[width=0.1\linewidth]{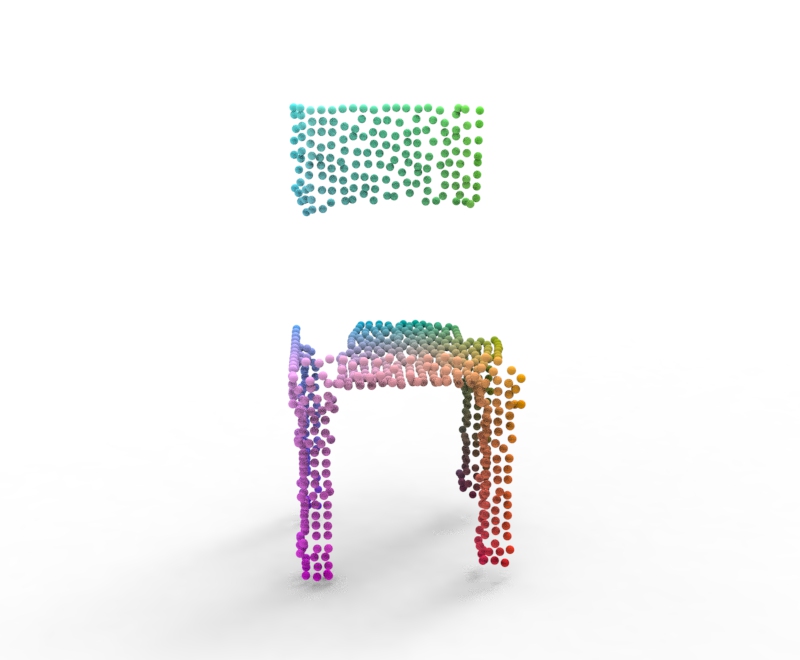}}
	\subfigure{
		\includegraphics[width=0.1\linewidth]{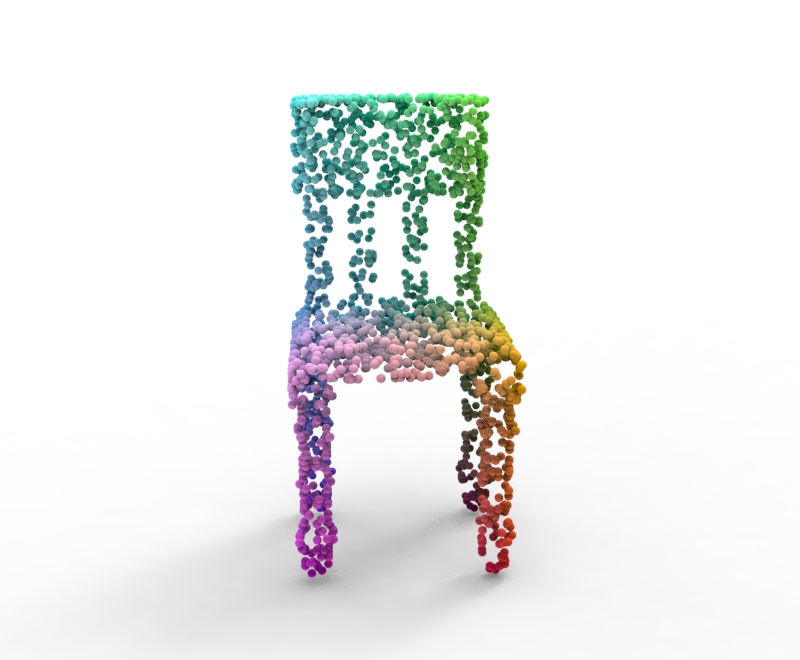}}
	\subfigure{
		\includegraphics[width=0.1\linewidth]{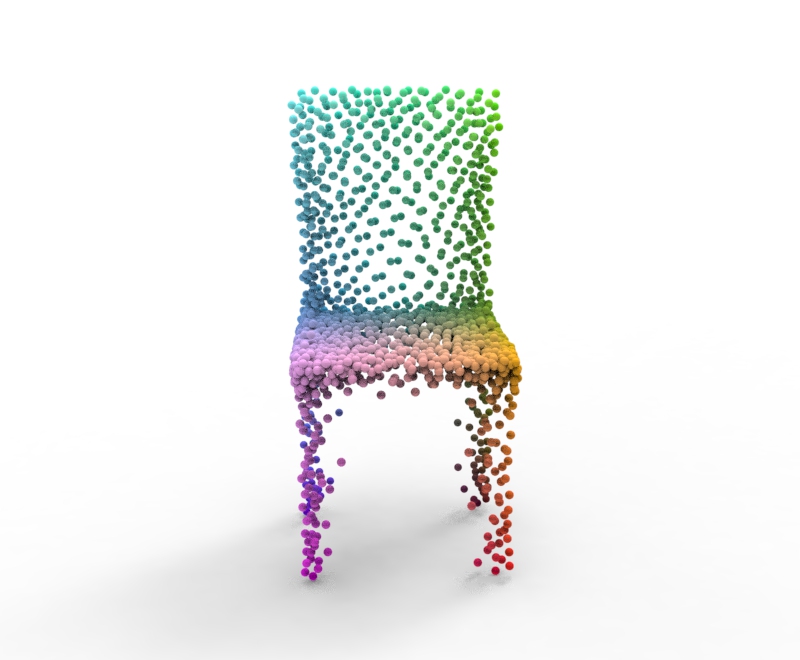}}
	\subfigure{
		\includegraphics[width=0.1\linewidth]{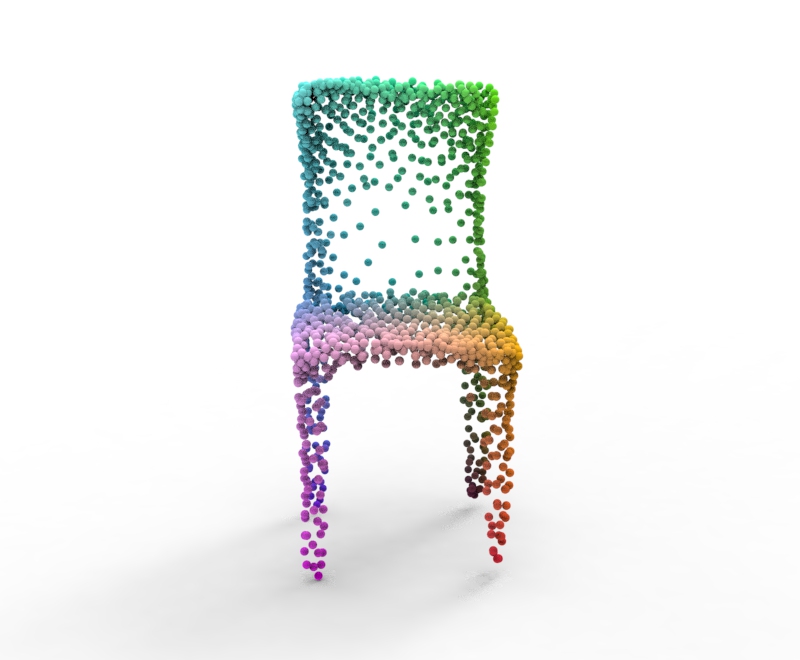}}
	\subfigure{
		\includegraphics[width=0.1\linewidth]{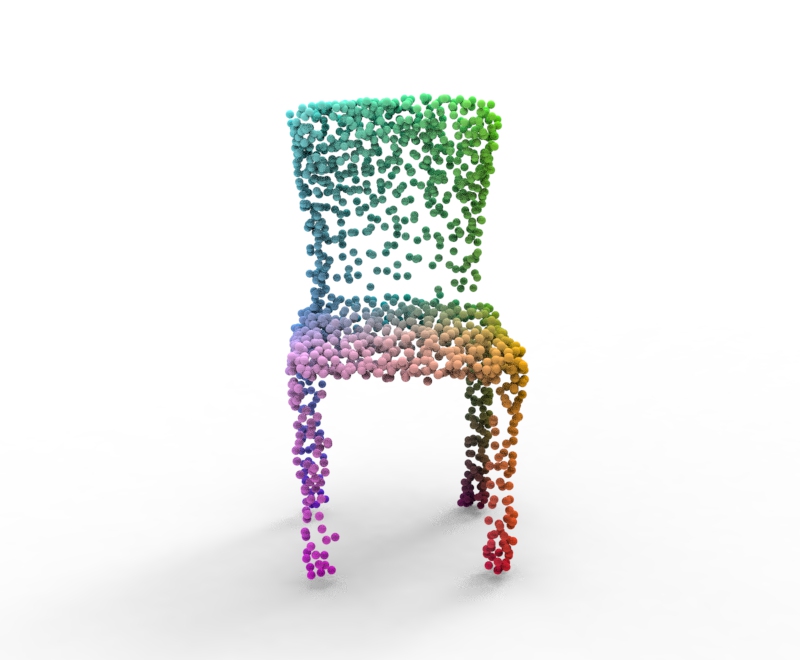}}
	\subfigure{
		\includegraphics[width=0.1\linewidth]{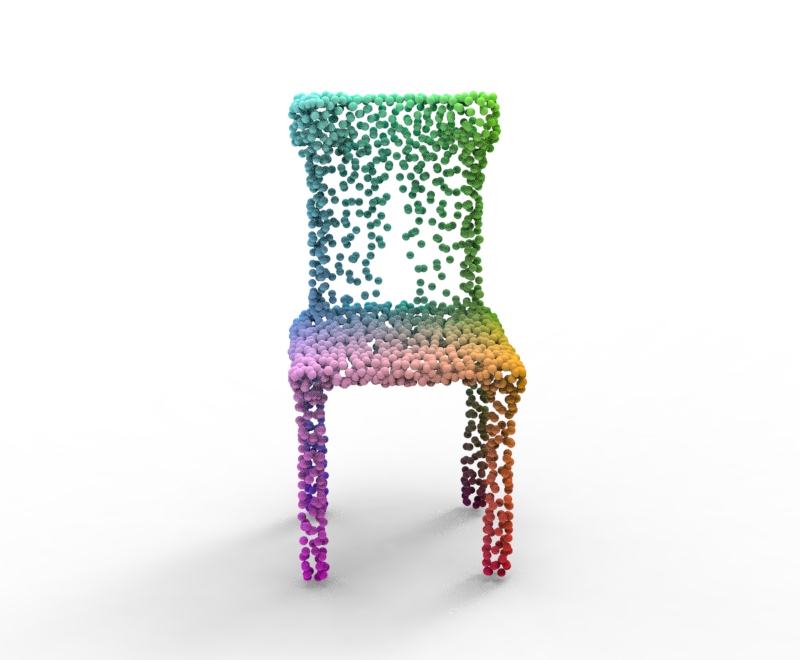}}
	\subfigure{
		\includegraphics[width=0.1\linewidth]{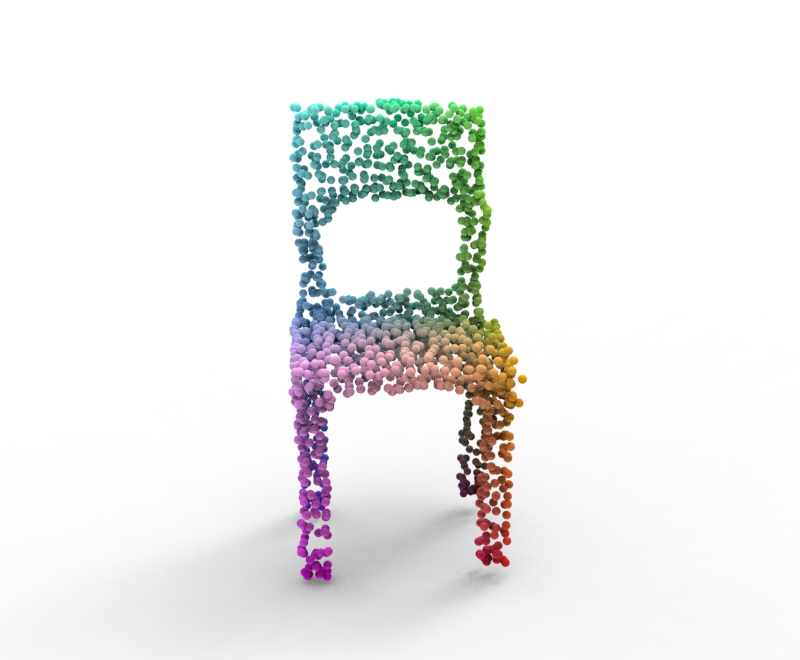}}
	\subfigure{
		\includegraphics[width=0.1\linewidth]{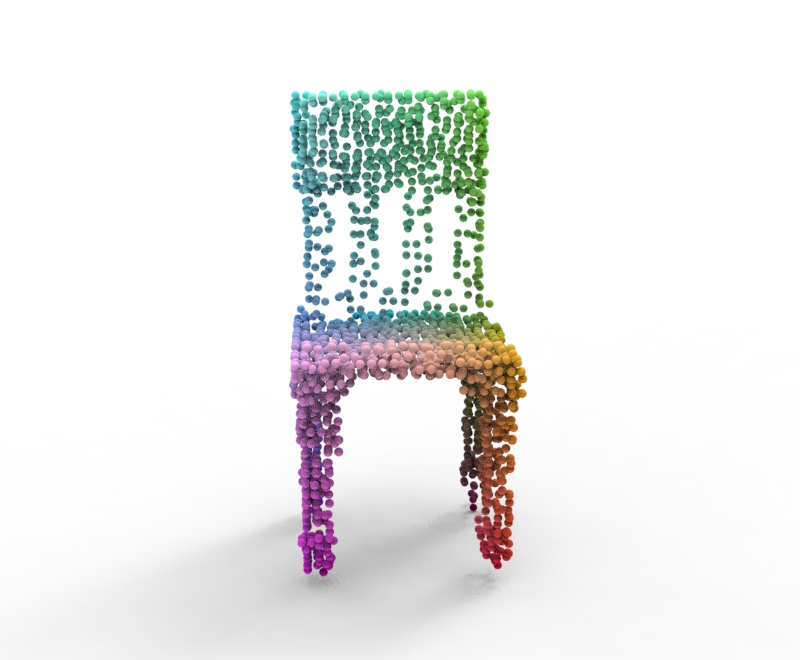}}
	\\
	\subfigure{
		\includegraphics[width=0.08\linewidth]{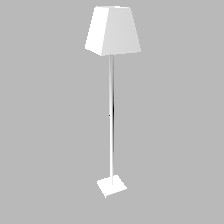}}
	\subfigure{
		\includegraphics[width=0.1\linewidth]{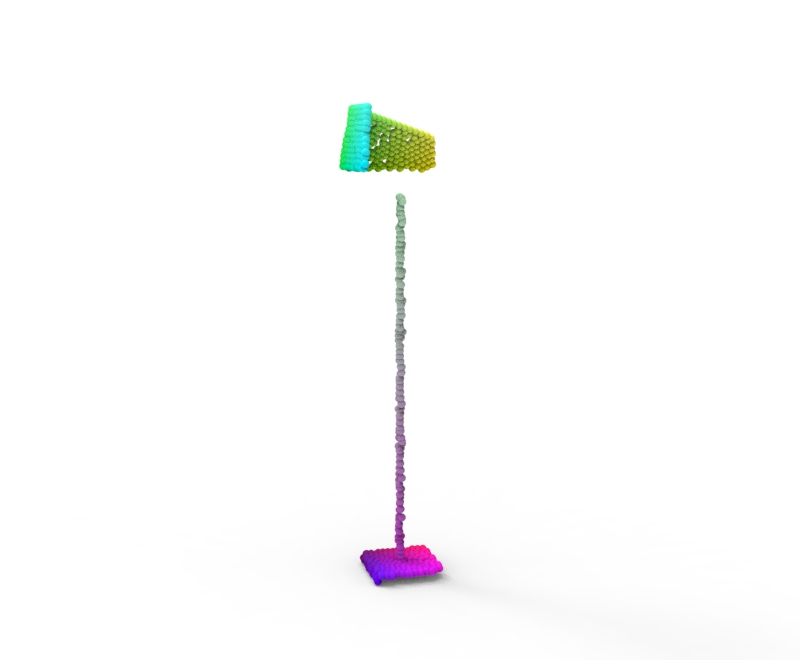}}
	\subfigure{
		\includegraphics[width=0.1\linewidth]{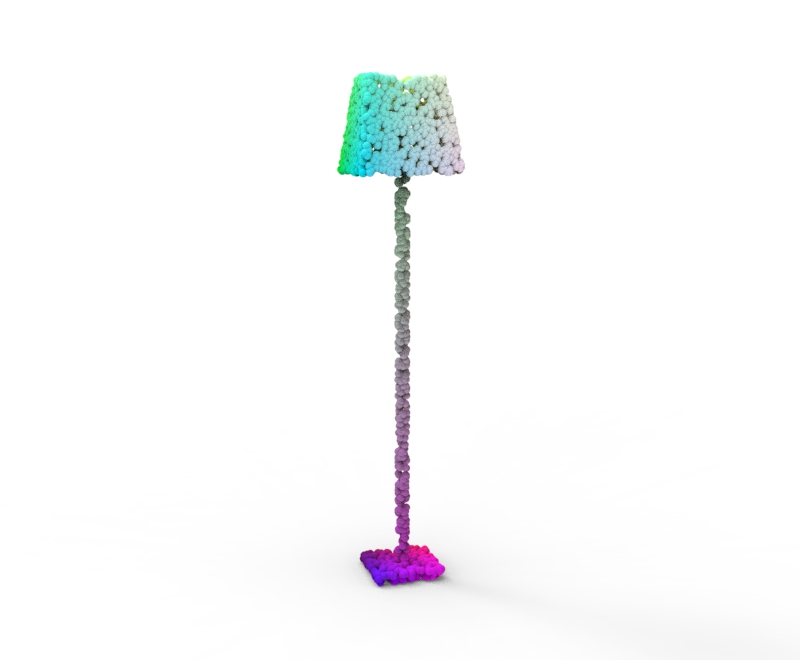}}
	\subfigure{
		\includegraphics[width=0.1\linewidth]{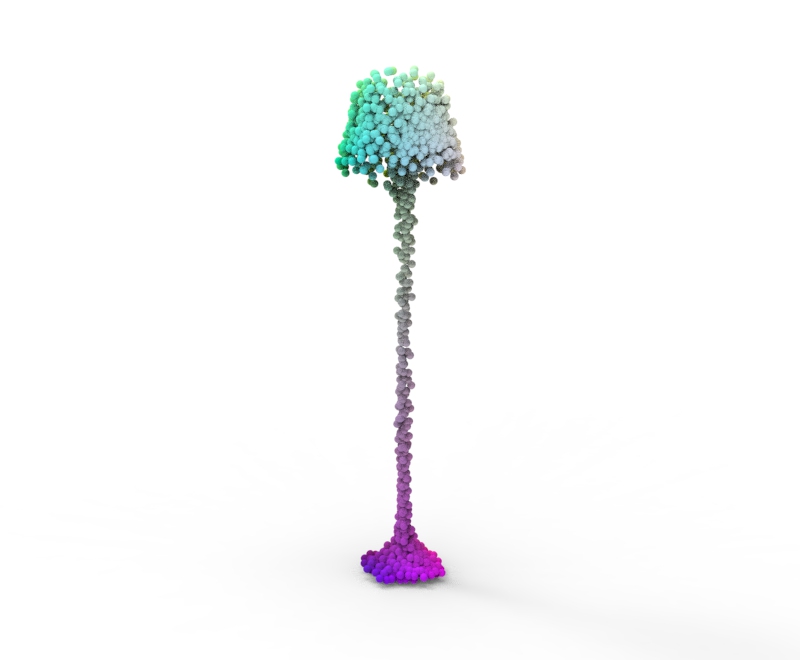}}
	\subfigure{
		\includegraphics[width=0.1\linewidth]{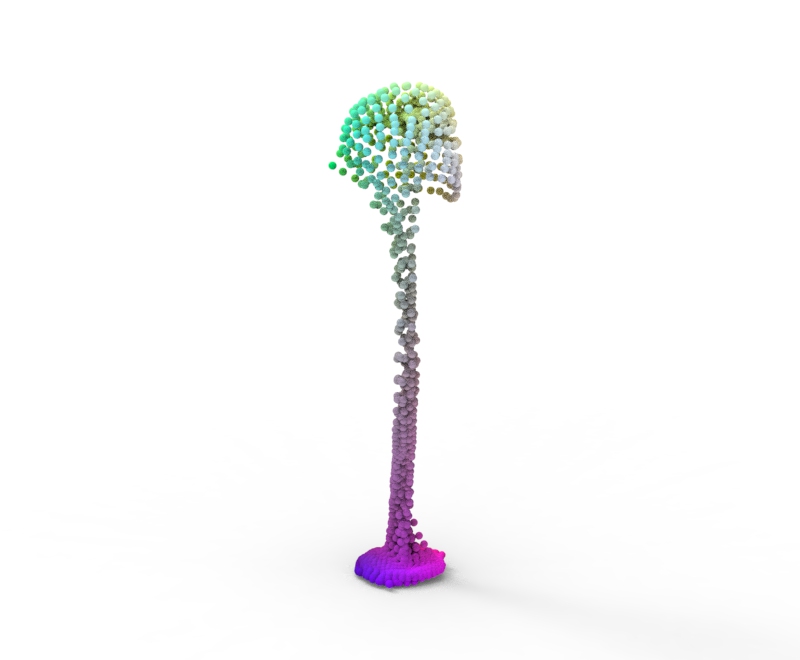}}
	\subfigure{
		\includegraphics[width=0.1\linewidth]{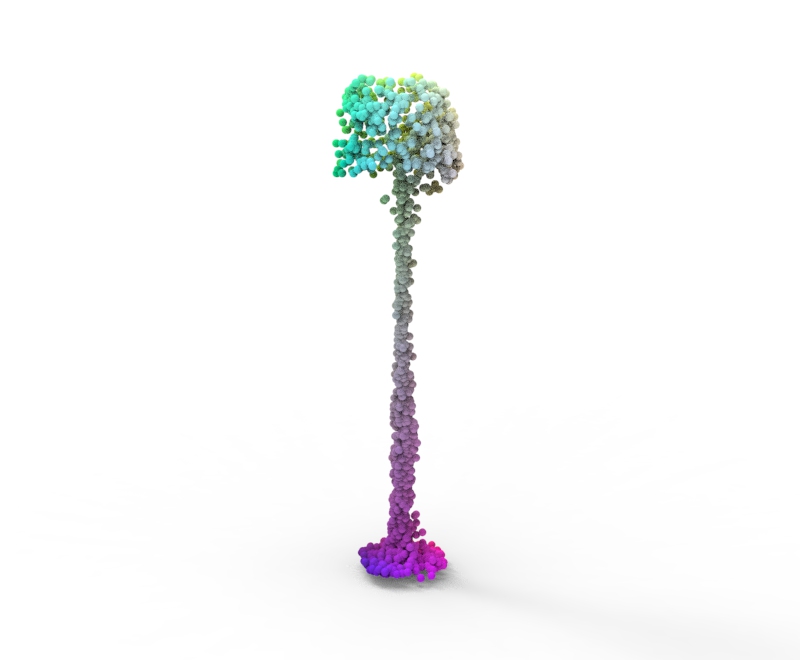}}
	\subfigure{
		\includegraphics[width=0.1\linewidth]{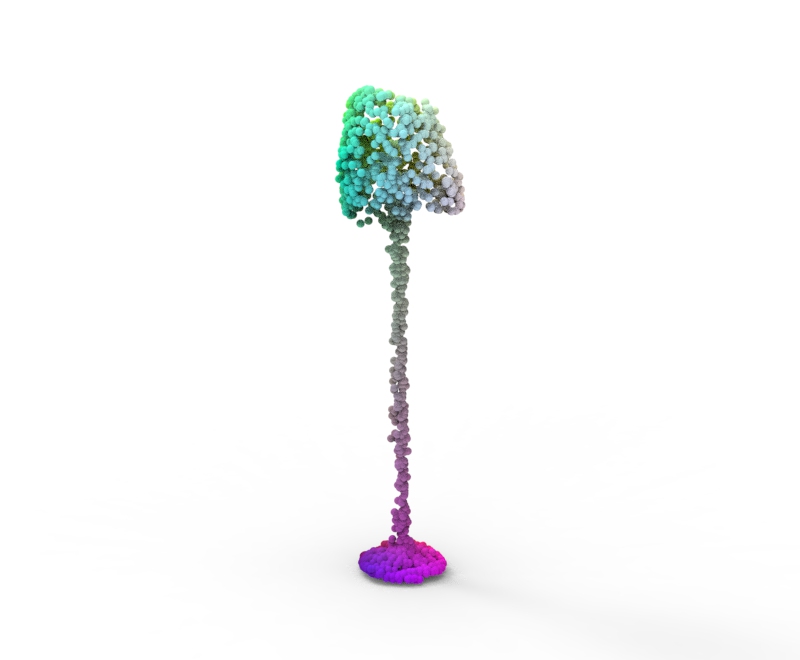}}
	\subfigure{
		\includegraphics[width=0.1\linewidth]{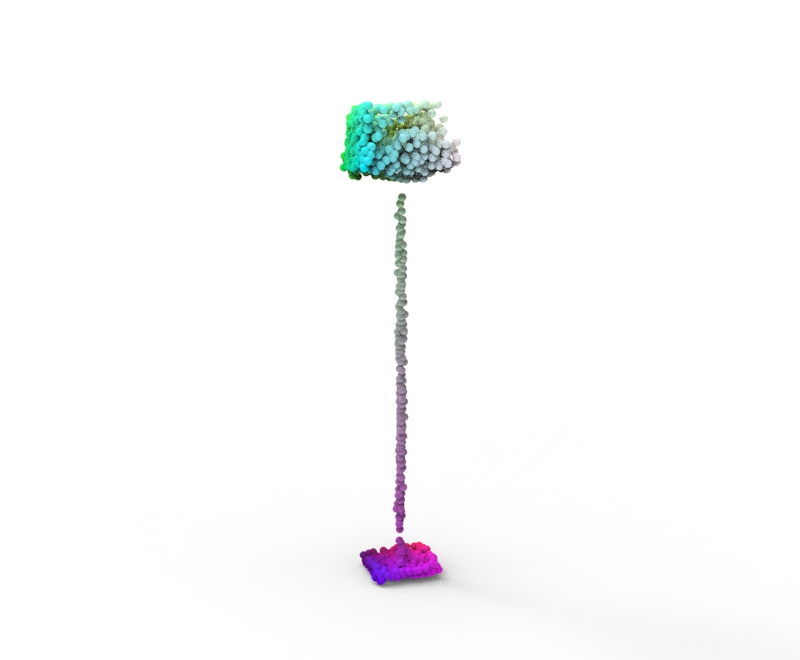}}
	\subfigure{
		\includegraphics[width=0.1\linewidth]{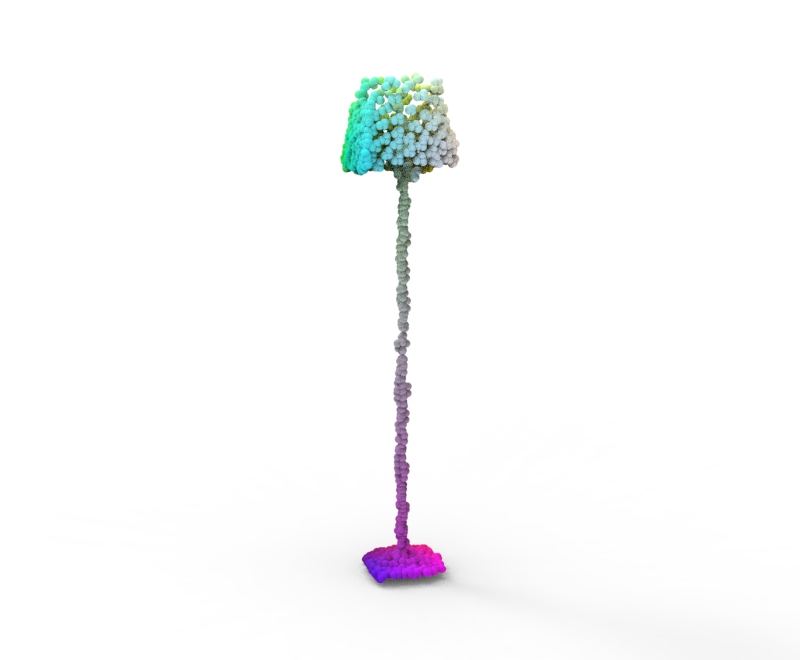}}
	\\
	\subfigure{
		\includegraphics[width=0.08\linewidth]{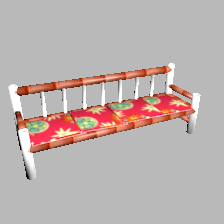}}
	\subfigure{
		\includegraphics[width=0.1\linewidth]{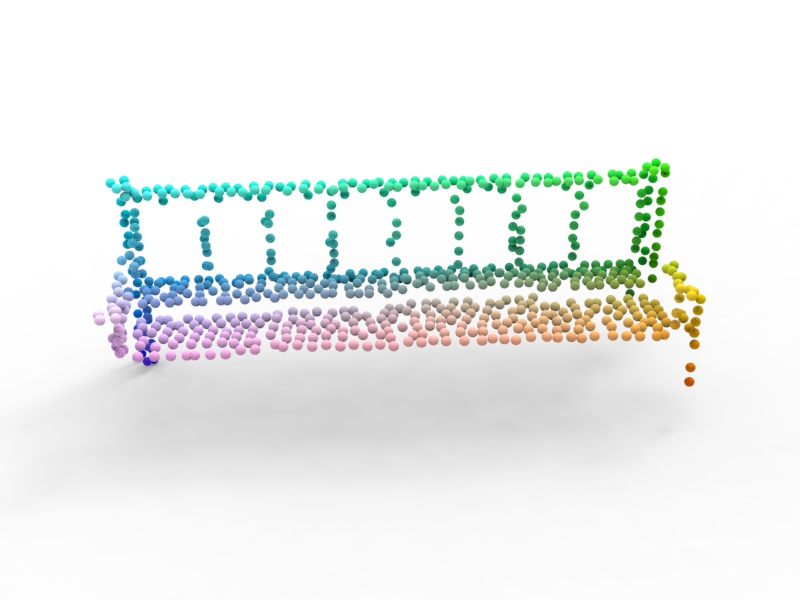}}
	\subfigure{
		\includegraphics[width=0.1\linewidth]{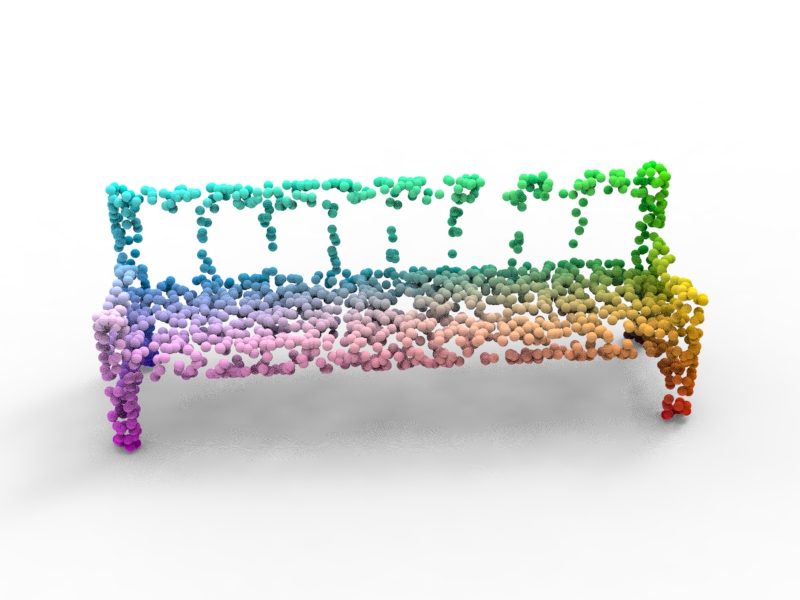}}
	\subfigure{
		\includegraphics[width=0.1\linewidth]{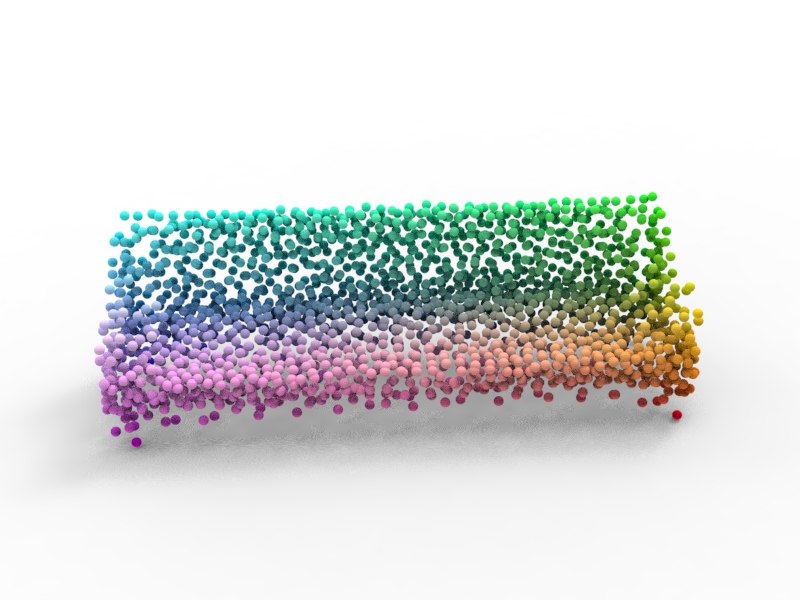}}
	\subfigure{
		\includegraphics[width=0.1\linewidth]{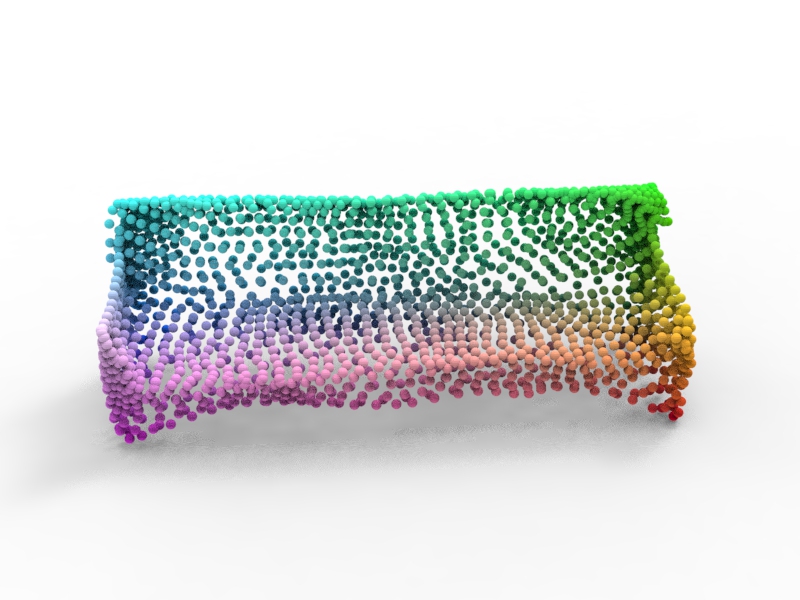}}
	\subfigure{
		\includegraphics[width=0.1\linewidth]{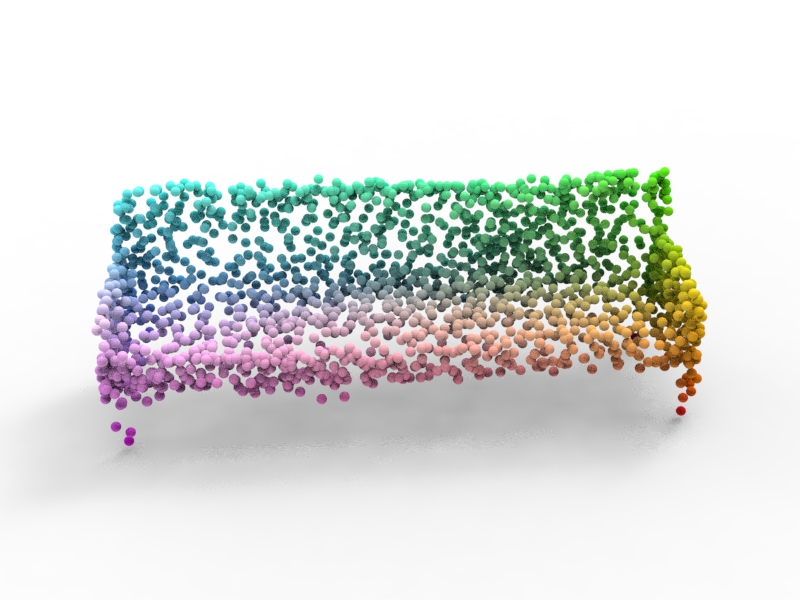}}
	\subfigure{
		\includegraphics[width=0.1\linewidth]{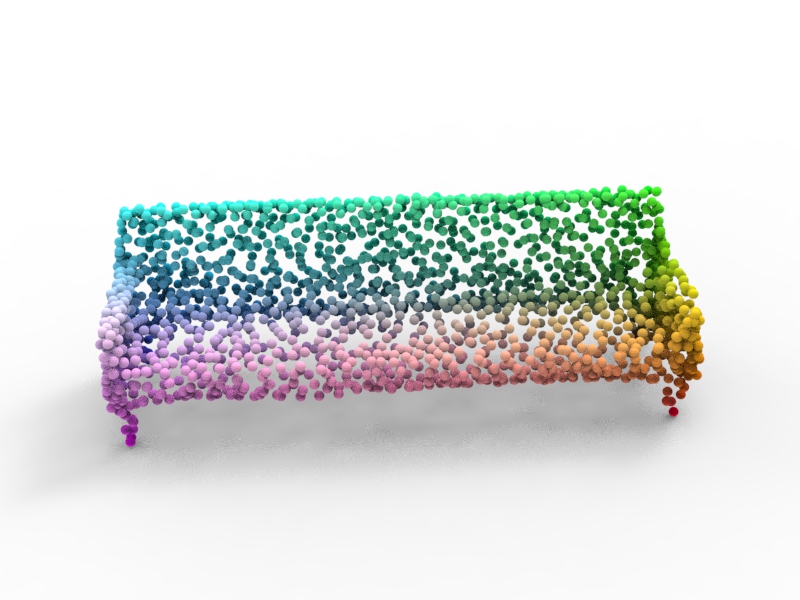}}
	\subfigure{
		\includegraphics[width=0.1\linewidth]{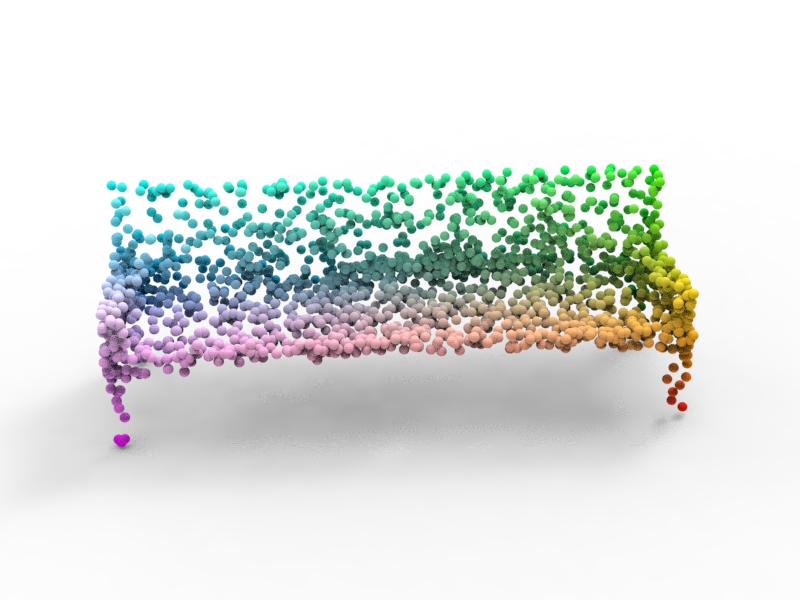}}
	\subfigure{
		\includegraphics[width=0.1\linewidth]{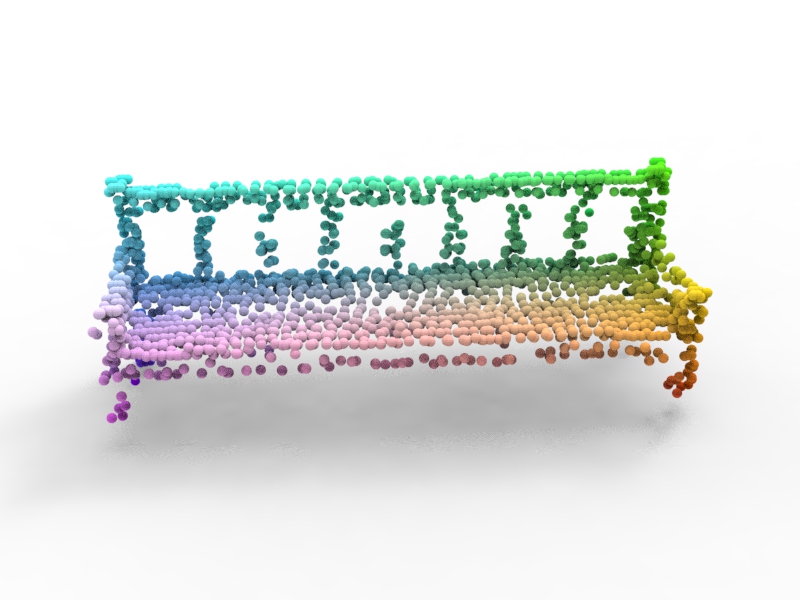}}
	\\
	\subfigure{
		\includegraphics[width=0.08\linewidth]{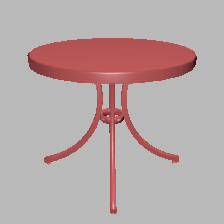}}
	\subfigure{
		\includegraphics[width=0.1\linewidth]{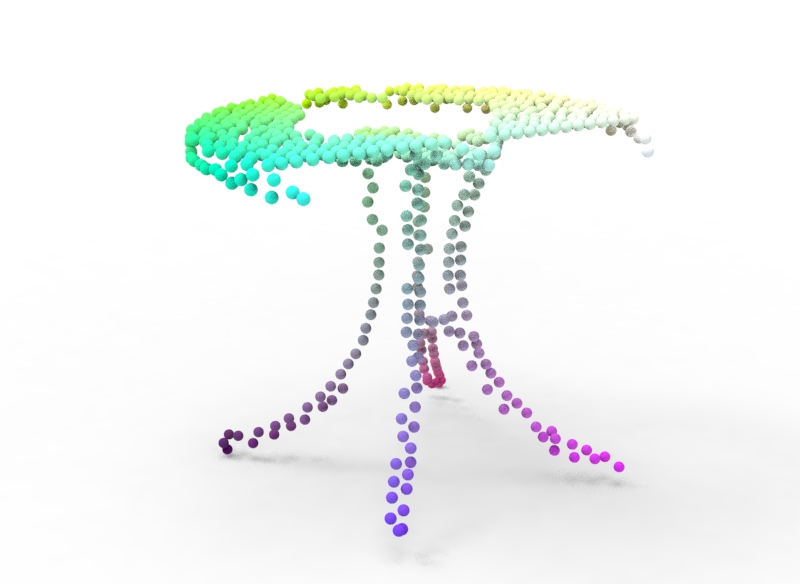}}
	\subfigure{
		\includegraphics[width=0.1\linewidth]{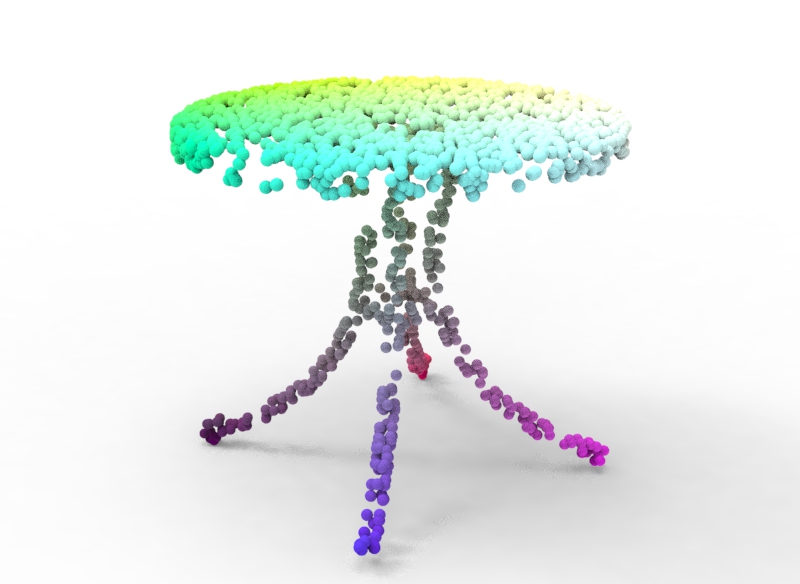}}
	\subfigure{
		\includegraphics[width=0.1\linewidth]{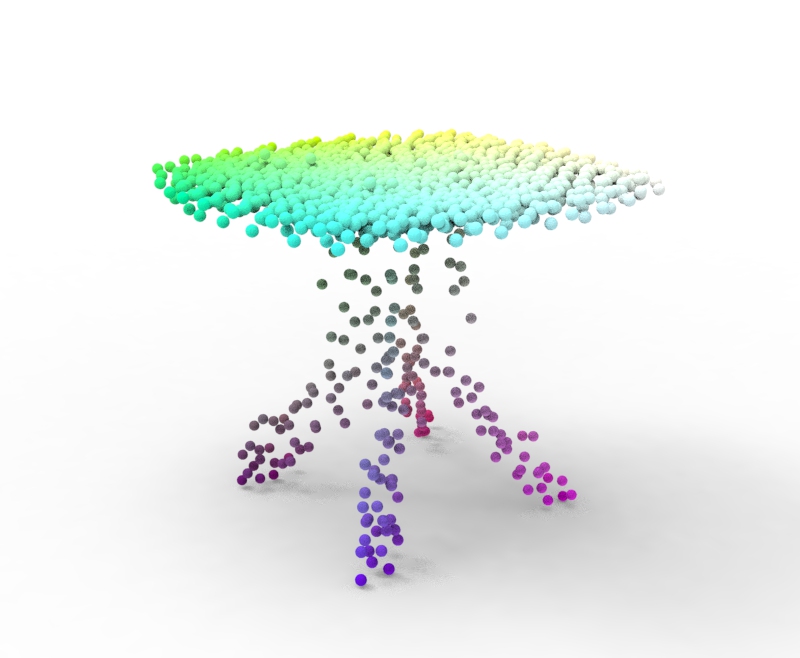}}
	\subfigure{
		\includegraphics[width=0.1\linewidth]{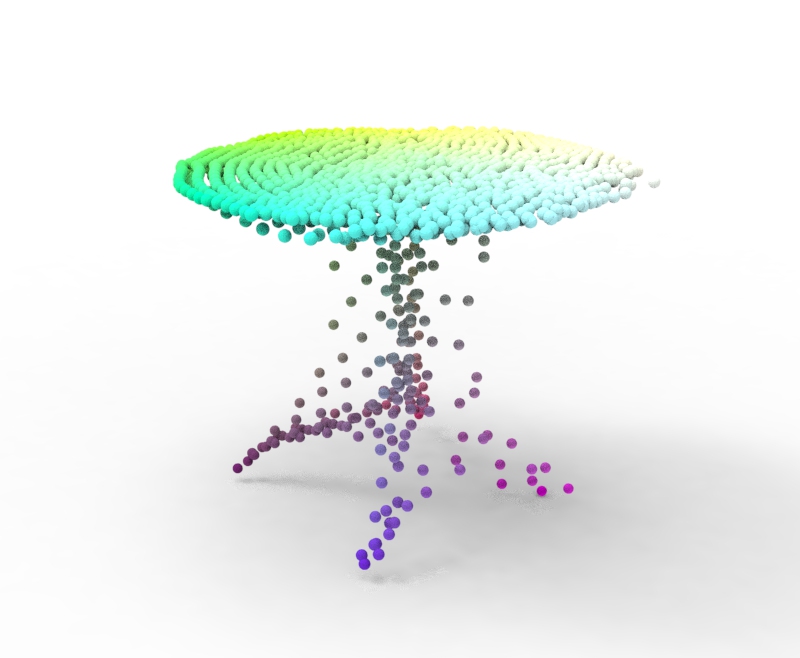}}
	\subfigure{
		\includegraphics[width=0.1\linewidth]{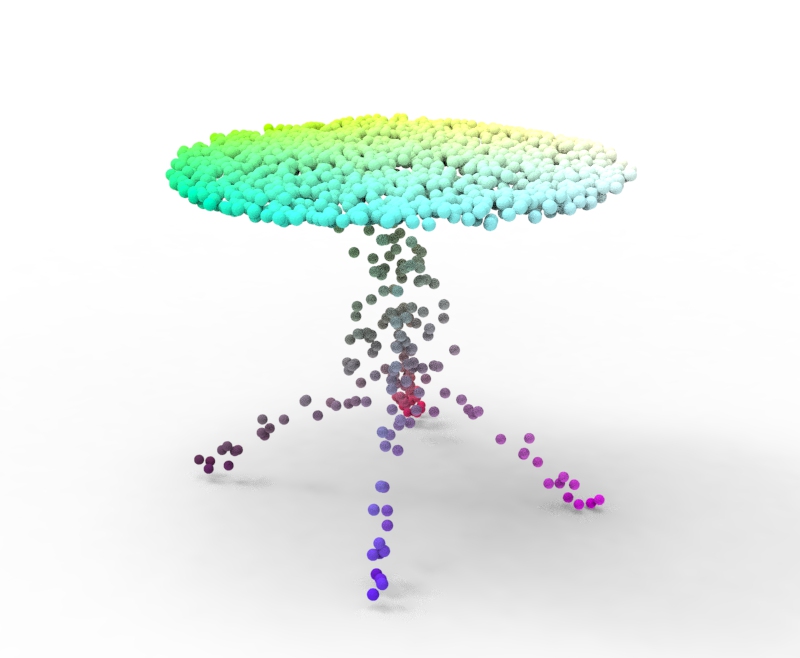}}
	\subfigure{
		\includegraphics[width=0.1\linewidth]{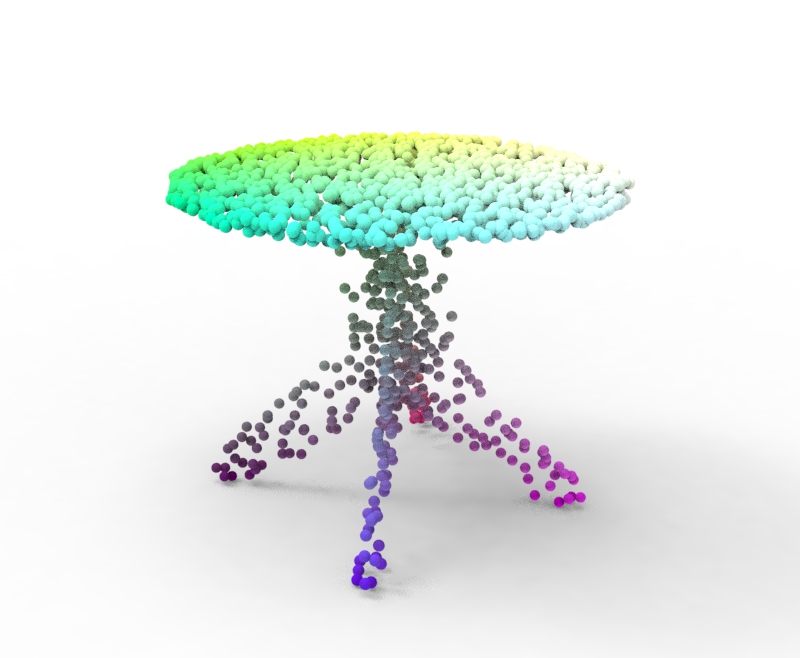}}
	\subfigure{
		\includegraphics[width=0.1\linewidth]{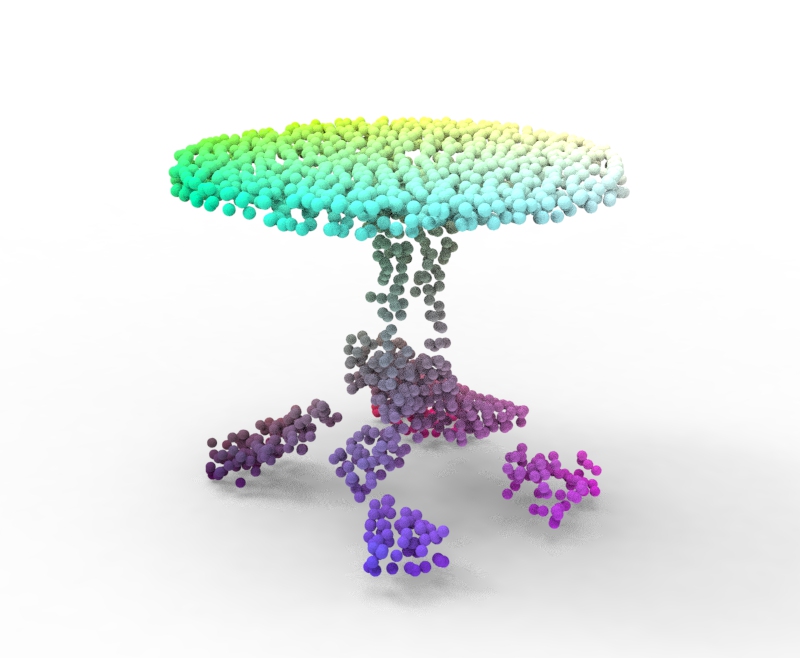}}
	\subfigure{
		\includegraphics[width=0.1\linewidth]{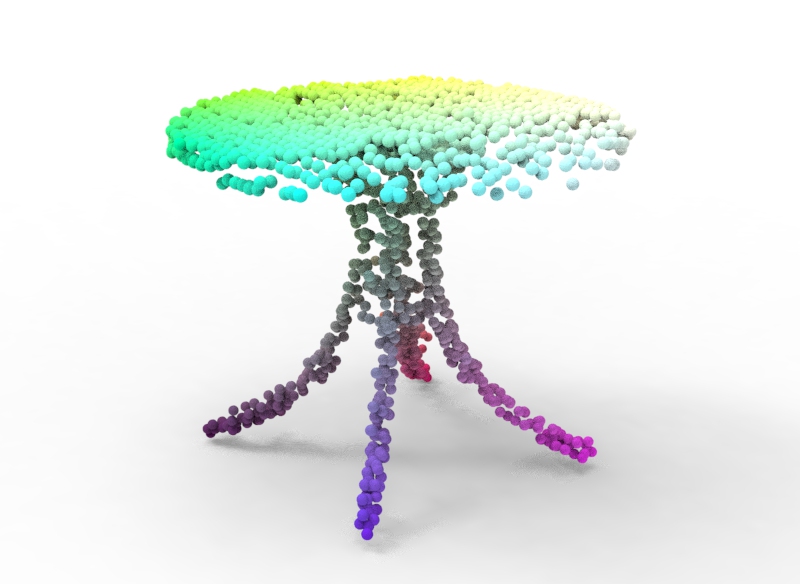}}
	\\
	\setcounter{subfigure}{0}
	\subfigure[]{
		\includegraphics[width=0.08\linewidth]{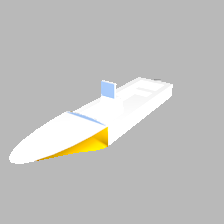}}
	\subfigure[]{
		\includegraphics[width=0.1\linewidth]{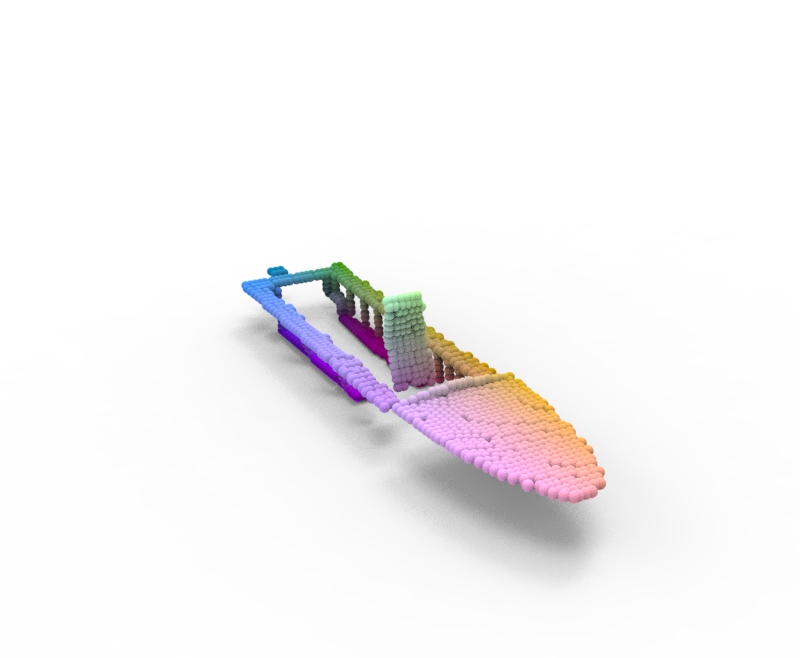}}
	\subfigure[]{
		\includegraphics[width=0.1\linewidth]{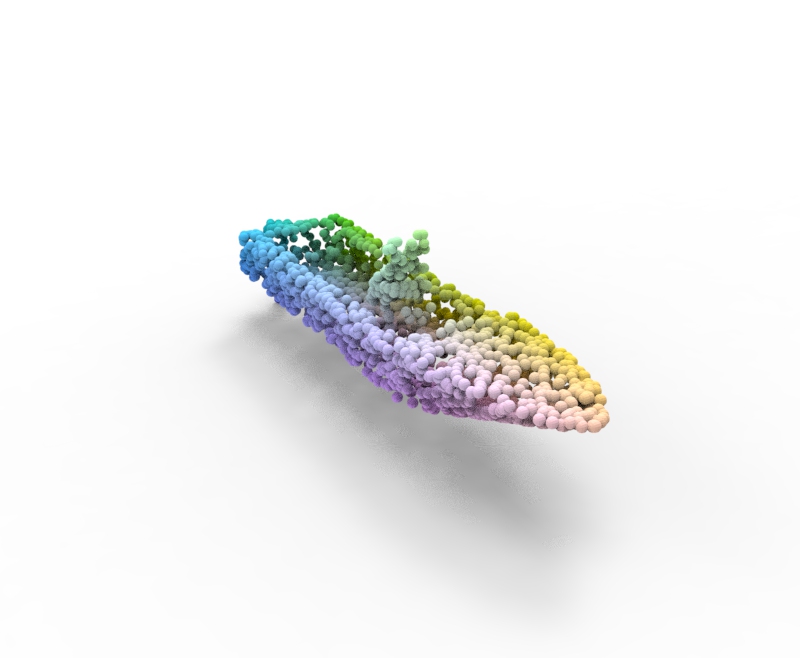}}
	\subfigure[]{
		\includegraphics[width=0.1\linewidth]{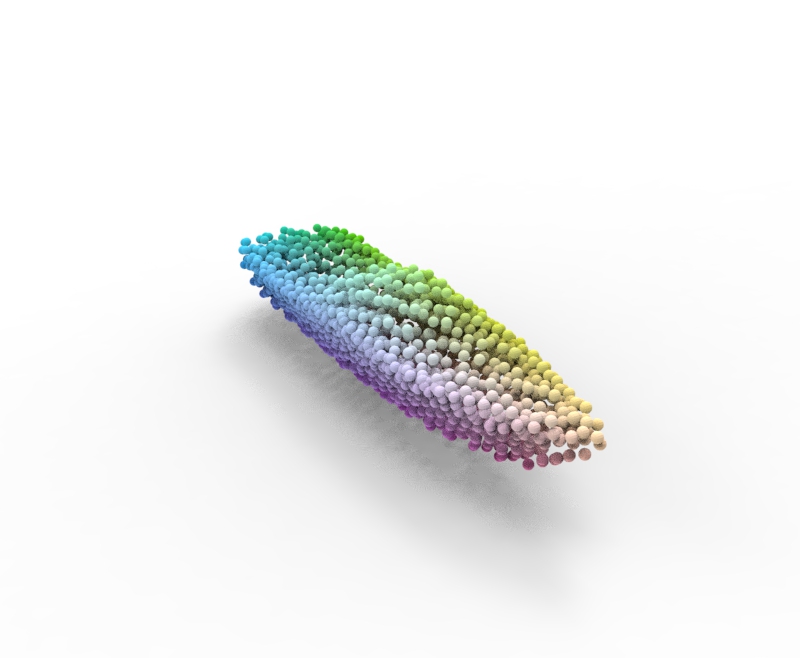}}
	\subfigure[]{
		\includegraphics[width=0.1\linewidth]{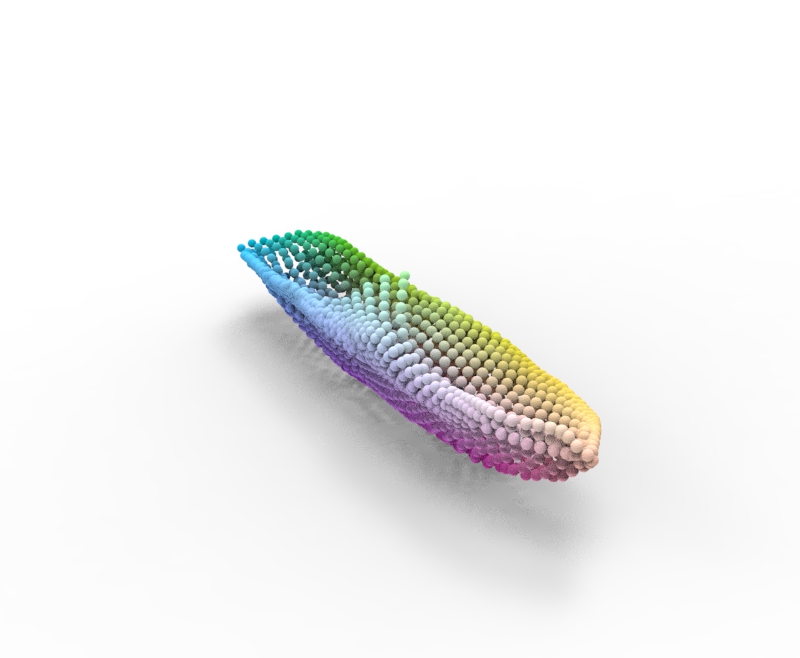}}
	\subfigure[]{
		\includegraphics[width=0.1\linewidth]{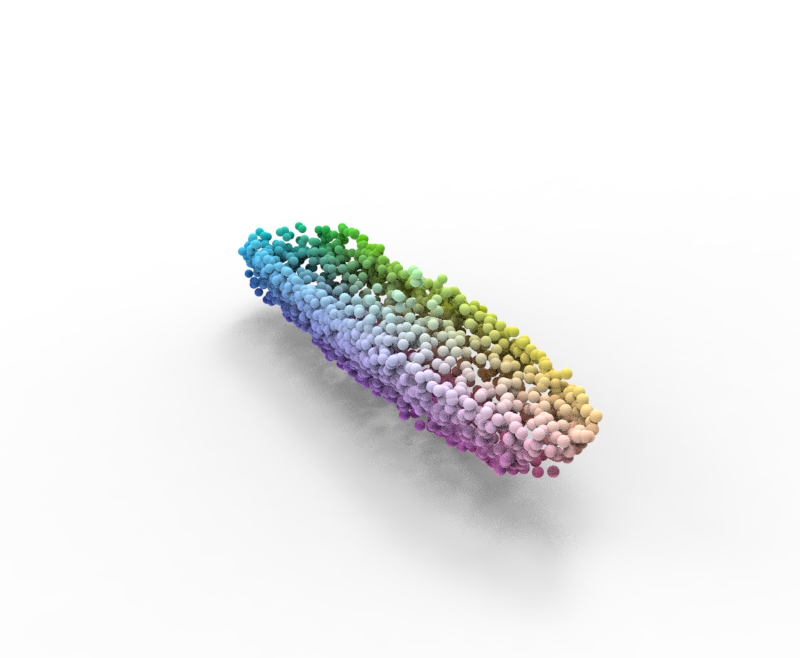}}
	\subfigure[]{
		\includegraphics[width=0.1\linewidth]{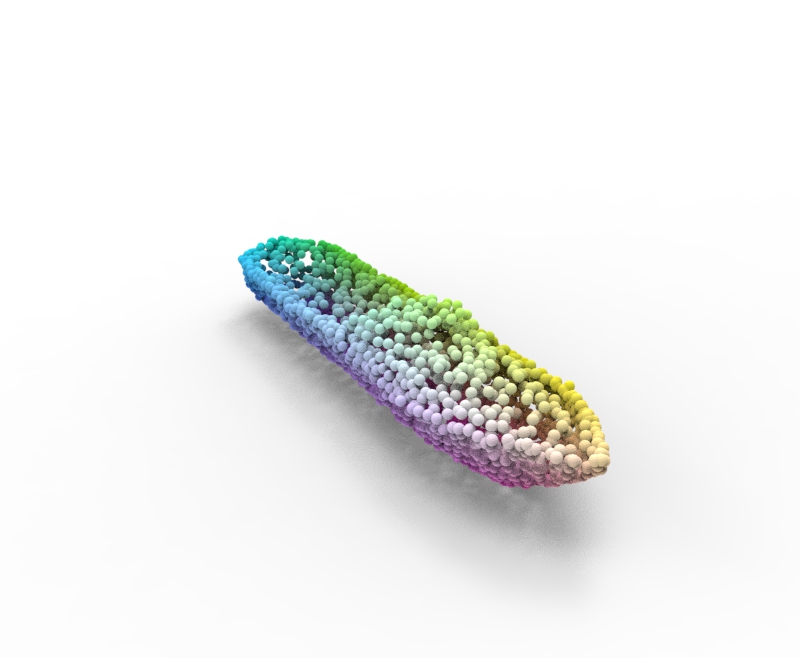}}
	\subfigure[]{
		\includegraphics[width=0.1\linewidth]{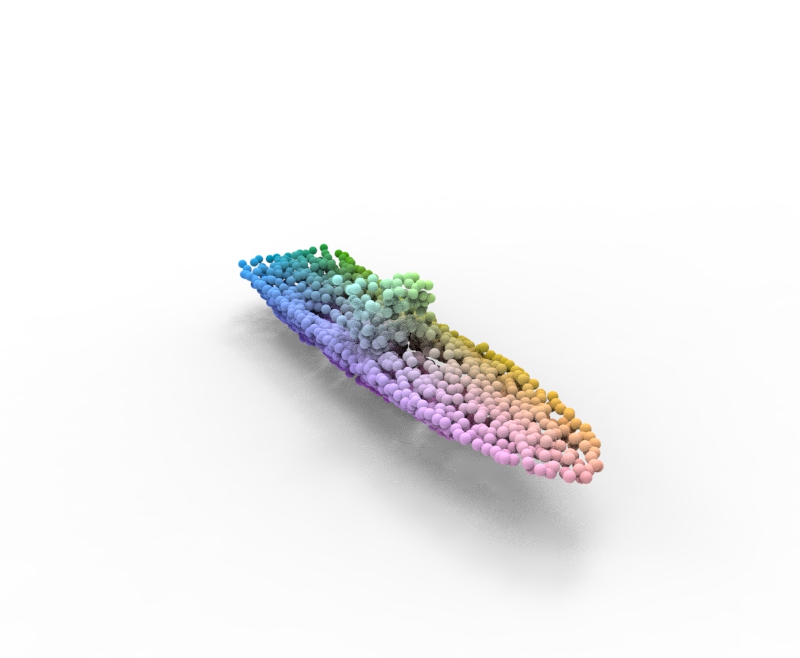}}
	\subfigure[]{
		\includegraphics[width=0.1\linewidth]{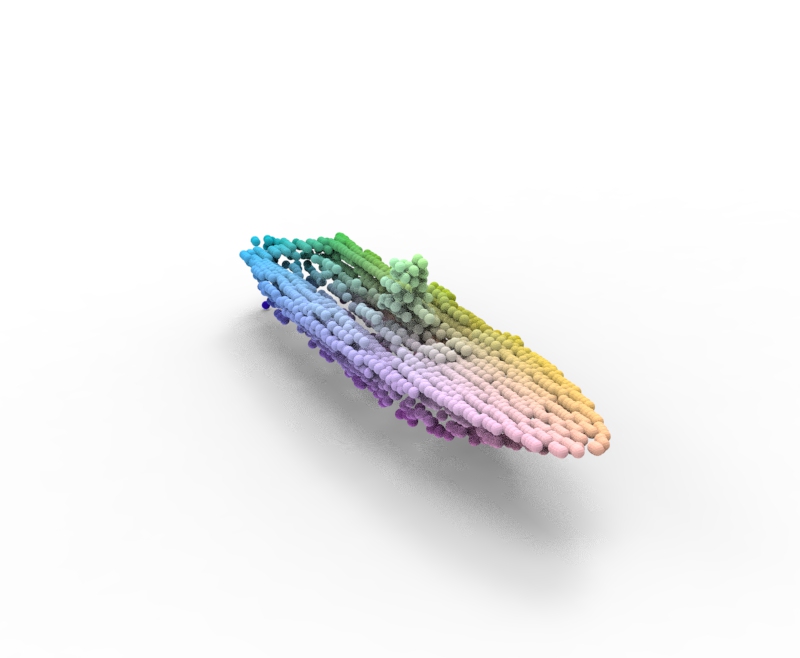}}
	\caption{Comparison of the visual results by different methods on the synthetic dataset. Note that we change the background color of the view images in our dataset from black to grey for better visualization. (a) Input image (b) Partial points. (c) Ground truth. (d) PCN \cite{12}. (e) FoldingNet \cite{11}. (f) TopNet \cite{17}. (g) Cycle \cite{2}. (h) Inversion \cite{3}. (i) Ours}
	\label{visual_comparison}
\end{figure*}

\subsection{Implementation Details}
The point numbers of $\mathbf{P}_{in}$, $\mathbf{P}_0$, $\mathbf{P}_1$, $\mathbf{P}_2$, $\mathbf{P}_c$ and $\mathbf{P}_{out}$ are 2048, 256, 512, 512, 2048 and 2048, respectively. The upsampling rate $r$ is 4 and the number of the sampled 2D points $M$ is 4096.
We used the Adam optimizer to train our model with a batch size of 32. The initial learning rate is $1e-4$, and it exponentially decays by 0.7 for every 10 epochs in the CSR stage and by 0.1 for every 10 epochs in the VSR stage. It takes 200 and 50 epochs to converge for the CSR stage and VSR stage, respectively.
For a fair comparison, we train a separate model for each category following previous unsupervised methods.
The training process is conducted on an NVIDIA RTX 3090 GPU with Intel(R) Xeon(R) CPUs.
\subsection{Comparison with State-of-the-Art Methods}

\begin{table*}[htbp]
	\centering
	\renewcommand\arraystretch{1.25}
	\caption{Results on KITTI dataset in terms of MMD\(\downarrow\) (scaled by \(10^{4}\)). ``VSR" denotes the View-assisted Shape Refinement stage. The \textbf{bold} numbers represent the best results of the unsupervised methods.}
	\label{kitti_results}
	\begin{tabular}{l|c|c|c|c|c|c|c}
		\toprule[1.2pt]
		Methods             & FoldingNet \cite{11} & PCN \cite{12} & TopNet \cite{17} & Cycle \cite{2} & Inversion \cite{3} & Cross-PCC w/o VSR & Cross-PCC \\ \hline 
		Supervised          & yes & yes & yes & no & no & no & no \\ \hline
		MMD          & 6.9	& 6.7 & 7.3 & 9.0 & 10.0 & 10.8 & \textbf{8.5}	 \\ \bottomrule[1.2pt]
	\end{tabular}
\end{table*}

\begin{table*}[htbp]
	\centering
    \renewcommand\arraystretch{1.25}
	\caption{Comparison with cross-modal point cloud completion method. The reported values are  L2 CDs\(\downarrow\) (scaled by \(10^{4}\)) on 3D-EPN benchmark.  Note that the reported values to the left and right of the slash ``/" are the ${\rm CD}_{min}$ and ${\rm CD}_{avg}$, respectively.}
	\label{cross_modal}
	\begin{tabular}{l|c|c|c|c|c|c|c|c|c|c}
		\toprule[1.2pt]
		Methods  & Supervised&Average&Plane&Cabinet&Car&Chair&Lamp&Couch&Table&Watercraft \\ \hline  
		CSDN \cite{zhu2023csdn}  & yes &  6.3/7.2  &  2.5/2.6   & 7.7/9.1  &  6.0/6.1     &8.0/9.4  &  6.8/7.3   & 8.7/10.2   & 6.9/8.4 & 4.2/4.9 \\ \hline
		Cross-PCC (ours)          & no	& 8.4/9.1 & 2.3/2.4 & 13.0/14.0 &7.4/7.6 & 9.8/10.6 & 7.8/9.3 & 11.5/12.1 & 9.6/10.7	& 6.1/6.8	\\ \bottomrule[1.2pt]
	\end{tabular}
\end{table*}

\subsubsection{Results on synthetic data}
\label{section_comparison}

We first conduct experiments on two widely-used synthetic benchmarks proposed in 3D-EPN \cite{8} and PCN \cite{12}. Table \ref{epn_results} shows the quantitative results of our method and other state-of-the-art methods, including supervised methods requiring ground-truth complete 3D point clouds as supervision (i.e., 3D-EPN \cite{8}, FoldingNet \cite{11}, PCN \cite{12}, TopNet \cite{17}, SnowflakeNet \cite{18})  and unsupervised methods (i.e., PCL2PCL \cite{1}, Cycle \cite{2}, Inversion \cite{3}). From Table \ref{epn_results}, it can be seen that our method surpasses the previous unsupervised methods by a large margin. Specifically, compared with the previous best unsupervised method Cycle, the results of our method are reduced by 4.6/3.9 in terms of the average ${\rm CD}_{min}$ and ${\rm CD}_{avg}$ among all categories. Notably, our results under the ``lamp" and ``table" categories surpass the best unsupervised results by 15.2/13.7 and 11.3/10.2, respectively. 
Besides, compared to the supervised methods, although our method still falls behind the latest state-of-the-art method SnowflakeNet \cite{18} by a relatively large margin, we achieve results that are extremely close to those of some supervised methods, i.e., FoldingNet, PCN, TopNet. The average CD gap between our and these methods is less than 0.8/1.5. Moreover, our method even outperforms these methods under some categories. For example, on the lamp category, the ${\rm CD}_{min}$/${\rm CD}_{avg}$ of our method is lower than PCN by 5.2/3.7.

The quantitative results on the PCN benchmark are shown in Table \ref{pcn_results}. Compared to the best unsupervised method, our method improves the completion performance by 4.9/4.2 in terms of the ${\rm CD}_{min}$ and ${\rm CD}_{avg}$.
Notably, our method reduces the CD on the ``Lamp" and ``Table" categories by 7.5/6.3 and 8.1/7.0, respectively.
In addition, the results of our method are very close to those of some supervised methods with a minimum gap of 1.5/2.2 in terms of the averaged CD. On the ``Lamp" and ``Table" categories, the CD results of our method exceed the FoldingNet, PCN, and TopNet by 4.8/3.6, 2.5/1.3 and 4.3/3.1, respectively.

Figure \ref{visual_comparison} shows the qualitative results of our method and the state-of-the-art methods on the synthetic dataset. From Figure \ref{visual_comparison}, we can see that our method can effectively recover the missing part while preserving the partial input. Both the supervised and unsupervised methods are apt to alter the partial input and thus generate a shape with a large difference from the input and ground truth. Our method, however, preserves the input shape completely. In addition, the supervised and unsupervised methods do poor in recovering some details, such as holes and thin bars. For example, these methods fill the hollows in the backs of the chair and couch in Figure \ref{visual_comparison}, which is inconsistent with the ground truth. In contrast, our method accurately predicts the bars and holes.

\begin{figure}[htbp]
	\centering  
	\subfigbottomskip=1pt 
	\subfigure{
		\includegraphics[width=0.15\linewidth]{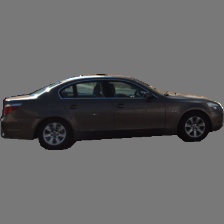}}
	\hspace{-3mm}
	\subfigure{
		\includegraphics[width=0.2\linewidth]{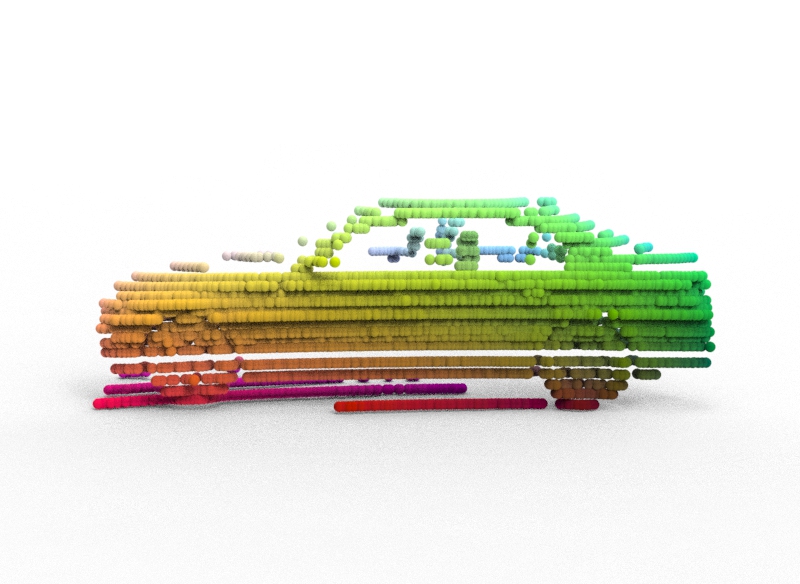}}
	\hspace{-3mm}
	\subfigure{
		\includegraphics[width=0.2\linewidth]{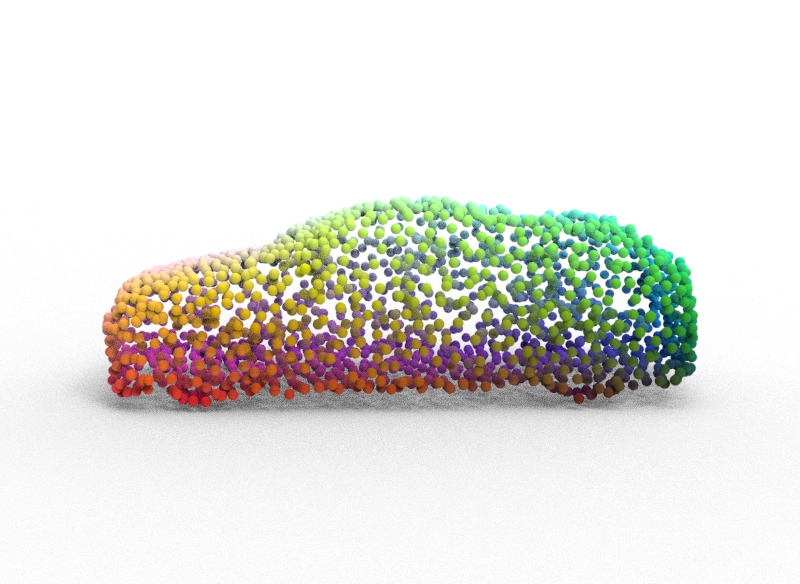}}
	\hspace{-3mm}
	\subfigure{
		\includegraphics[width=0.2\linewidth]{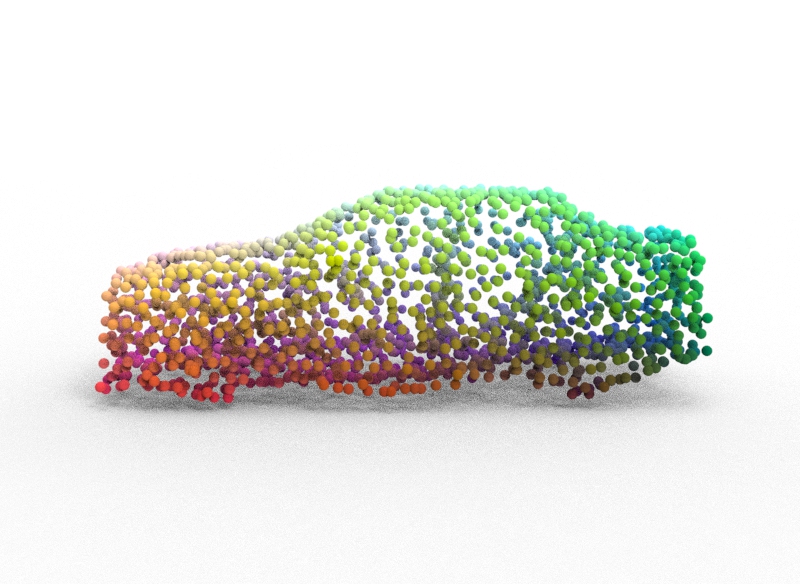}}
	\hspace{-3mm}
	\subfigure{
		\includegraphics[width=0.2\linewidth]{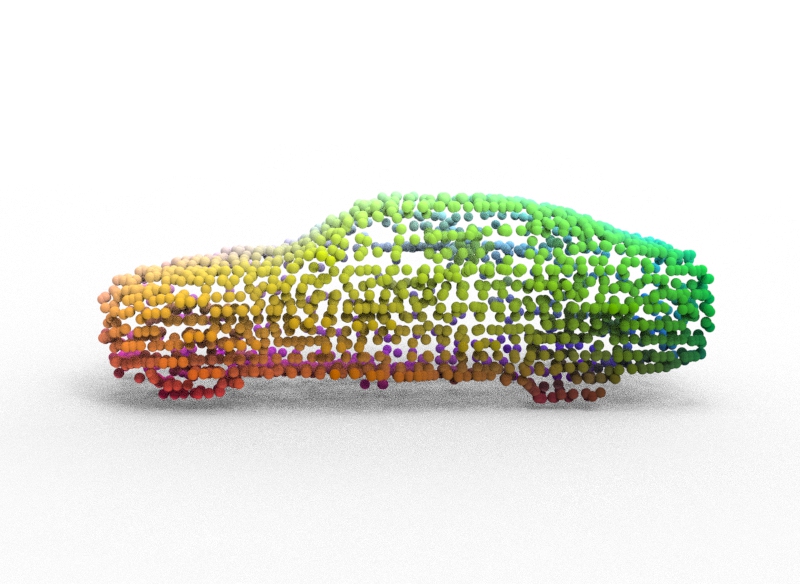}}
	\\
	\subfigure{
		\includegraphics[width=0.15\linewidth]{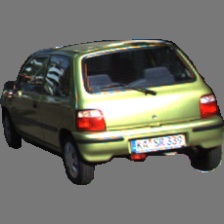}}
	\hspace{-3mm}
	\subfigure{
		\includegraphics[width=0.2\linewidth]{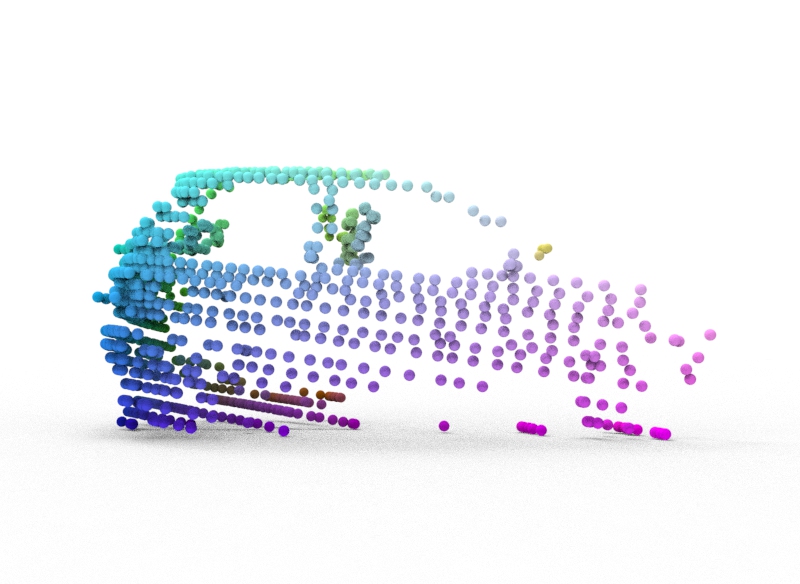}}
	\hspace{-3mm}
	\subfigure{
		\includegraphics[width=0.2\linewidth]{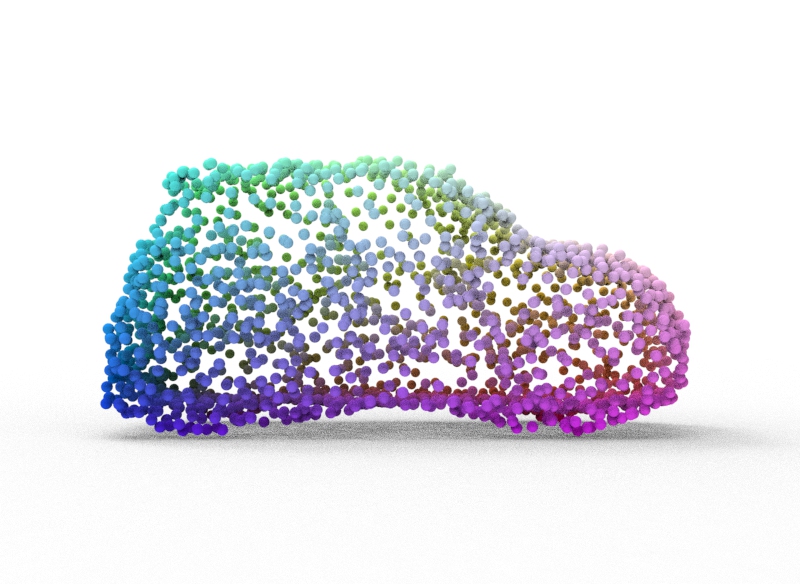}}
	\hspace{-3mm}
	\subfigure{
		\includegraphics[width=0.2\linewidth]{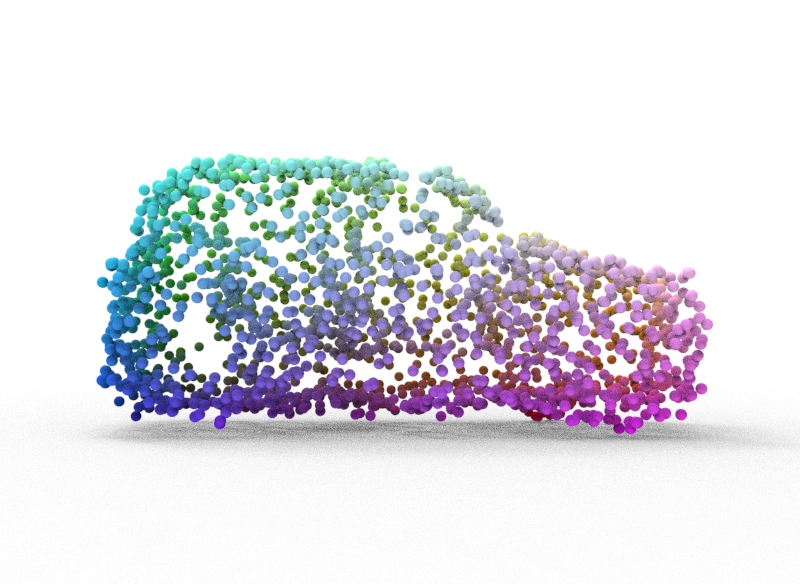}}
	\hspace{-3mm}
	\subfigure{
		\includegraphics[width=0.2\linewidth]{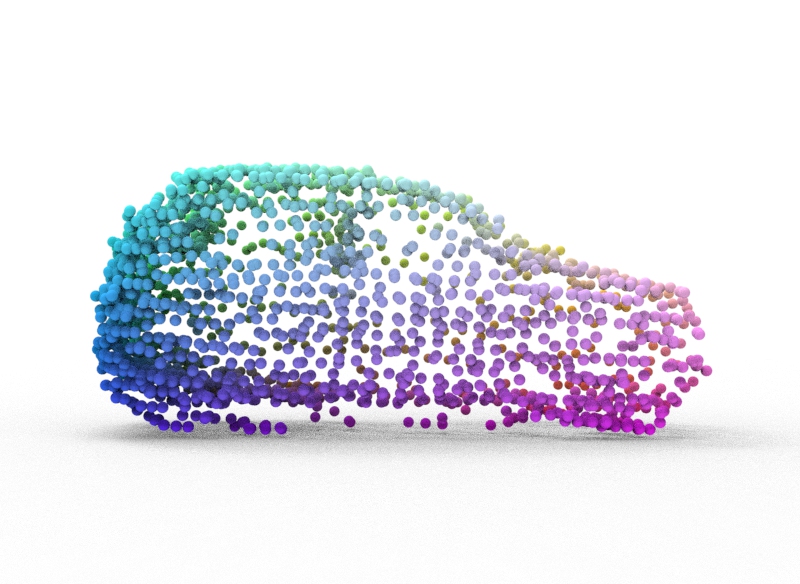}}
	\\
	\setcounter{subfigure}{0}
	\subfigure[]{
		\includegraphics[width=0.15\linewidth]{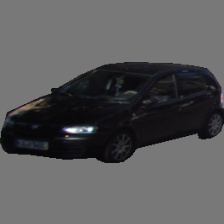}}
	\hspace{-3mm}
	\subfigure[]{
		\includegraphics[width=0.2\linewidth]{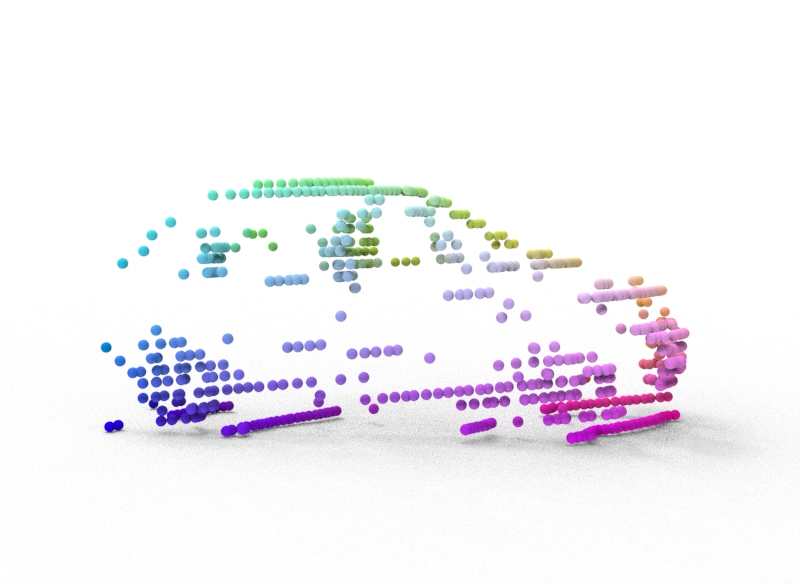}}
	\hspace{-3mm}
	\subfigure[]{
		\includegraphics[width=0.2\linewidth]{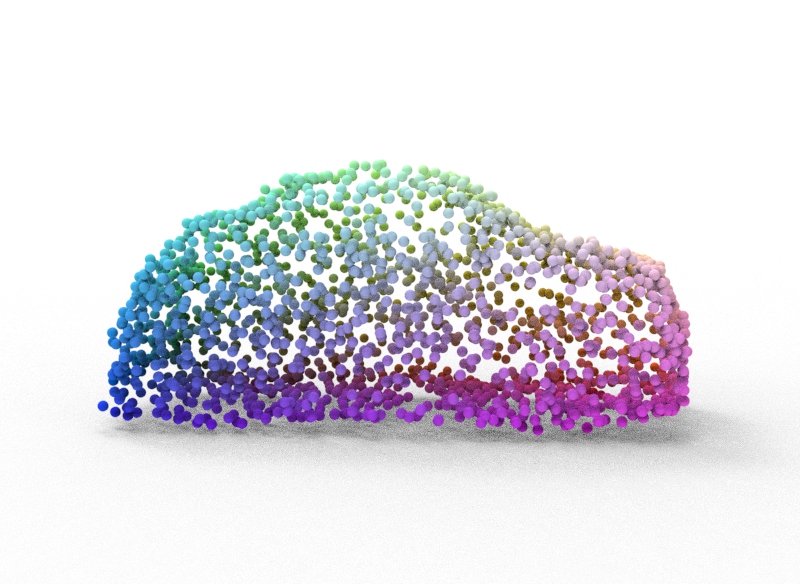}}
	\hspace{-3mm}
	\subfigure[]{
		\includegraphics[width=0.2\linewidth]{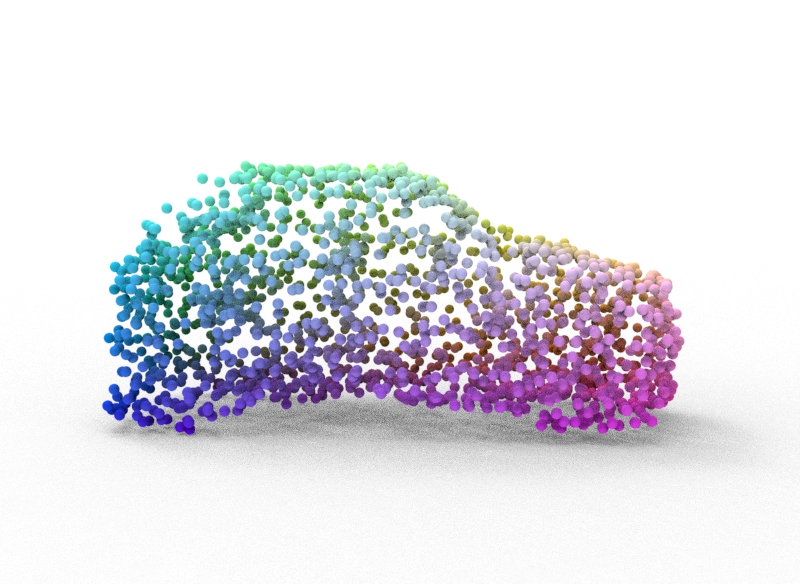}}
	\hspace{-3mm}
	\subfigure[]{
		\includegraphics[width=0.2\linewidth]{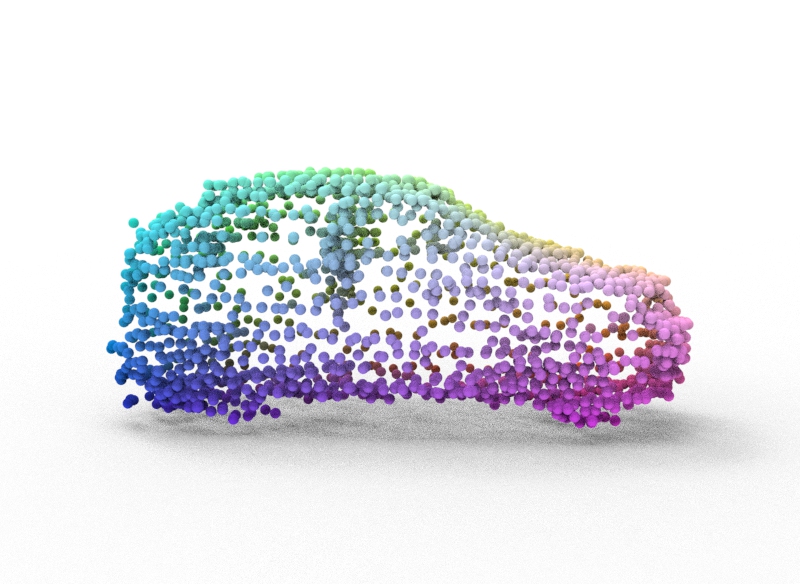}}

	\caption{Comparison of visual results between our methods and unsupervised methods on the KITTI dataset. (a) Input image (b) Input partial points. (c) Cycle \cite{2}. (d) Inversion \cite{3}. (e) Ours.}
	\label{kitti_visual}
\end{figure}

\subsubsection{Results on real-world scans}
Our completion model can be extended to real-world scans. We test our models on KITTI data which are extracted from real-world scenes.
Table \ref{kitti_results} shows the quantitative results of our model and other state-of-the-art methods, including supervised and unsupervised methods. 
Compared with the latest unsupervised methods, our method yields the best result in terms of MMD. And our method is comparable with some supervised methods with a little behind.
These results demonstrate that our model is capable of predicting reasonable complete cars according to the sparse and partial point clouds.
Figure \ref{kitti_visual} shows some visual results of our methods and two unsupervised methods, i.e., Cycle \cite{2} and Inversion \cite{3}. We can find that our model can predict reasonable shapes while doing better in preserving the partial input compared with the unsupervised methods.

\subsubsection{Comparison with cross-modal point cloud completion methods}
In addition to comparing our approach with unimodal methods, we also conduct a comparative analysis with the \textit{supervised} cross-modal point cloud completion method named CSDN. Note that CSDN was trained using 24-view images for each object in the original work. In contrast, only 8-view images per object were used for training our framework and CSDN in this work.
As shown in Table \ref{cross_modal}, our method falls behind CSDN in terms of results across seven categories. However, the performance gap is generally less than 2 for most categories, and our method slightly outperforms CSDN in the plane category. These findings indicate that our method remains competitive with supervised cross-modal point cloud completion methods, as the performance disparity is not substantial.

\begin{table}[thbp]
	\centering
    \renewcommand\arraystretch{1.25}
	\caption{ Quantitative comparison with single view reconstruction methods on 3D-EPN benchmark. Note that the reported values represent average CD across all categories. The \textbf{bold} numbers represent the best results across all methods.}
	\label{svr_comparison}
	\begin{tabular}{l|c|c|c}
		\toprule[1.2pt]
		Methods             & Supervised & ${\rm CD}_{min}$ & ${\rm CD}_{avg}$ \\ \hline  
		PSGN \cite{fan2017point}      & yes & 16.3 & 23.2  \\ 
		2DPM \cite{20}    & no & 27.7  & 33.8 \\ \hline
		Cross-PCC (ours)          & no	& \textbf{8.4} & \textbf{9.1} \\ \bottomrule[1.2pt]
	\end{tabular}
\end{table}

\begin{table}[htbp]
	\centering
    \renewcommand\arraystretch{1.25}
	\caption{ Comparison of FLOPs and model size.}
	\label{flops_comparison}
	\begin{tabular}{l|c|c|c|c}
		\toprule[1.2pt]
		Methods             & SnowflakeNet & Inversion & CSDN  &Ours \\ \hline  
		FLOPs (G)      &  7.38  & 1.35 & 14.21 & 11.51 \\   
		Model size (M)    & 73.60 & 156.41 & 67.53 & 98.90 \\ \bottomrule[1.2pt]
	\end{tabular}
\end{table}

\begin{table*}[htbp]
	\centering
	\renewcommand\arraystretch{1.25}
	\caption{Results comparison of our image feature branch, view calibrator and offset predictor. ``\Checkmark" and ``\XSolid" represent the model with or without the component, respectively. 'IFB', 'VC' and 'OP' represent the image feature branch, view calibrator and offset predictor, respectively. The results are reported on the 3D-EPN dataset.}
	\label{component}
	\begin{tabular}{c|c|c|c|c|c|c|c|c|c|c|c|c}
		\toprule[1.2pt]
		\multicolumn{4}{c|}{Component} & \multicolumn{9}{c}{Datasets}                   \\ \hline
		IFB         & 1st VC	& OP	& 2nd VC     & Average & Plane & Cabinet & Car & Chair & Lamp	& Couch	& Table	& Watercraft \\ \hline
		\XSolid    & \XSolid	& \XSolid	& \XSolid    & 13.3  & 3.5 & 19.5 & 11.7 & 15.0 & 14.0 & 16.6	& 17.3	& 9.3 \\
		\Checkmark    & \XSolid	& \XSolid	& \XSolid    & 11.4/11.9  & 3.3/3.3 & 16.1/17.3 & 11.4/11.5 & 12.5/12.8 & 10.6/11.9 & 14.8/15.1	& 14.8/15.3	& 7.8/8.5 \\
		\Checkmark    & \Checkmark	& \XSolid	& \XSolid    & 10.5/11.3  & 2.8/3 & 15.9/17.1 & 10.7/11.0 & 11.5/12.2 & 9.1/10.7 & 14.2/14.7	& 13.2/14.3	& 7.3/8.1 \\
		\Checkmark    & \Checkmark	& \Checkmark	& \XSolid & 8.5/9.2  & 2.3/2.5 & 13.1/14.1 & 7.4/7.7 & 10.0/10.7 & 7.9/9.4 & 11.7/12.2	& 9.8/10.8	& 6.1/6.9 \\
		\Checkmark    & \Checkmark	& \Checkmark	& \Checkmark & \textbf{8.4/9.1} & \textbf{2.3/2.4} & \textbf{13.0/14.0} & \textbf{7.4/7.6} & \textbf{9.8/10.6} & \textbf{7.8/9.3} & \textbf{11.5/12.1} & \textbf{9.6/10.7}	& \textbf{6.1/6.8} \\ \bottomrule[1.2pt]
	\end{tabular}
\end{table*}

\subsubsection{Comparison with single-view image reconstruction methods}
The single-view image reconstruction methods reconstruct a complete 3D shape from a single 2D image. The supervised methods drive the training process by measuring the similarity between predicted results and corresponding complete 3D shapes, whereas unsupervised methods rely on 2D images along with differentiable rendering techniques to guide the learning process. In contrast to single-view image reconstruction methods, our Cross-PCC is specifically designed for point cloud completion, aiming to fill in the missing parts of the incomplete input point cloud. The images we employed serve as auxiliary information to support the point cloud completion process. To assess the performance of our method in comparison to single-view image reconstruction approaches, we conduct comparative experiments involving PSGN (supervised) and 2DPM (unsupervised). The results in Table \ref{svr_comparison} demonstrate that our method significantly outperforms PSGN and 2DPM, with ${\rm CD}_{min}$/ ${\rm CD}_{avg}$ values that are 0.9/1.5 and  2.3/2.7  times smaller than PSGN and 2DPM, respectively. In our analysis, there are two key factors contributing to this result. Firstly, our designed model exhibits greater power when compared to the relatively simpler models of PSGN and 2DPM. Secondly, the utilization of incomplete 3D point cloud inputs, combined with the incorporation of partial matching loss, facilitates improved perception and inference of complete 3D shapes.

\subsubsection{Comparison of computational complexity and model size}
In Table \ref{flops_comparison}, we compared the FLOPs (Floating Point Operations) and model sizes of our Cross-PCC and some previous models,  
where it can be seen that our model has higher FLOPs compared to SnowflakeNet and Inversion, but lower than model CSDN. This observation suggests that although our model involves the operation of projecting point clouds onto 2D planes from multiple viewpoints, its computational complexity remains moderate. This is because the projection operation mainly involves matrix computations without requiring rendering-based operations such as rasterization. Additionally, the size of our model is slightly larger than SnowflakeNet and CSDN, but significantly smaller than Inversion.

\subsection{Ablation Study}

\subsubsection{Effect of each component}
To analyze the effectiveness of three key components of our model, i.e., the image feature branch (2D encoder and modality fusion module), view calibrator and offset predictor, we conduct comprehensive ablation studies on the 3D-EPN benchmark. The experimental settings are consistent with those in Section \ref{section_comparison}.

As Table \ref{component} shows, without any components, the model yields poor results. After adding the 2D encoder, the model reduces the average CD by 1.9/1.4. It proves that our 2D encoder and modality fusion module can extract useful 2D features from images and fuse them with point cloud features effectively.

Further, after embedding our calibrator behind the generation decoder and offset predictor, we improve the results by 0.9/0.6 and 0.1/0.1, respectively. 
After the calibrated point clouds are refined by the offset predictor, the average CD decreases significantly by 2.0/2.1. It means that the offset predictor can predict more precise coordinates for calibrated points.

\begin{table}[htbp]
	\centering
    \renewcommand\arraystretch{1.25}
	\caption{ Quantitative comparison of the effects achieved by using varying numbers of view images for supervision and calibration. Note that the reported values represent average CD across all categories. The \textbf{bold} numbers represent the best results across all methods.}
	\label{view_num}
	\begin{tabular}{l|c|c|c}
		\toprule[1.2pt]
		View number for supervision             & 1 & 4 & 8 \\ \hline  
		${\rm CD}_{min}$/${\rm CD}_{avg}$       & 9.1/9.9  & 8.8/9.7 & \textbf{8.4/9.1} \\ \hline\hline
        View number for calibration           & 1       & 4   & 8 \\ \hline  
		${\rm CD}_{min}$/${\rm CD}_{avg}$ & 8.4/9.1 & 7.9/8.5 & \textbf{7.8/8.4} \\ \bottomrule[1.2pt]
	\end{tabular}
\end{table}

\begin{figure}[htbp]
	\centering  
	\subfigbottomskip=1pt 
	\subfigure{
		\includegraphics[width=0.19\linewidth]{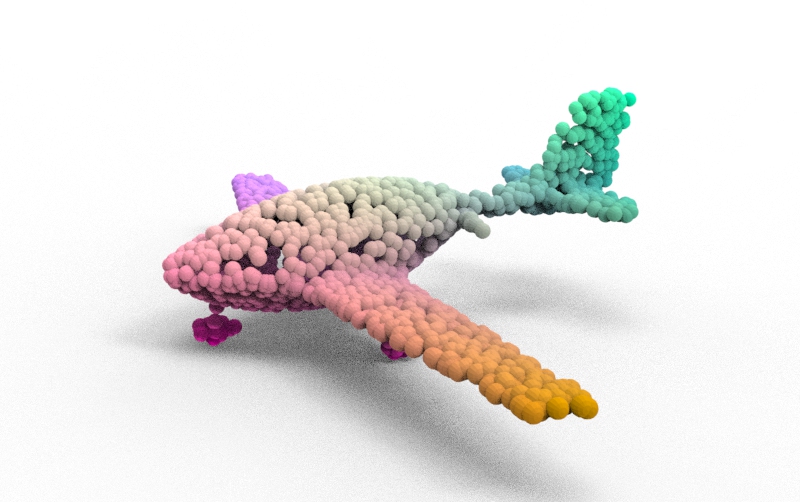}}
	\hspace{-3mm}
	\subfigure{
		\includegraphics[width=0.19\linewidth]{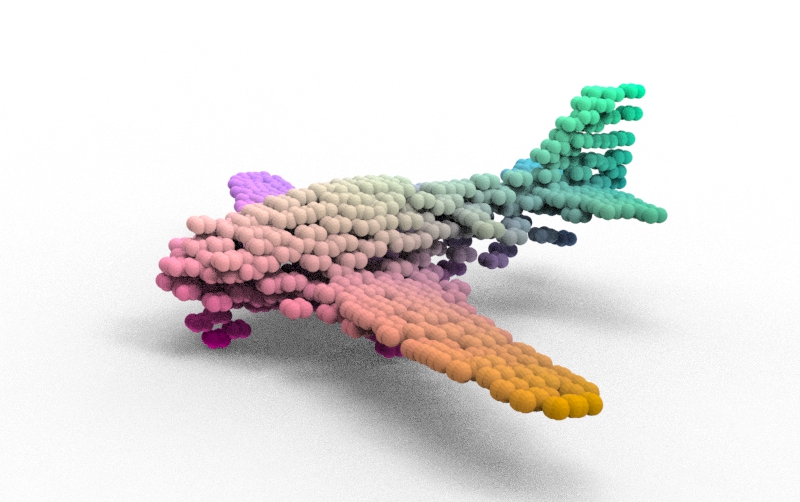}}
	\hspace{-3mm}
	\subfigure{
		\includegraphics[width=0.19\linewidth]{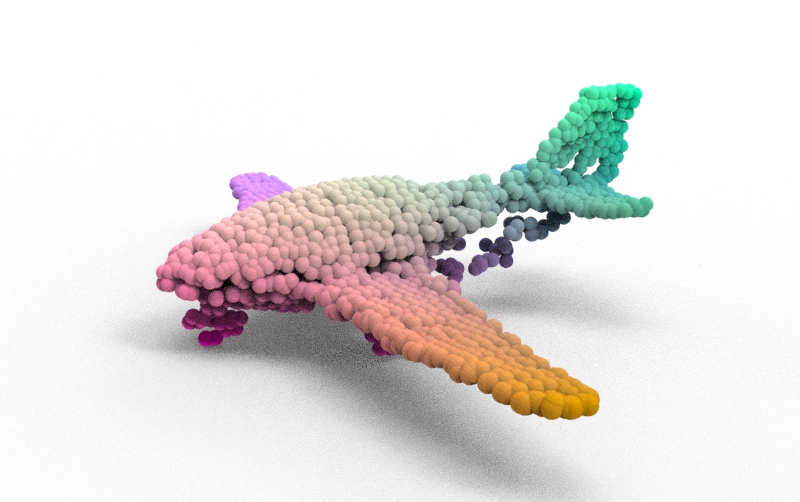}}
	\hspace{-3mm}
	\subfigure{
		\includegraphics[width=0.19\linewidth]{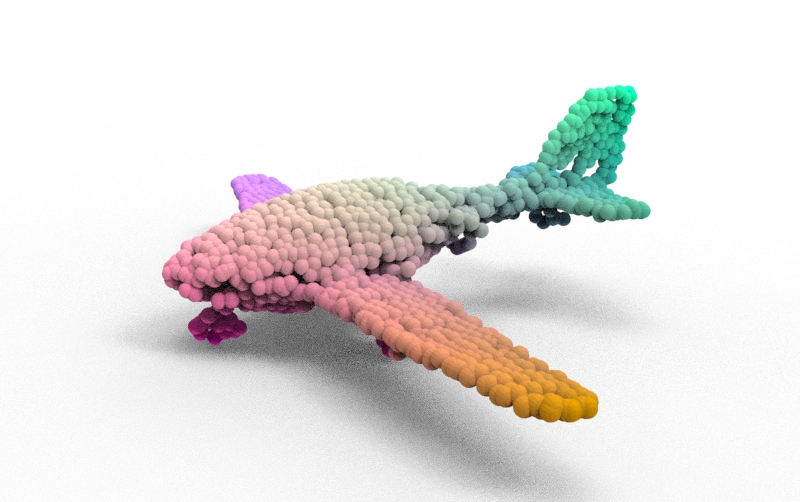}}
	\hspace{-3mm}
	\subfigure{
		\includegraphics[width=0.19\linewidth]{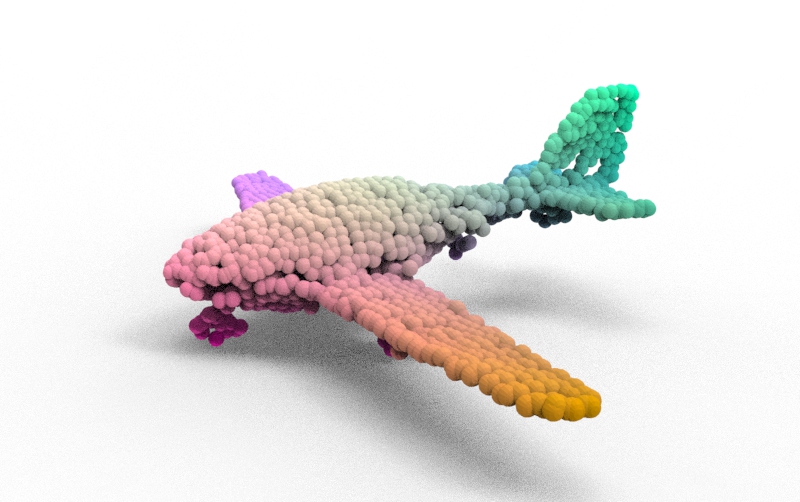}}
	\\
	\setcounter{subfigure}{0}
	\subfigure[]{
		\includegraphics[width=0.19\linewidth]{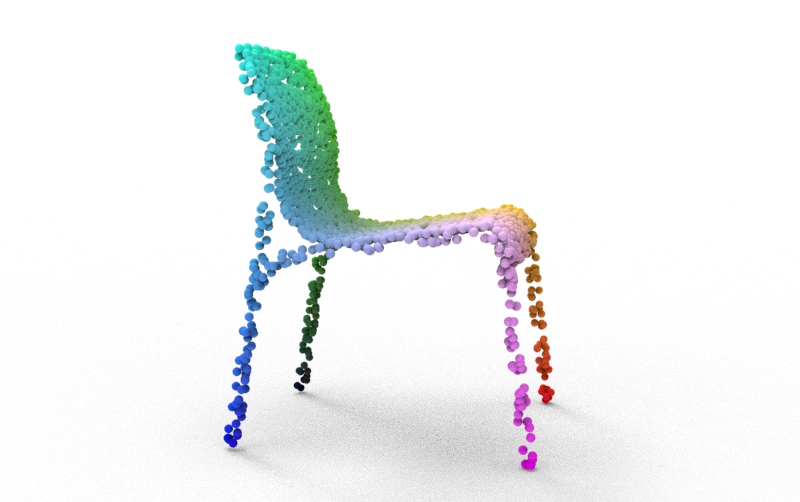}}
	\hspace{-3mm}
	\subfigure[]{
		\includegraphics[width=0.19\linewidth]{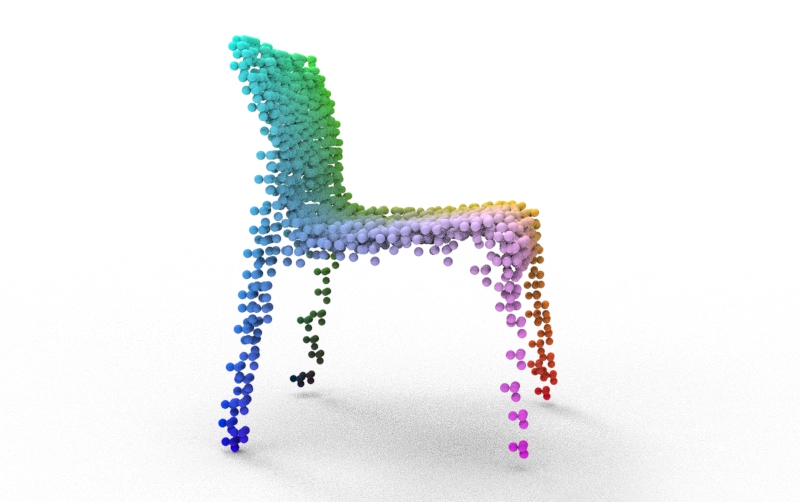}}
	\hspace{-3mm}
	\subfigure[]{
		\includegraphics[width=0.19\linewidth]{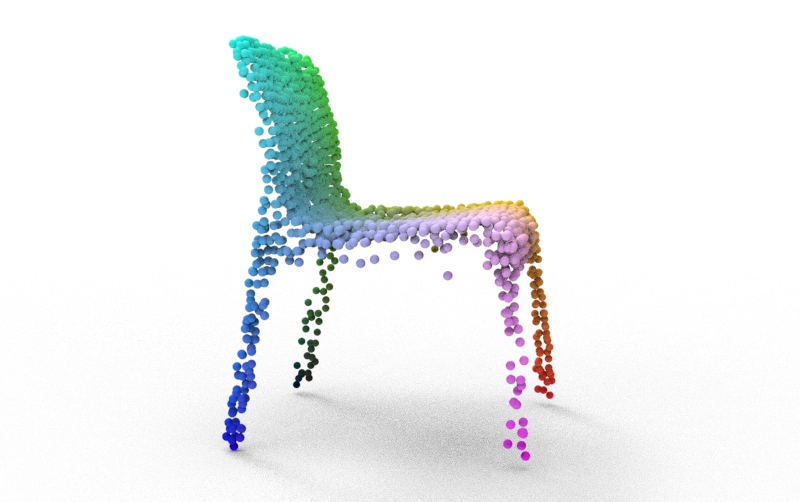}}
	\hspace{-3mm}
	\subfigure[]{
		\includegraphics[width=0.19\linewidth]{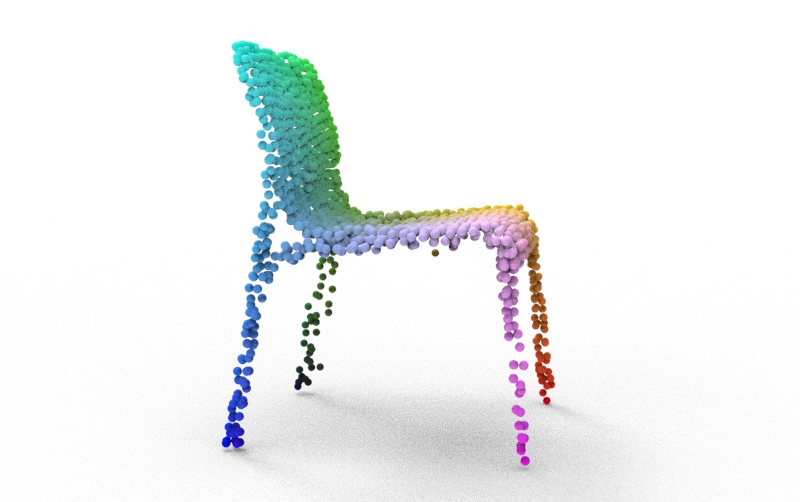}}
	\hspace{-3mm}
	\subfigure[]{
		\includegraphics[width=0.19\linewidth]{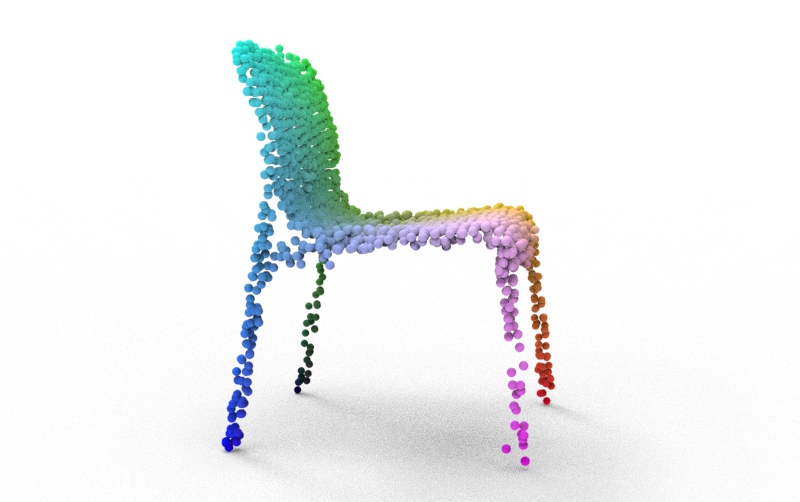}}

	\caption{Visualization of results after calibration with varying numbers of views. (a) ground truth (b) without calibration (c) calibration with (c) 1 view (d)  4 views (e)  8 views.}
	\label{cal_num_vis}
\end{figure}

\begin{figure*}[htbp]
	\centering  
	\subfigbottomskip=1pt 
	\subfigure{
		\includegraphics[width=0.09\linewidth]{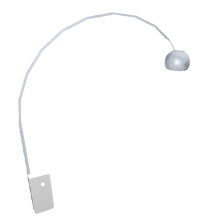}}
	\hspace{-2mm}
	\subfigure{
		\includegraphics[width=0.12\linewidth]{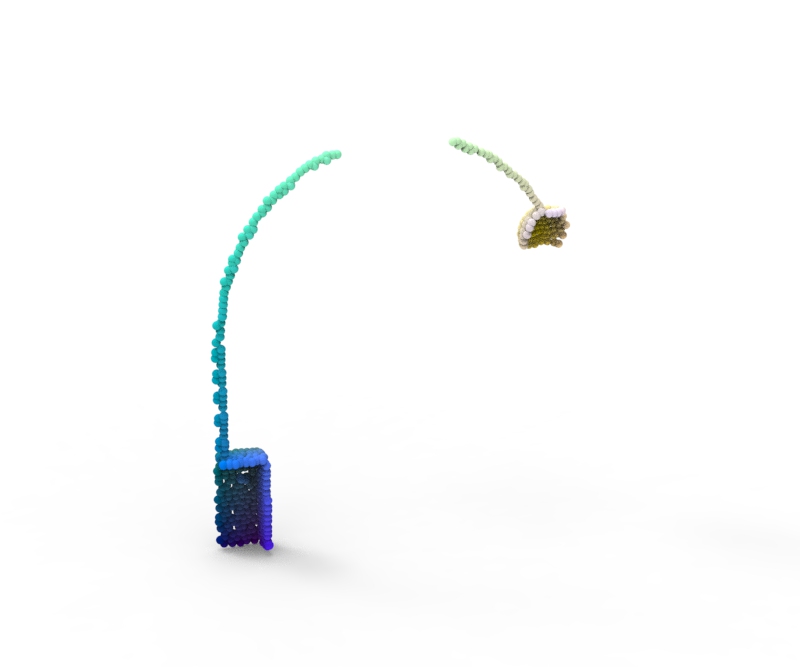}}
	\hspace{-2mm}
	\subfigure{
		\includegraphics[width=0.12\linewidth]{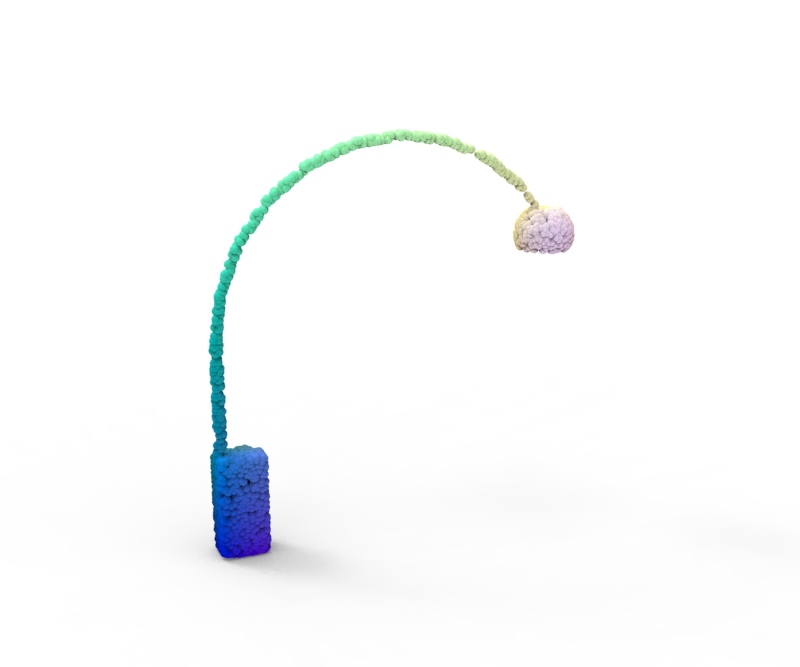}}
	\hspace{-2mm}
	\subfigure{
		\includegraphics[width=0.12\linewidth]{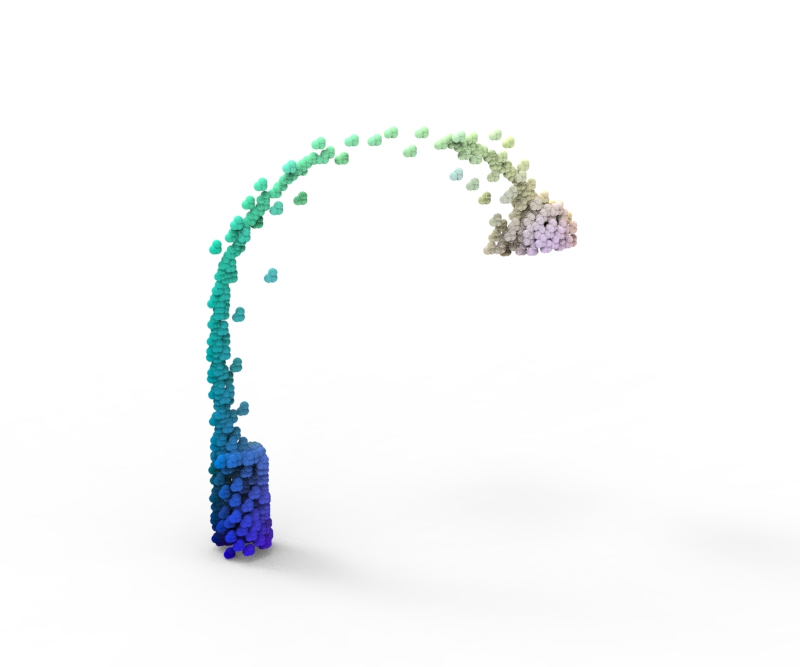}}
	\hspace{-2mm}
	\subfigure{
		\includegraphics[width=0.12\linewidth]{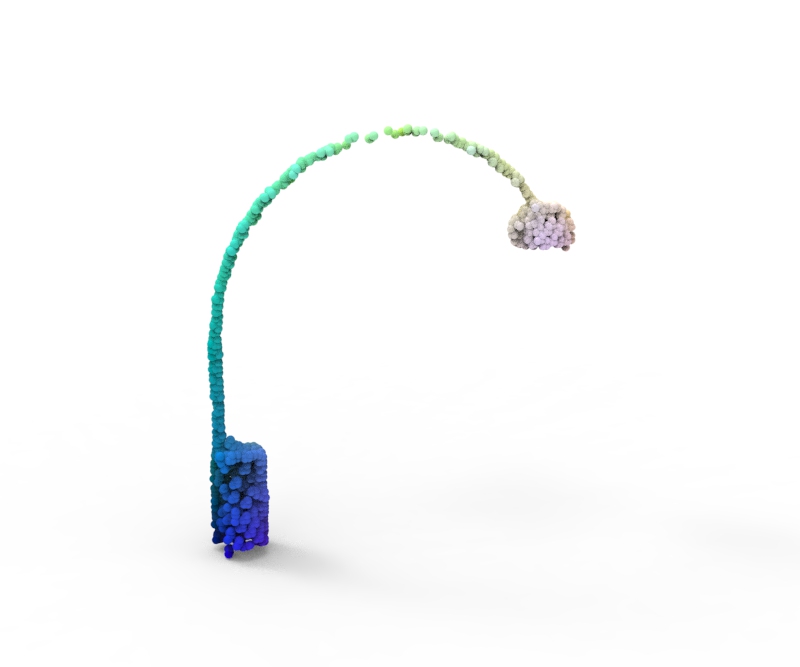}}
	\hspace{-2mm}
	\subfigure{
		\includegraphics[width=0.12\linewidth]{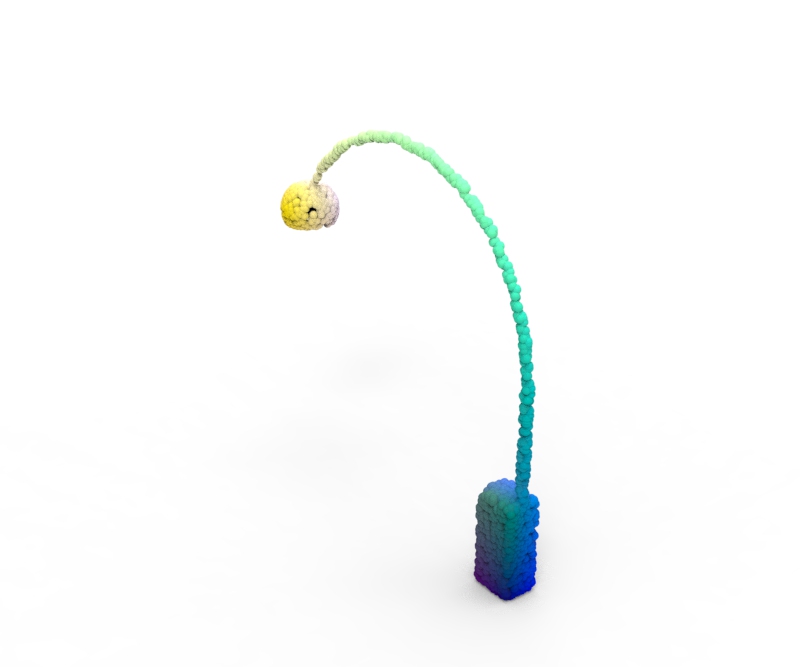}}
	\hspace{-2mm}
	\subfigure{
		\includegraphics[width=0.12\linewidth]{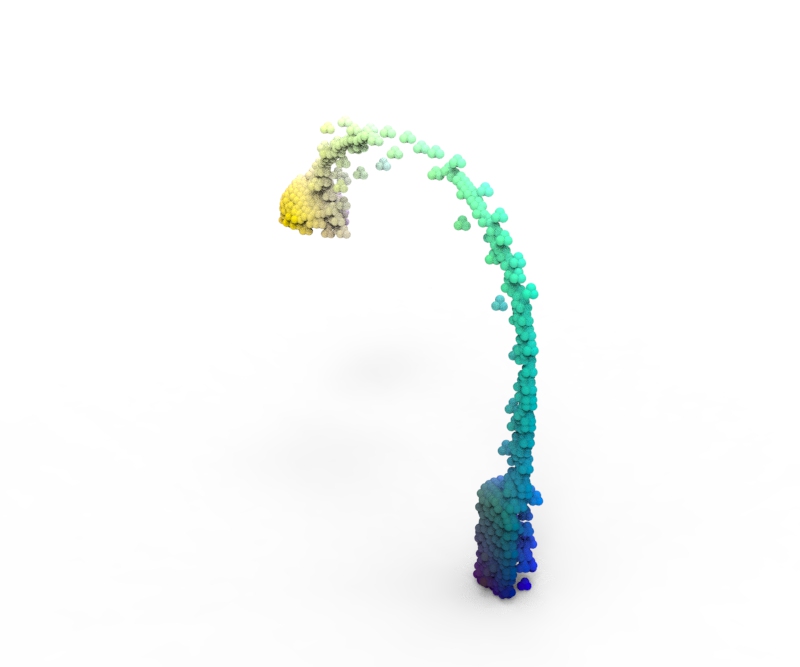}}
	\hspace{-2mm}
	\subfigure{
		\includegraphics[width=0.12\linewidth]{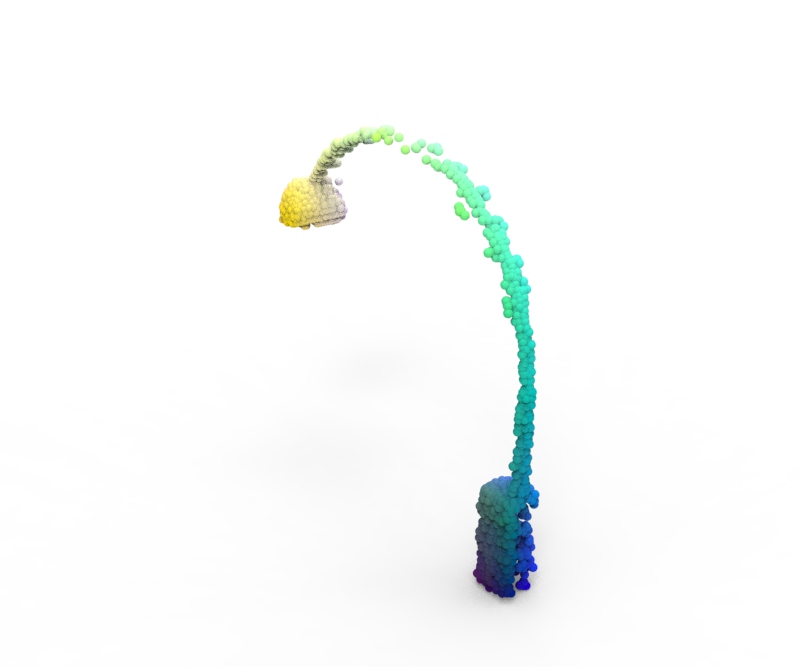}}
	\\
	\subfigure{
		\includegraphics[width=0.09\linewidth]{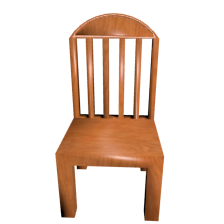}}
	\hspace{-2mm}
	\subfigure{
		\includegraphics[width=0.12\linewidth]{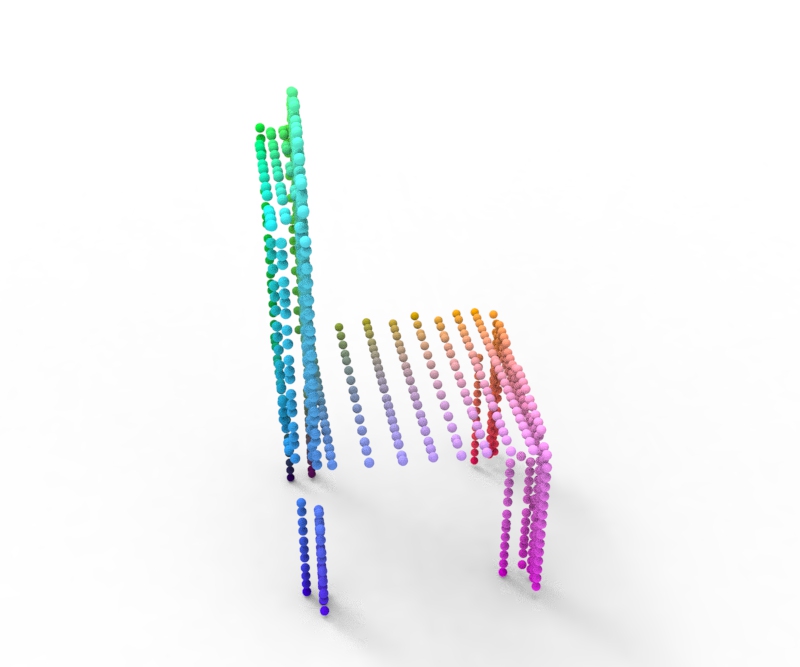}}
	\hspace{-2mm}
	\subfigure{
		\includegraphics[width=0.12\linewidth]{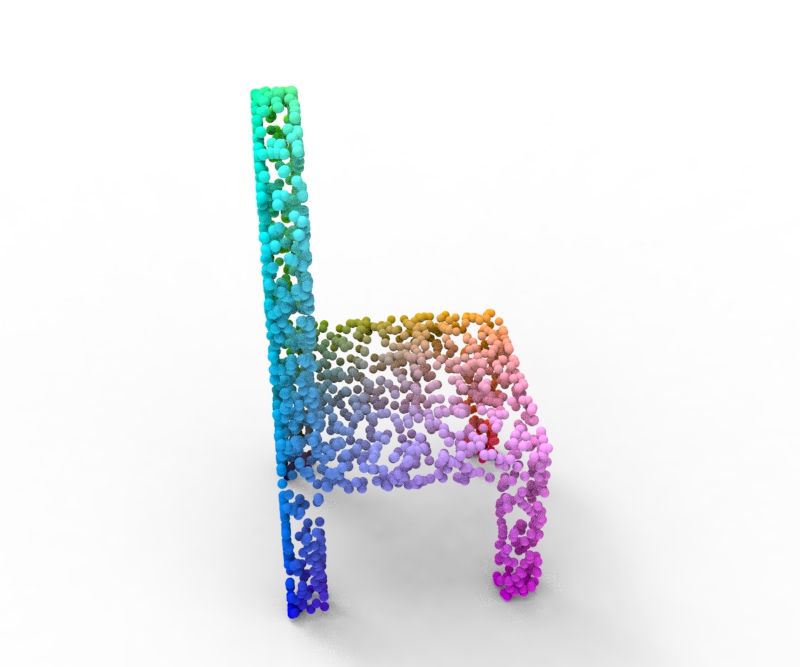}}
	\hspace{-2mm}
	\subfigure{
		\includegraphics[width=0.12\linewidth]{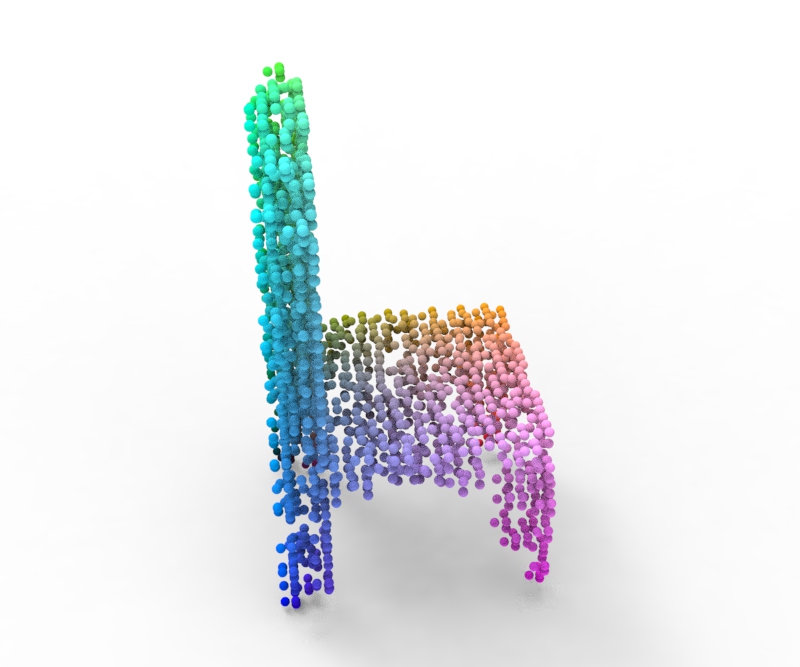}}
	\hspace{-2mm}
	\subfigure{
		\includegraphics[width=0.12\linewidth]{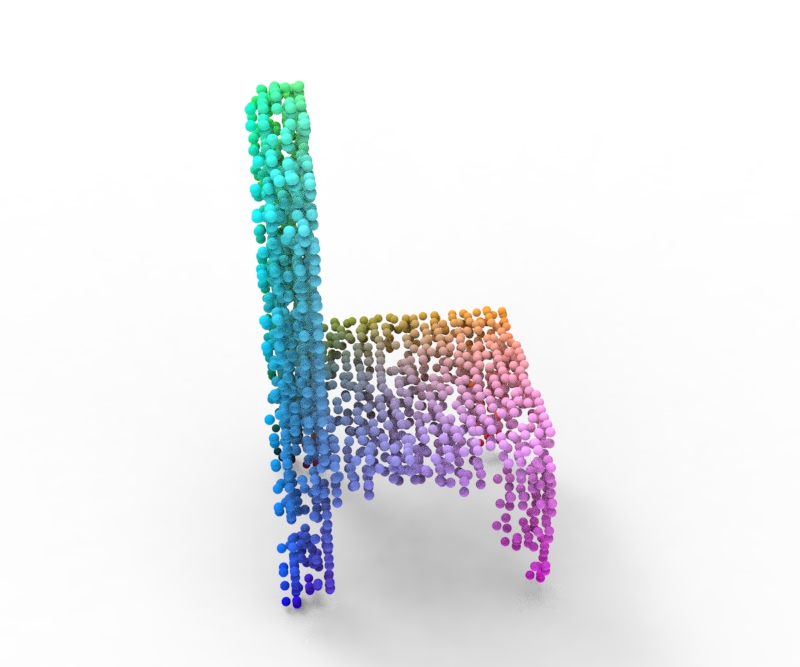}}
	\hspace{-2mm}
	\subfigure{
		\includegraphics[width=0.12\linewidth]{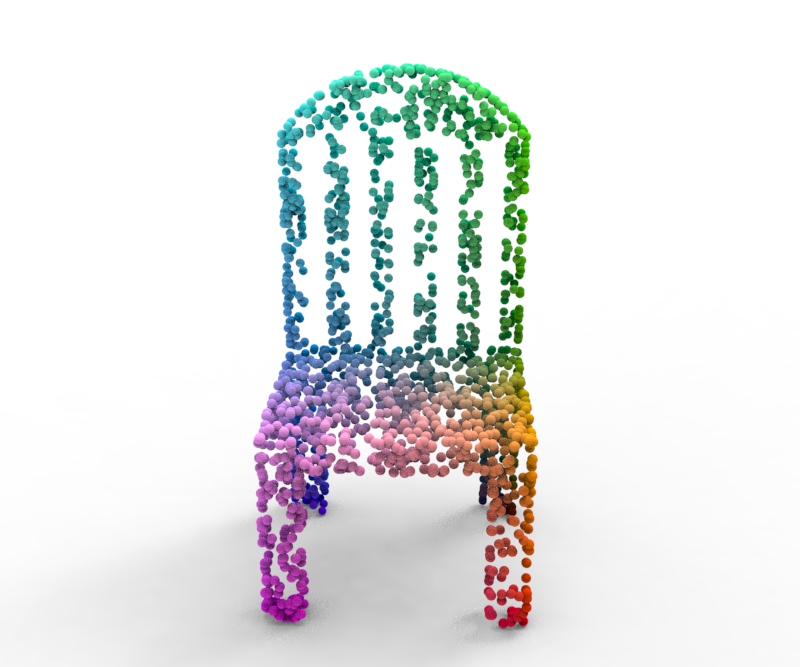}}
	\hspace{-2mm}
	\subfigure{
		\includegraphics[width=0.12\linewidth]{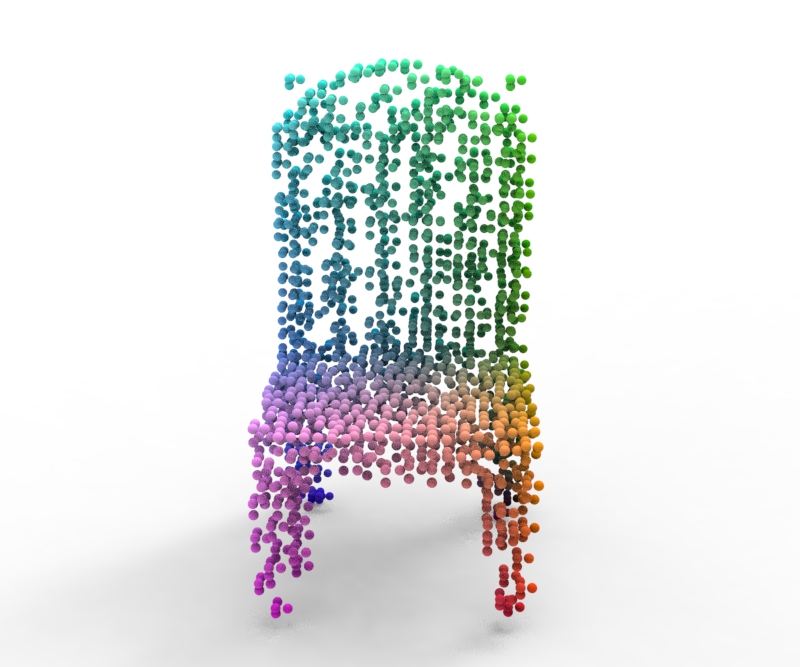}}
	\hspace{-2mm}
	\subfigure{
		\includegraphics[width=0.12\linewidth]{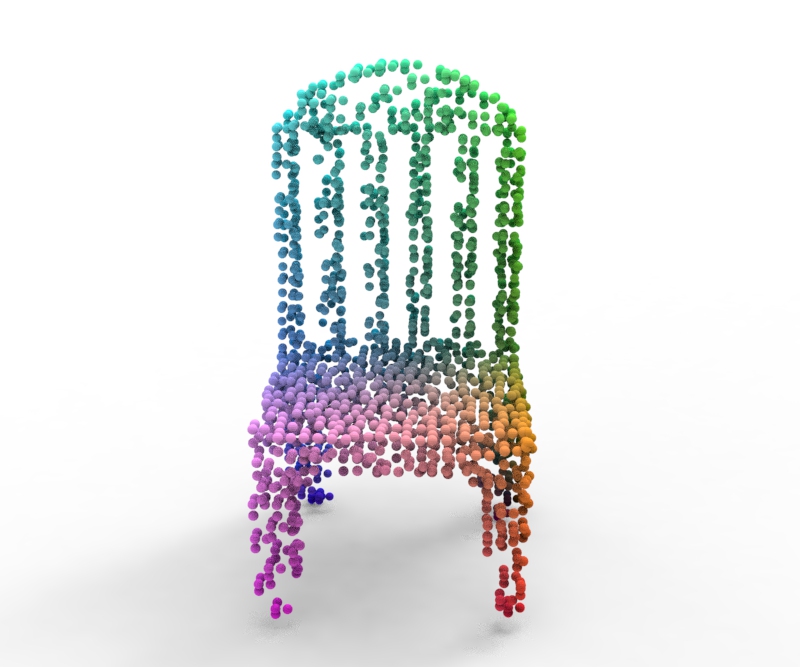}}
	\\
	\setcounter{subfigure}{0}
	\subfigure[]{
		\includegraphics[width=0.09\linewidth]{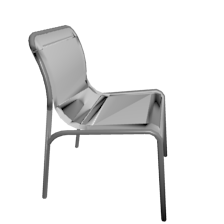}}
	\hspace{-2mm}
	\subfigure[]{
		\includegraphics[width=0.12\linewidth]{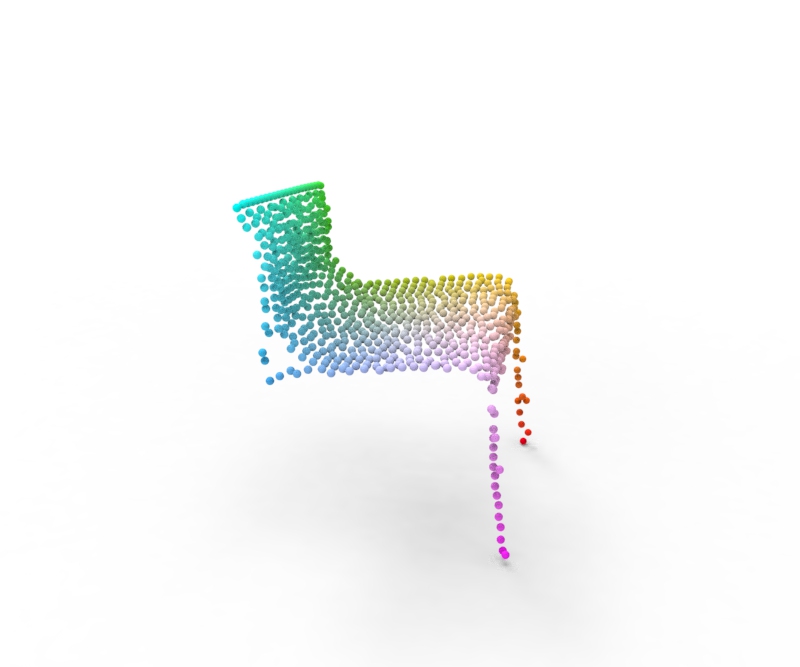}}
	\hspace{-2mm}
	\subfigure[]{
		\includegraphics[width=0.12\linewidth]{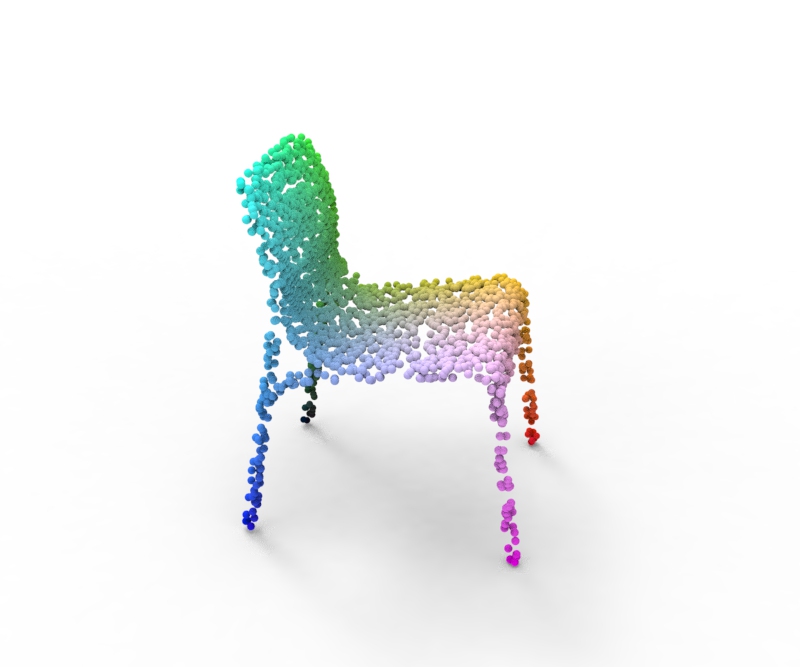}}
	\hspace{-2mm}
	\subfigure[]{
		\includegraphics[width=0.12\linewidth]{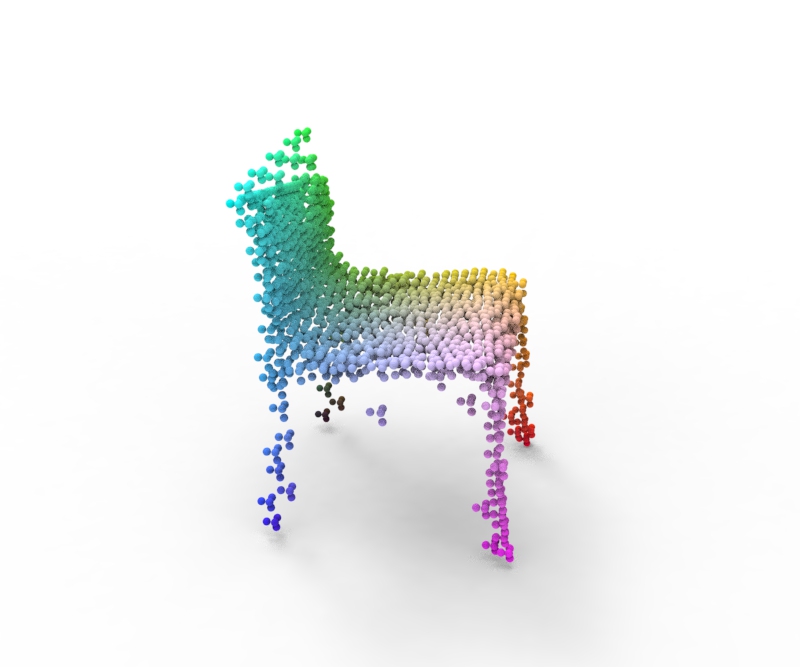}}
	\hspace{-2mm}
	\subfigure[]{
		\includegraphics[width=0.12\linewidth]{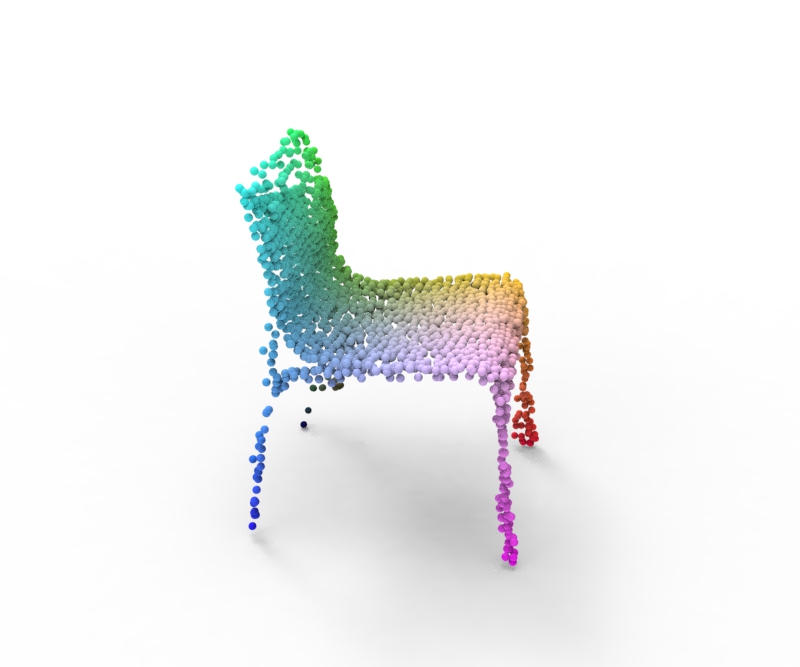}}
	\hspace{-2mm}
	\subfigure[]{
		\includegraphics[width=0.12\linewidth]{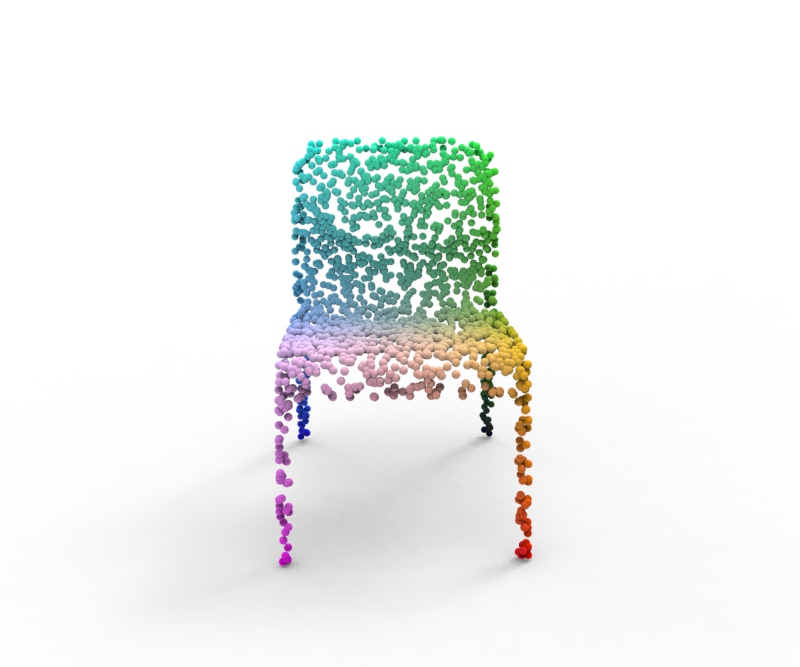}}
	\hspace{-2mm}
	\subfigure[]{
		\includegraphics[width=0.12\linewidth]{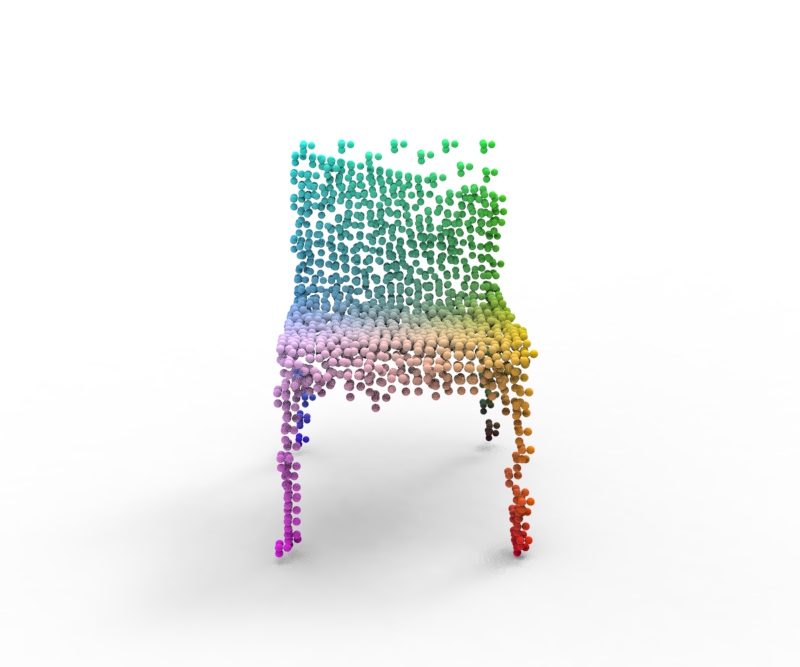}}
	\hspace{-2mm}
	\subfigure[]{
		\includegraphics[width=0.12\linewidth]{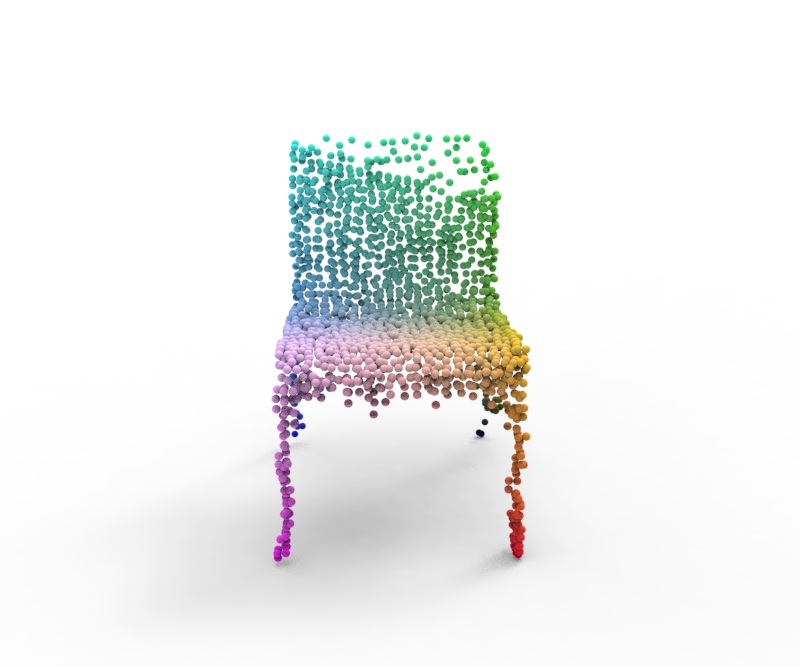}}

	\caption{Visual results of our VSR stage. (a) Input view images. (b) Input partial points. (c) Ground truth from the first view angle. (d) Predicted results without refinement from the first view angle. (e) Predicted results with refinement from the first view angle. (f) Ground truth from the second view angle. (g) Predicted results without refinement from the second view angle. (h) Predicted results with refinement from the second view angle.}
	\label{visual_refinement}
\end{figure*}

In addition, we conduct experiments to explore the effect of view numbers for calibration. As shown in Table \ref{view_num}, increasing the number of calibration views from 1 to 4 and 8 results in a reduction of 0.5/0.6 and 0.6/0.7 in CD values, respectively. The visual results in Figure \ref{cal_num_vis} show that utilizing a single view for calibration effectively eliminates some outliers. Increasing the number of views to four enhances the removal of a greater number of outliers. However, the effectiveness of outlier removal significantly diminishes when using eight views for calibration because there are only a few remaining outliers after calibration with four views.
Considering the trade-off between performance and computational complexity, we use a single view for calibration.

To better evaluate the effect of our VSR stage, as shown in Figure \ref{visual_refinement}, we visualize some results with or without refinement from two different angles. From Figures \ref{visual_refinement} (\textcolor{red}{d}) and (\textcolor{red}{g}), we can see that the CSR stage outputs many outliers around the object. Figures \ref{visual_refinement} (\textcolor{red}{e}) and (\textcolor{red}{h}) illustrate that the VSR stage effectively moves most of the outliers to accurate positions.

Our Cross-PCC adopts SnowflakeNet's 3D encoder, which consists of two point Transformer layers. To assess the influence of the point Transformer layers on our framework, we conduct comparative experiments by employing an encoder without point Transformer layers but with only three set abstraction layers. As Table \ref{transformer_encoder} shows, without using point Transformer layers, the performance of our framework decreases slightly. Such an observation also demonstrates that the advantage of our method does not stem from using more advanced feature representation modules but rather from its overall elegant design. 

\begin{table}[htbp]
	\centering
    \renewcommand\arraystretch{1.25}
	\caption{ Comparison of the 3D encoder with or without the Point Transformer Layers. Note that the reported values represent the average CD across all categories on 3D-EPN dataset. The \textbf{bold} numbers represent the best results across all methods.}
	\label{transformer_encoder}
	\begin{tabular}{l|c|c}
		\toprule[1.2pt]
		Methods                    & ${\rm CD}_{min}$ & ${\rm CD}_{avg}$ \\ \hline  
		w/o Transformer            & 8.6  & 9.4   \\ 
		w/ Transformer             & \textbf{8.4} & \textbf{9.1} \\ \bottomrule[1.2pt]
	\end{tabular}
\end{table}

\begin{table*}[htbp]
	\centering
    \renewcommand\arraystretch{1.25}
	\caption{Comparison of the effects between 2D and 3D supervision. The results are from 3D-EPN benchmark in terms of L2 CD\(\downarrow\) (scaled by \(10^{4}\)).  Note that the reported values to the left and right of the slash ``/" are the ${\rm CD}_{min}$ and ${\rm CD}_{avg}$, respectively. The \textbf{bold} numbers represent the best results.}
	\label{supervision_comparison}
	\begin{tabular}{l|c|c|c|c|c|c|c|c|c}
		\toprule[1.2pt]
		Methods              & Average & Plane & Cabinet & Car  & Chair & Lamp	& Couch	& Table	& Watercraft \\ \hline  
	  PCN (2D supervision)  & 14.6   & 3.9   & 17.5    & 9.2  & 15.8  & 19.0  & 14.5  & 26.1  & 10.9\\ 
        PCN (3D supervision)  & 7.6    & 2.0   & 8.0     & 5.0  & 9.0   & 13.0	& 8.0	& 10.0	& 6.0\\ \hline
		Cross-PCC (2D supervision)& 8.4/9.1& 2.3/2.4& 13.0/14.0& 7.4/7.6 & 9.8/10.6 & 7.8/9.3 & 11.5/12.1 & 9.6/10.7	& 6.1/6.8 \\
		Cross-PCC (3D supervision)&\textbf{6.0/6.7}&\textbf{1.8/1.9}&\textbf{8.8/9.5} &\textbf{5.4/5.6}&\textbf{7.4/8.2} & \textbf{6.1/7.4} &\textbf{7.6/8.3}&\textbf{7.2/8.1} & \textbf{4.0/4.6} \\ \bottomrule[1.2pt]
	\end{tabular}
\end{table*}

\begin{table*}[htbp]
	\centering
	\renewcommand\arraystretch{1.25}
	\caption{Standard deviations of CD values of eight view images. This metric measures the impact of view angles of input images on completion results. A lower value means lower impact. ``CSR" and ``VSR" denote the Coarse Shape Reconstruction stage and View-assisted Shape Refinement stage, respectively.}
	\label{angle_var}
	\begin{tabular}{c|c|c|c|c|c|c|c|c|c}
		\toprule[1.2pt]
		Category             & Average & Plane & Cabinet & Car & Chair & Lamp & Couch & Table	& Watercraft \\ \hline 
		CSR            & 0.390  & 0.021 & 0.792 & 0.039 & 0.262 & 0.967 & 0.177 & 0.377 & 0.487 \\
		VSR             &  0.525   & 0.106 & 0.670 & 0.154 & 0.481 & 1.178 & 0.373 & 0.698 & 0.537
		 \\ 
   \bottomrule[1.2pt]
	\end{tabular}
\end{table*}

\begin{figure*}[htbp]
	\centering  
	\subfigbottomskip=1pt 
	\subfigure{
		\includegraphics[width=0.1\linewidth]{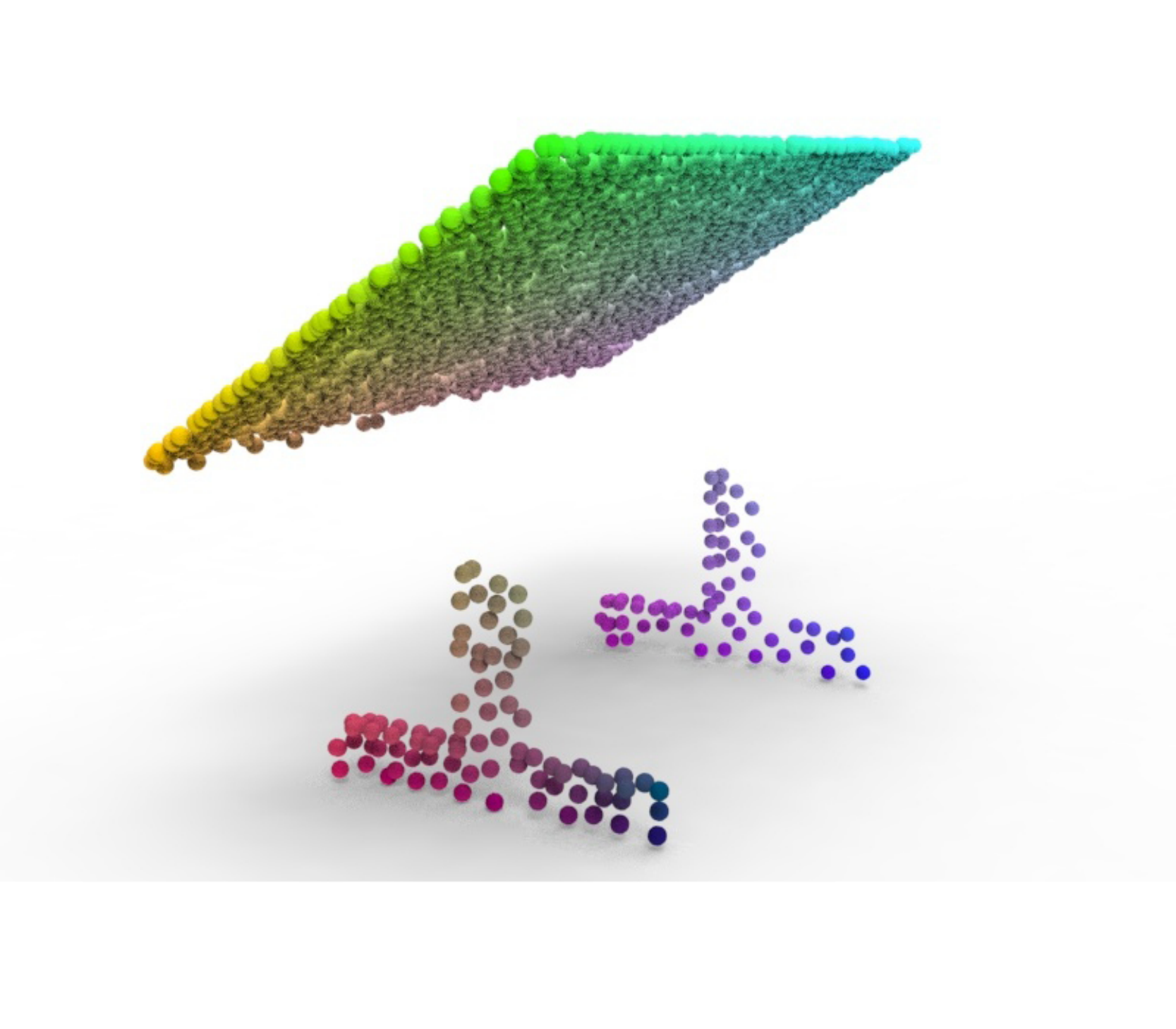}}
	\hspace{-5mm}
	\subfigure{
		\includegraphics[width=0.1\linewidth]{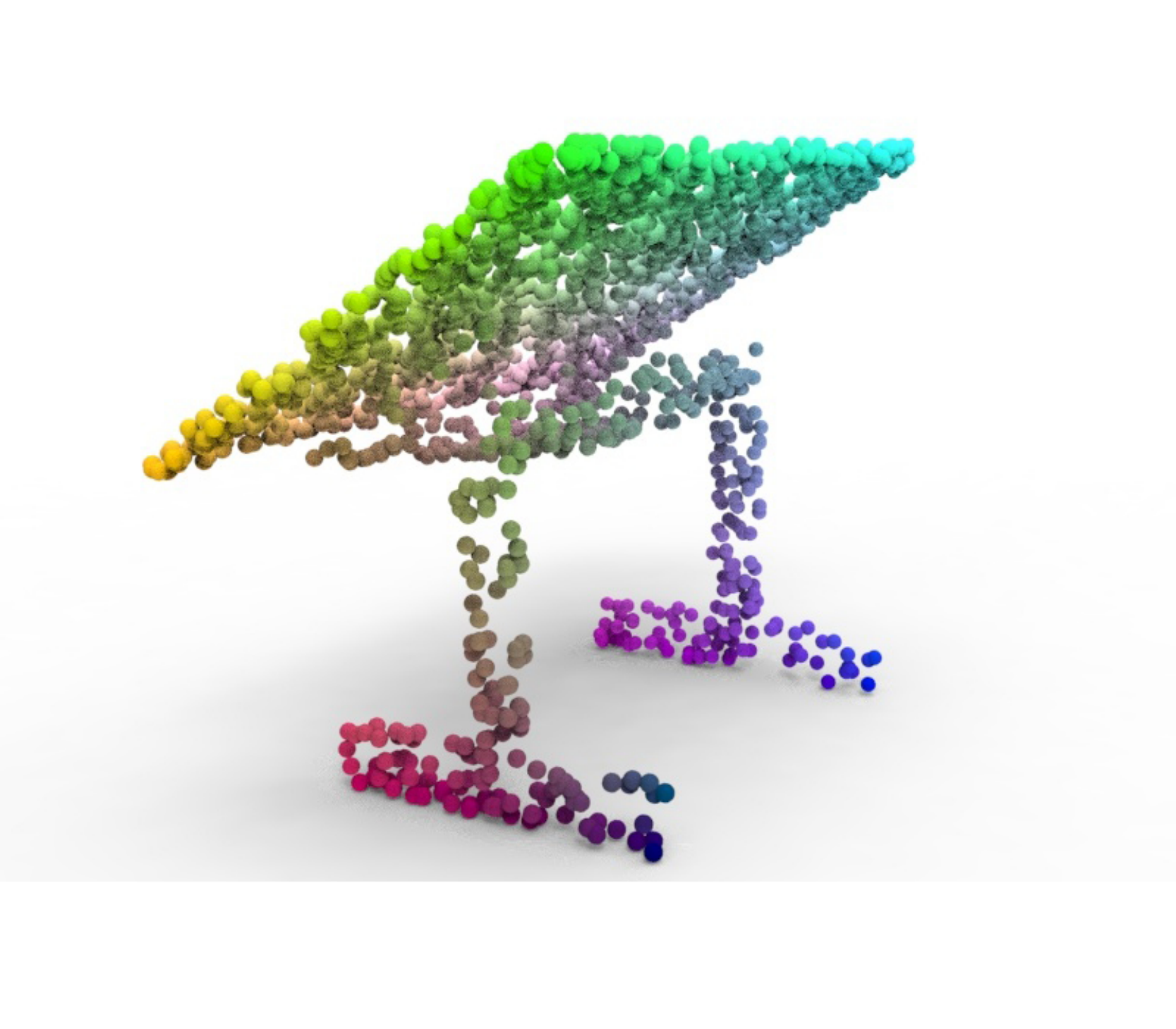}}
	\hspace{-5mm}
	\subfigure{
		\includegraphics[width=0.1\linewidth]{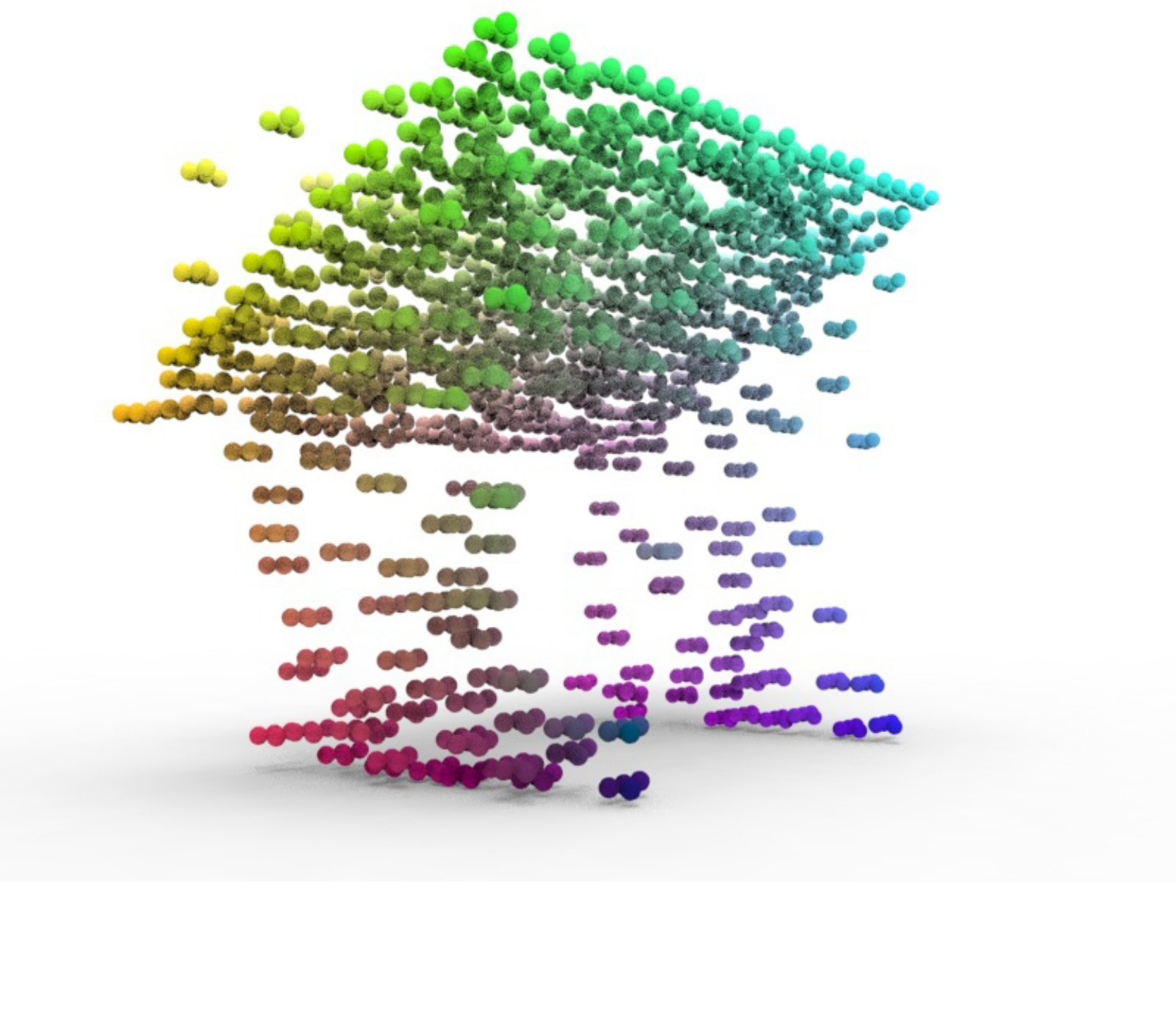}}
	\hspace{-5mm}
	\subfigure{
		\includegraphics[width=0.08\linewidth]{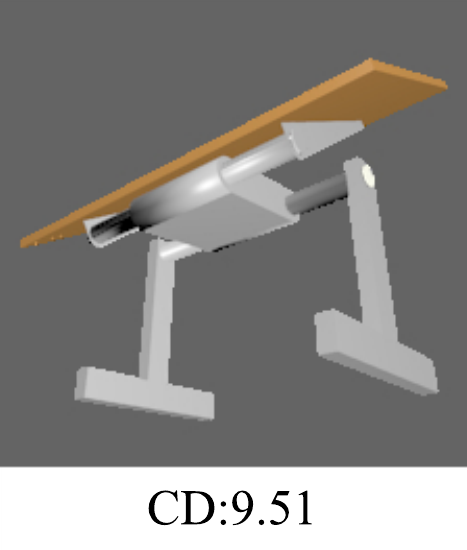}}
	\hspace{-2mm}
	\subfigure{
		\includegraphics[width=0.08\linewidth]{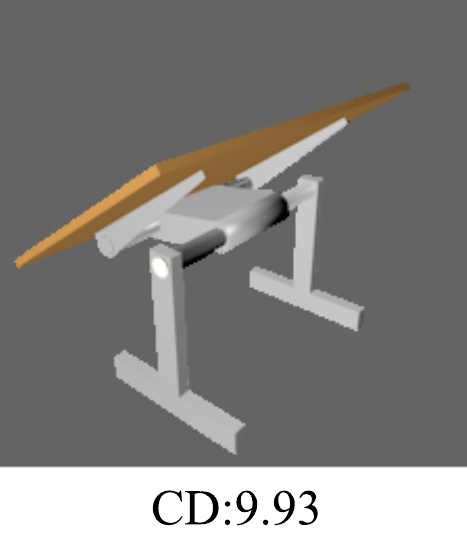}}
	\hspace{-2mm}
	\subfigure{
		\includegraphics[width=0.08\linewidth]{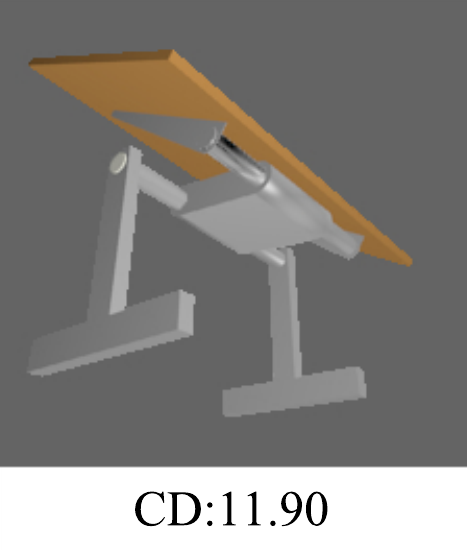}}
	\hspace{-2mm}
	\subfigure{
		\includegraphics[width=0.08\linewidth]{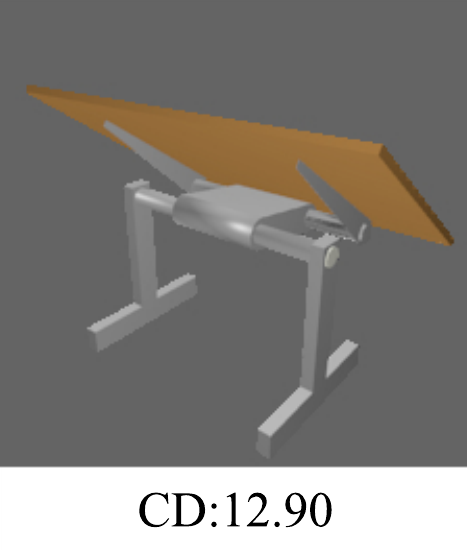}}
	\hspace{-2mm}
	\subfigure{
		\includegraphics[width=0.08\linewidth]{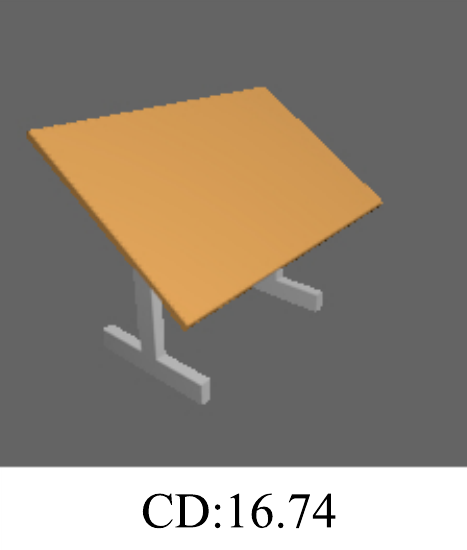}}
	\hspace{-2mm}
	\subfigure{
		\includegraphics[width=0.08\linewidth]{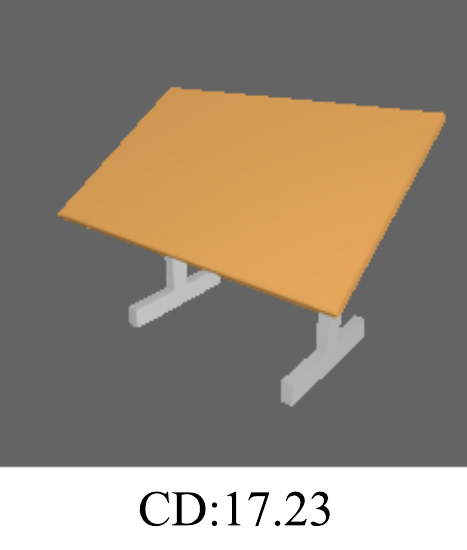}}
	\hspace{-2mm}
	\subfigure{
		\includegraphics[width=0.08\linewidth]{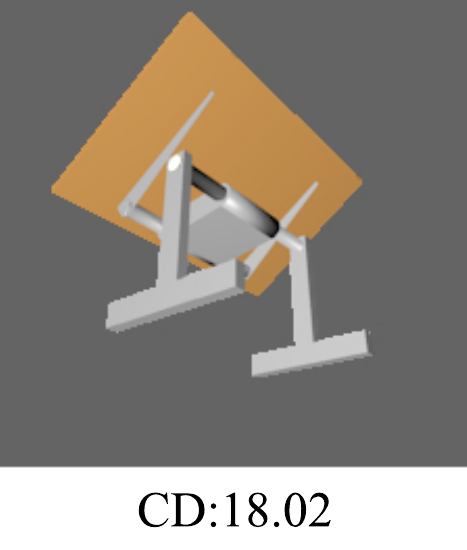}}
	\hspace{-2mm}
	\subfigure{
		\includegraphics[width=0.08\linewidth]{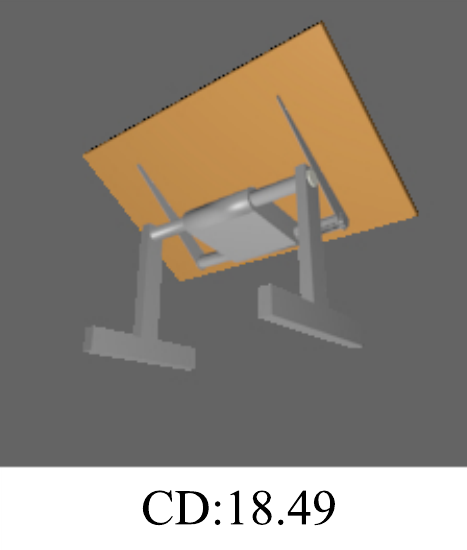}}
	\\
	\subfigure{
		\includegraphics[width=0.1\linewidth]{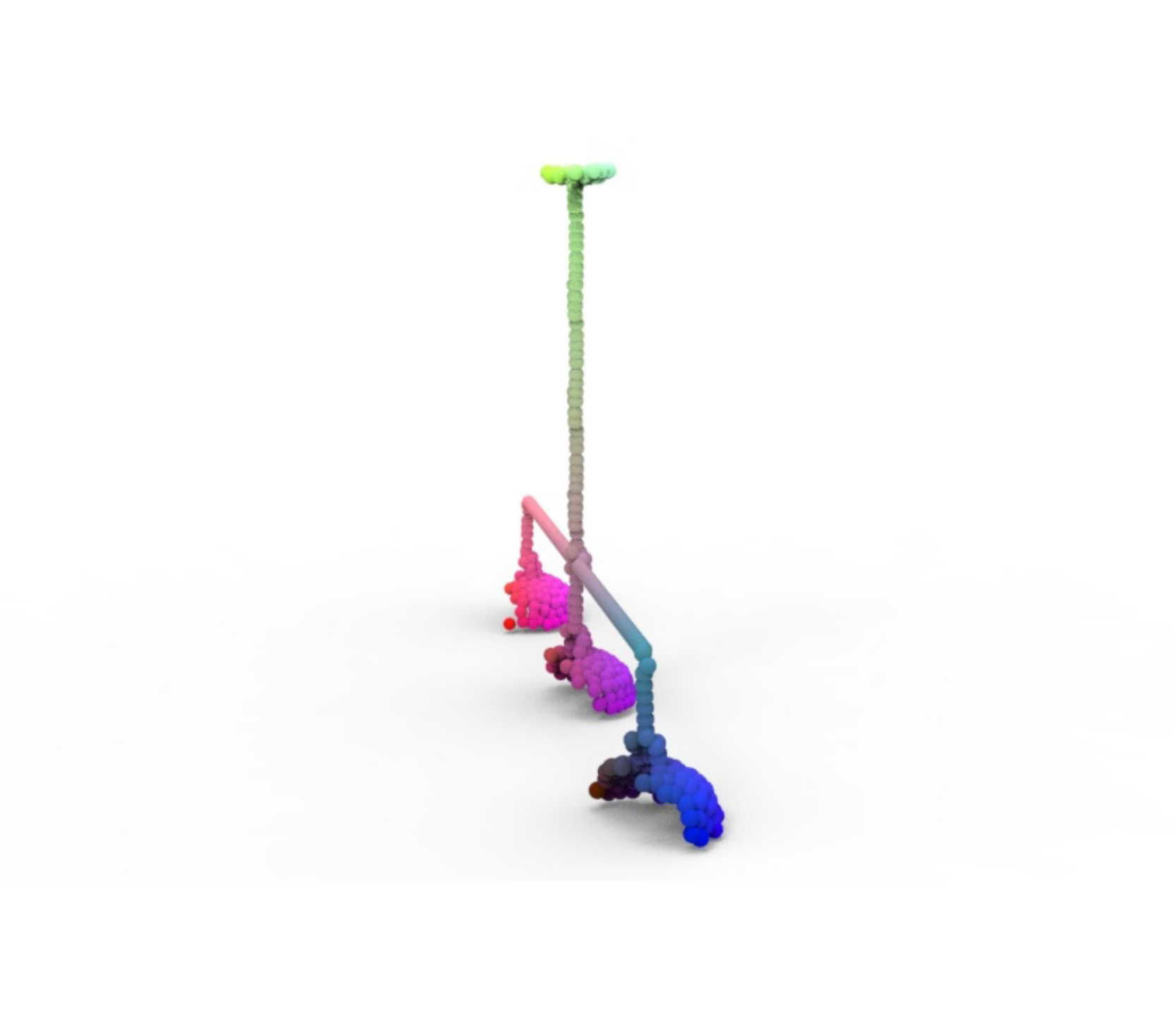}}
	\hspace{-5mm}
	\subfigure{
		\includegraphics[width=0.1\linewidth]{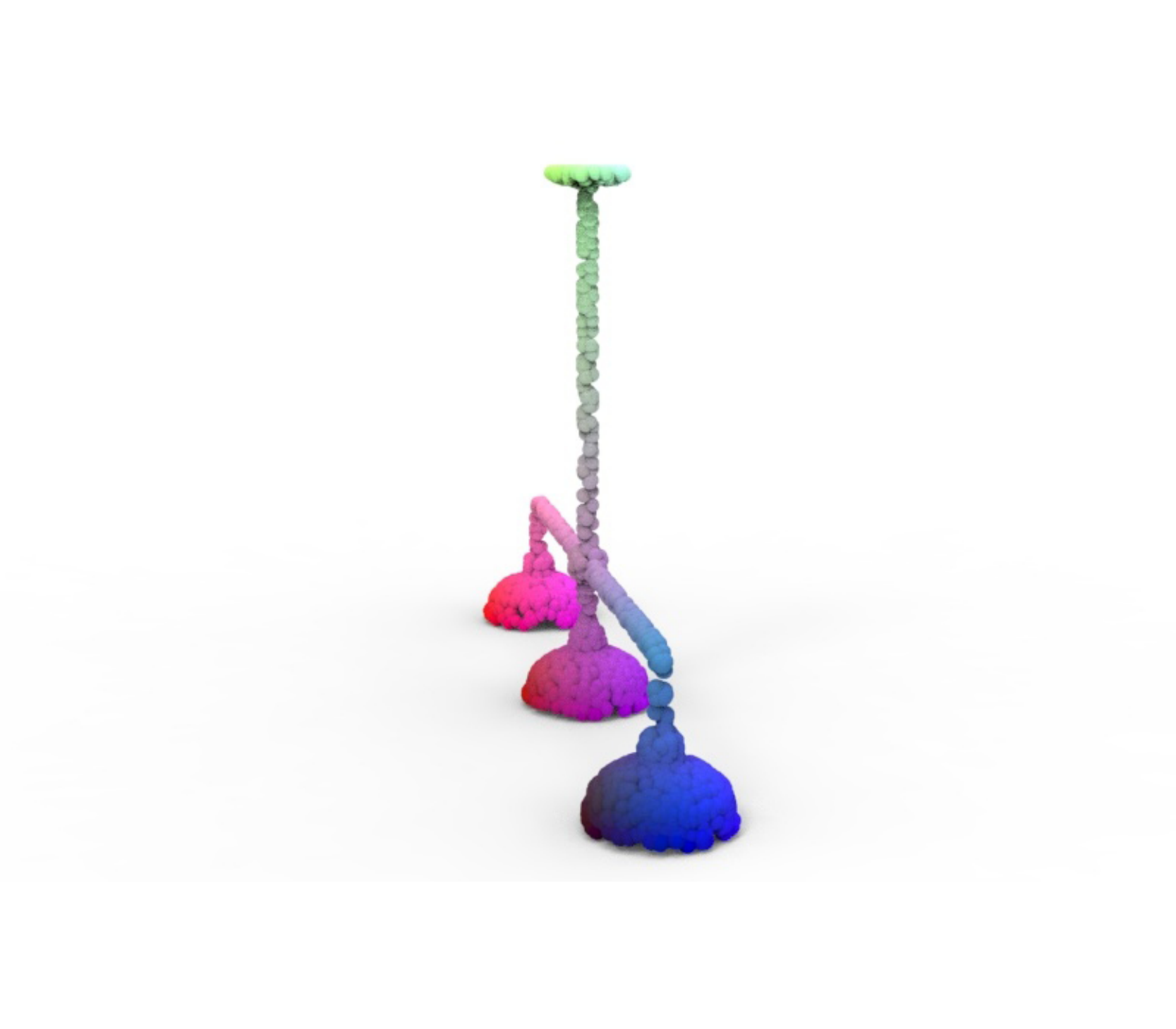}}
	\hspace{-5mm}
	\subfigure{
		\includegraphics[width=0.1\linewidth]{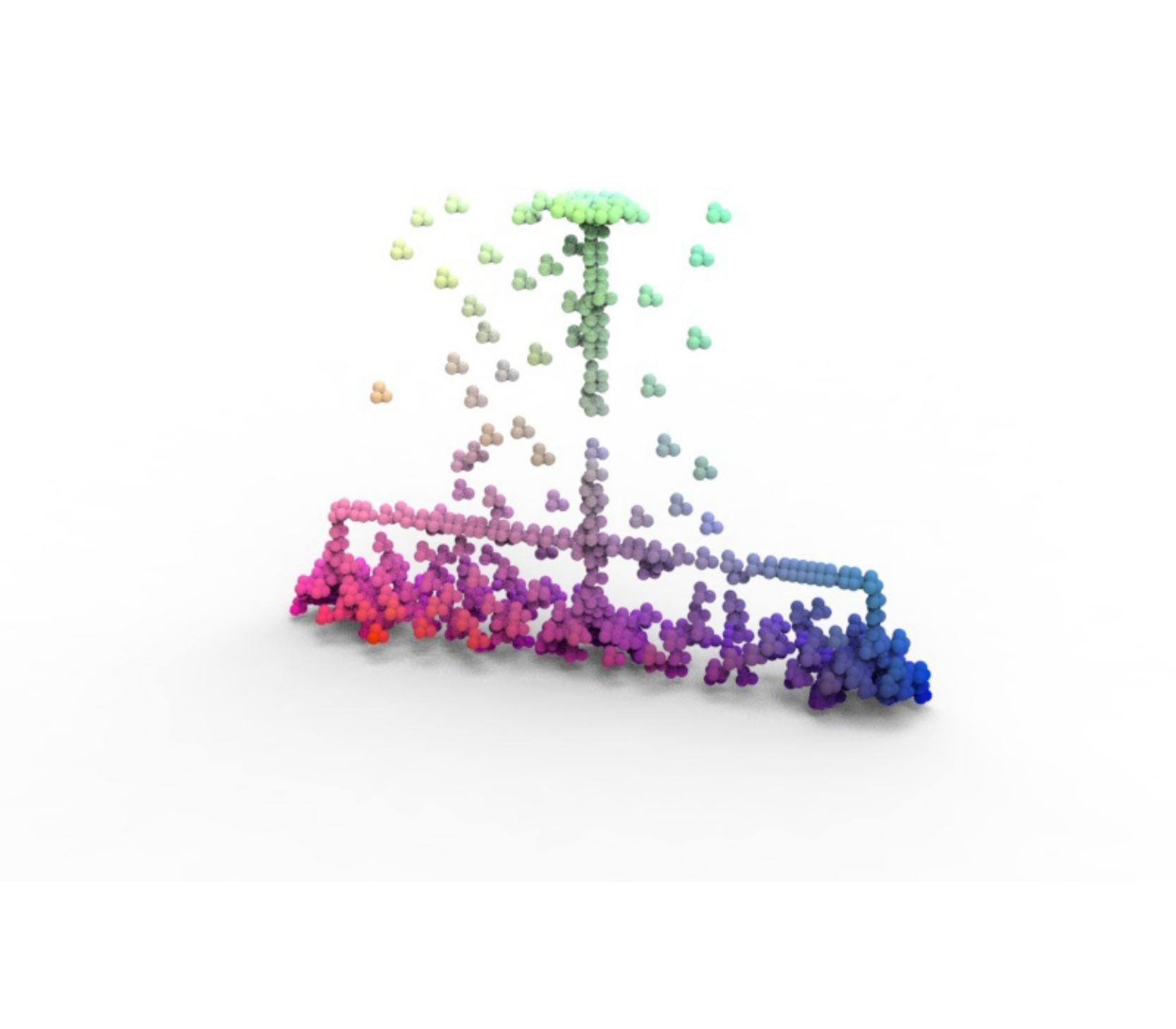}}
	\hspace{-5mm}
	\subfigure{
		\includegraphics[width=0.08\linewidth]{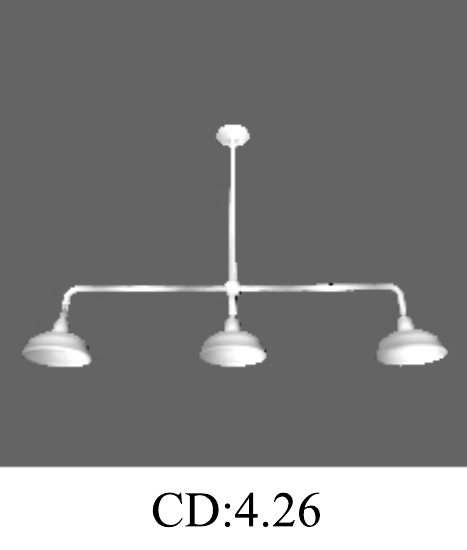}}
	\hspace{-2mm}
	\subfigure{
		\includegraphics[width=0.08\linewidth]{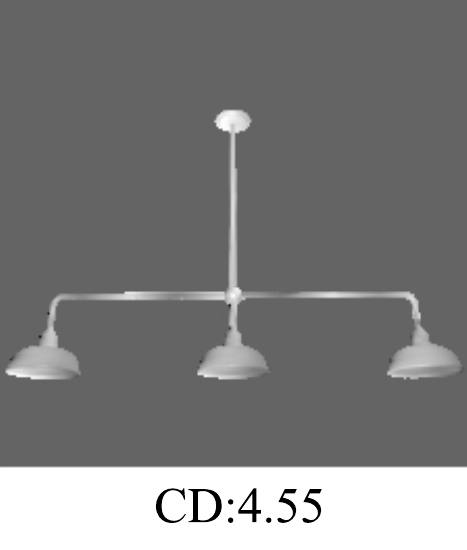}}
	\hspace{-2mm}
	\subfigure{
		\includegraphics[width=0.08\linewidth]{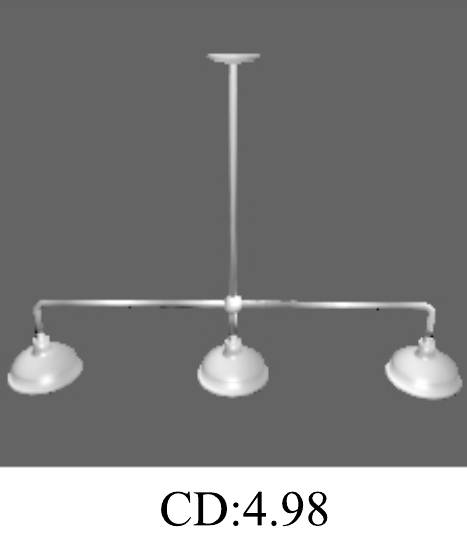}}
	\hspace{-2mm}
	\subfigure{
		\includegraphics[width=0.08\linewidth]{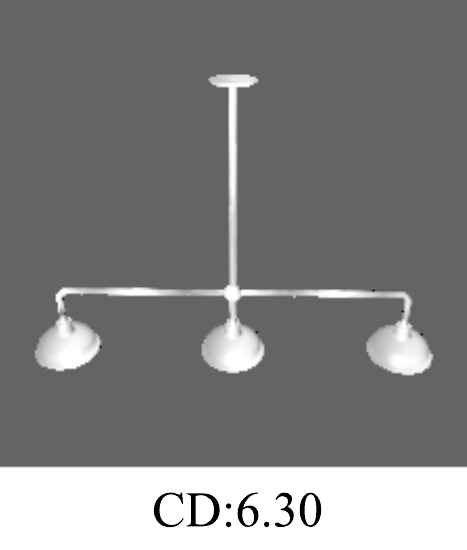}}
	\hspace{-2mm}
	\subfigure{
		\includegraphics[width=0.08\linewidth]{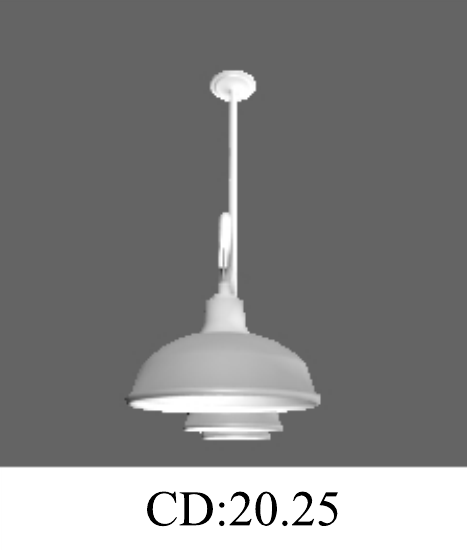}}
	\hspace{-2mm}
	\subfigure{
		\includegraphics[width=0.08\linewidth]{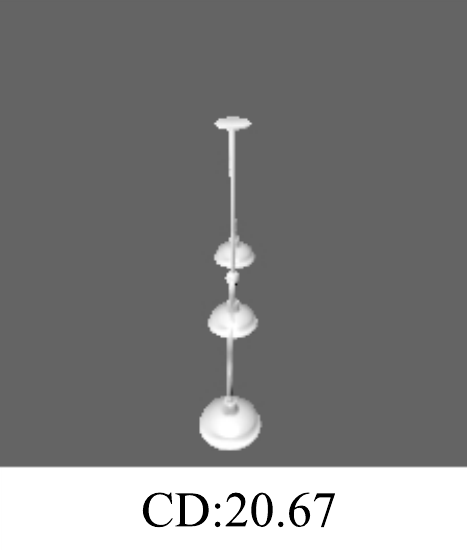}}
	\hspace{-2mm}
	\subfigure{
		\includegraphics[width=0.08\linewidth]{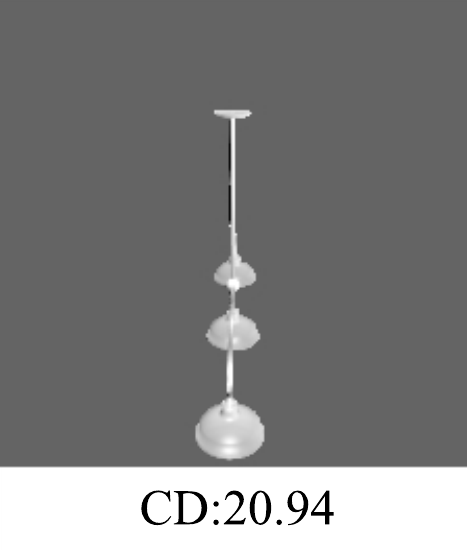}}
	\hspace{-2mm}
	\subfigure{
		\includegraphics[width=0.08\linewidth]{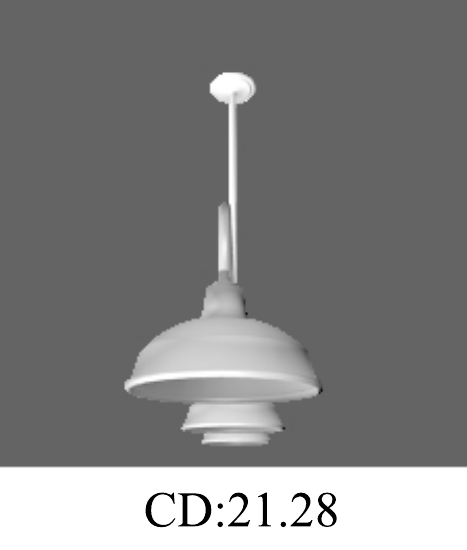}}
	\\
	\setcounter{subfigure}{0}
	\subfigure[]{
		\includegraphics[width=0.1\linewidth]{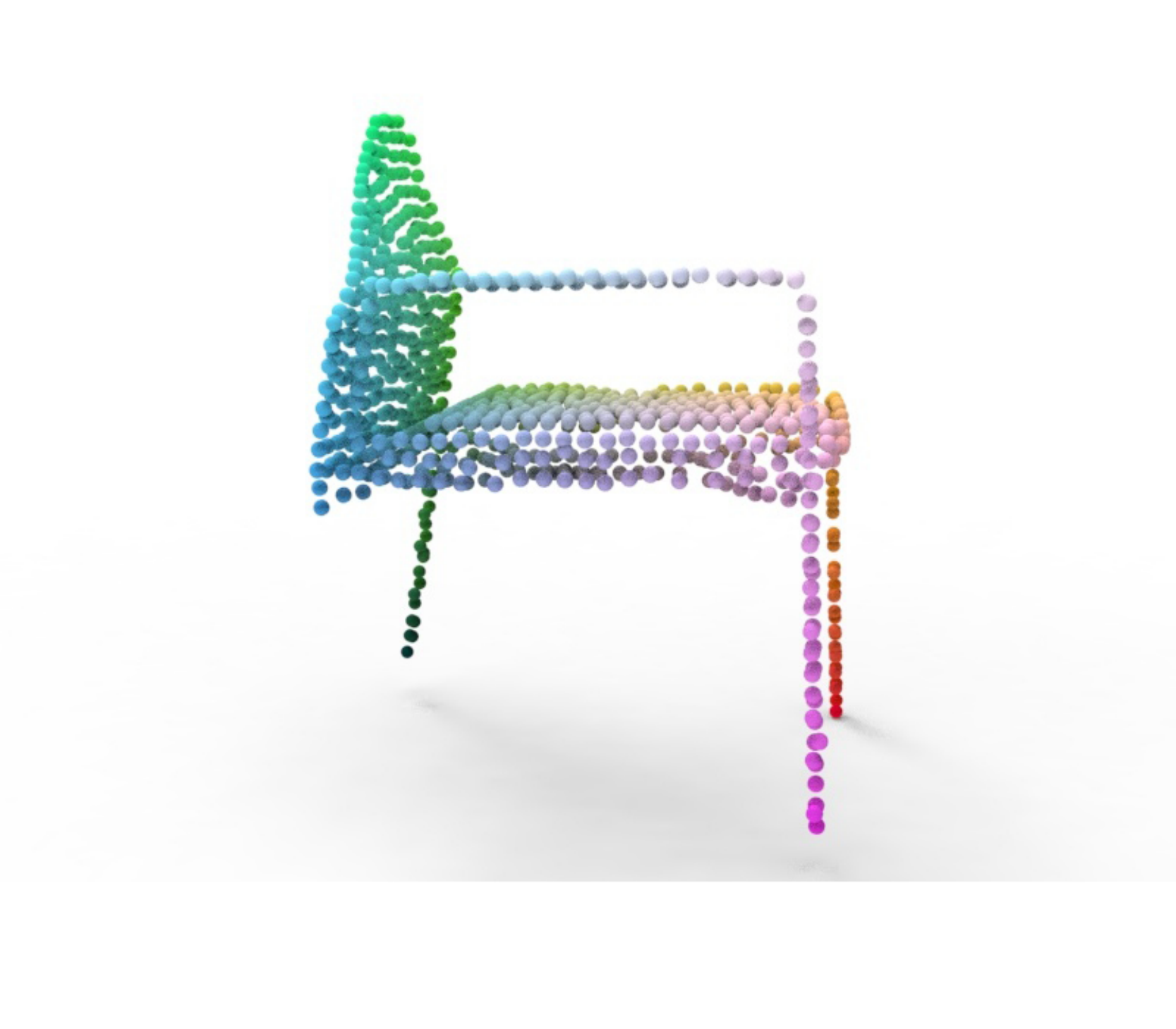}}
	\hspace{-5mm}
	\subfigure[]{
		\includegraphics[width=0.1\linewidth]{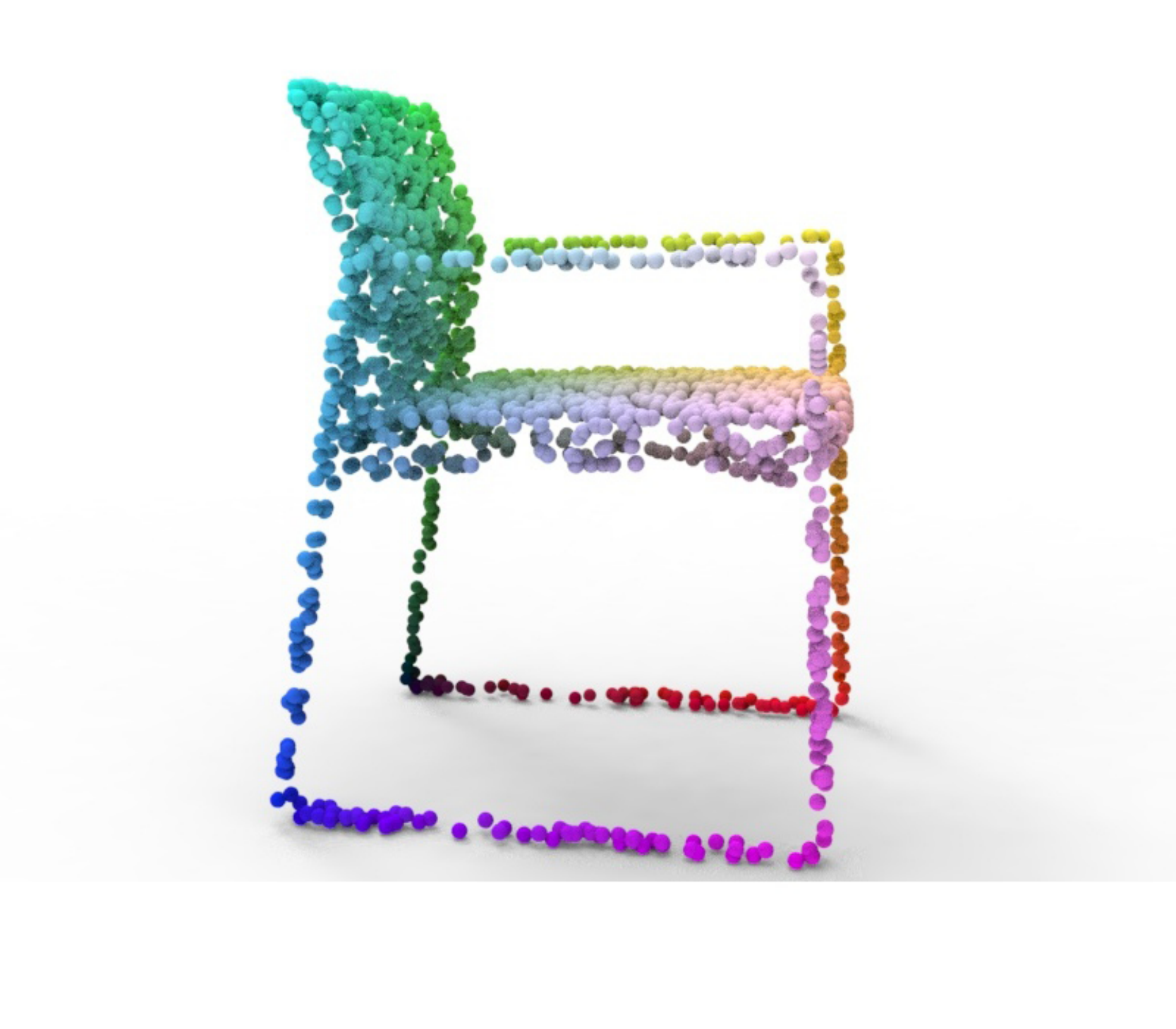}}
	\hspace{-5mm}
	\subfigure[]{
		\includegraphics[width=0.1\linewidth]{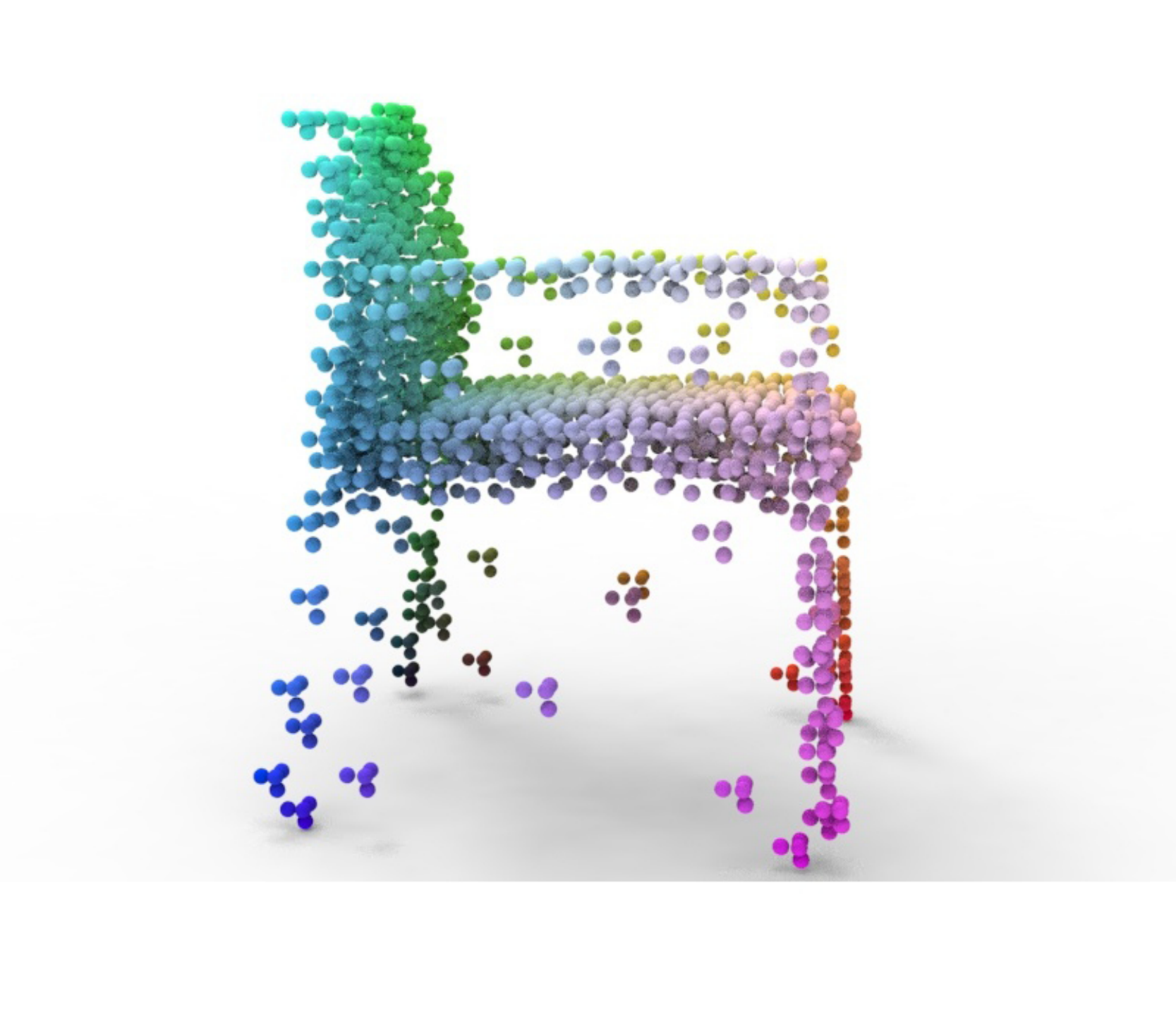}}
	\hspace{-5mm}
	\subfigure[]{
		\includegraphics[width=0.08\linewidth]{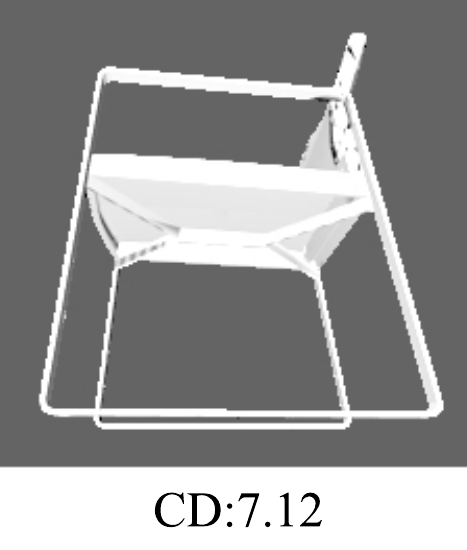}}
	\hspace{-2mm}
	\subfigure[]{
		\includegraphics[width=0.08\linewidth]{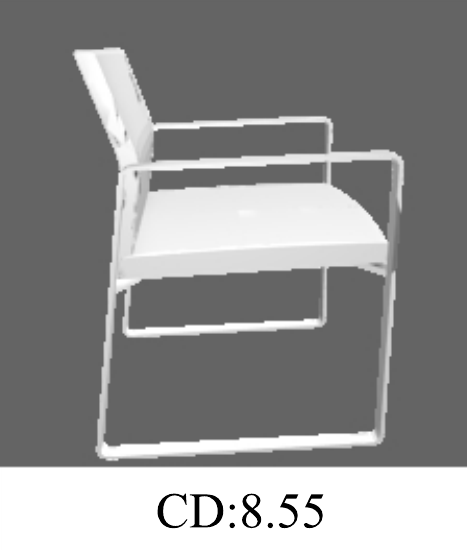}}
	\hspace{-2mm}
	\subfigure[]{
		\includegraphics[width=0.08\linewidth]{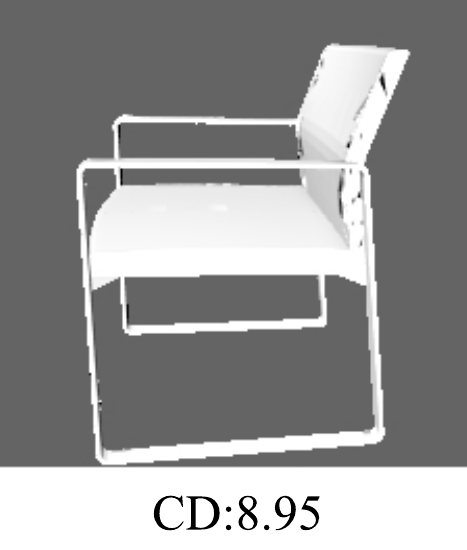}}
	\hspace{-2mm}
	\subfigure[]{
		\includegraphics[width=0.08\linewidth]{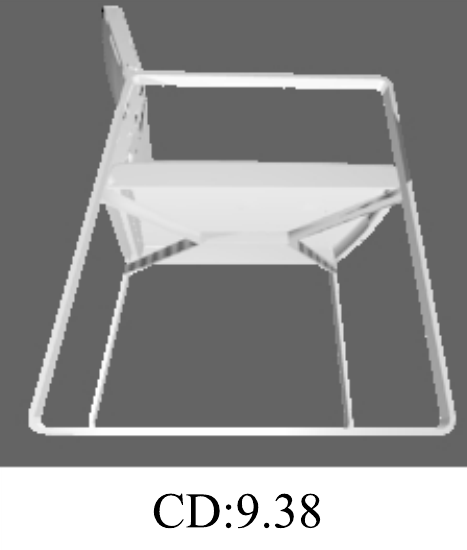}}
	\hspace{-2mm}
	\subfigure[]{
		\includegraphics[width=0.08\linewidth]{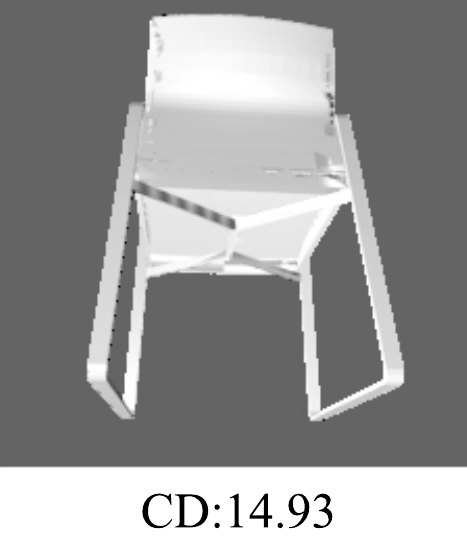}}
	\hspace{-2mm}
	\subfigure[]{
		\includegraphics[width=0.08\linewidth]{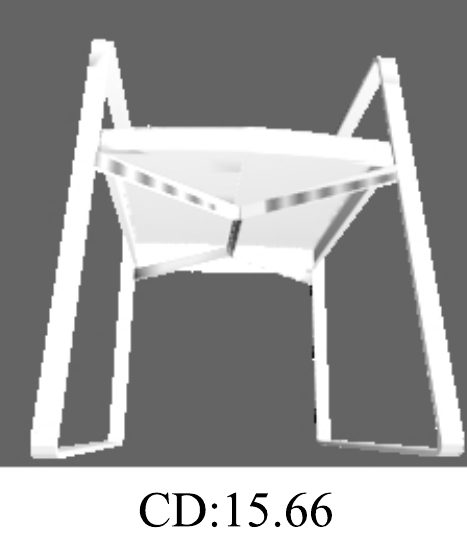}}
	\hspace{-2mm}
	\subfigure[]{
		\includegraphics[width=0.08\linewidth]{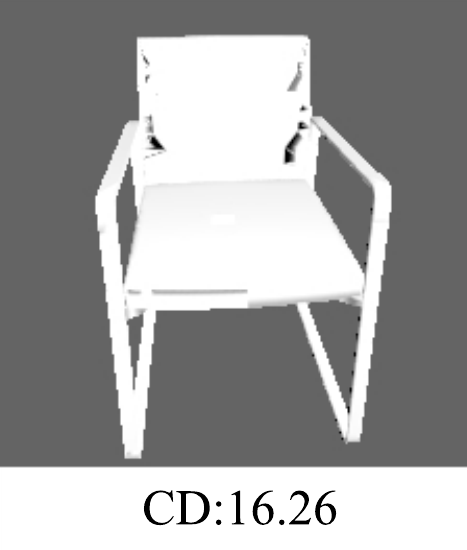}}
	\hspace{-2mm}
	\subfigure[]{
		\includegraphics[width=0.08\linewidth]{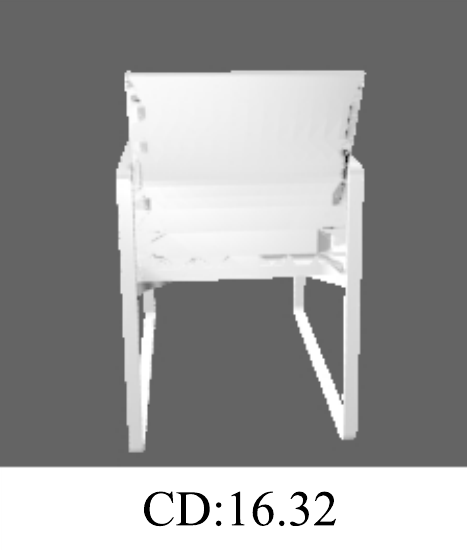}}

	\caption{Ranking of CD values of the results predicted by different view images. (a) Input partial points (b) Ground truth. (c) Predicted results without refinement. (d)$\sim$(k) CD rank of predicted results with refinement when inputting different angles of view images.}
	\label{view_angle_visual}
\end{figure*}

\subsubsection{Effect of 2D supervision}
To explore the effect of 2D supervision, we substitute our 2D supervision for the original 3D ground truth in PCN. Furthermore, we conduct experiments utilizing the complete 3D point cloud as supervision for our Cross-PCC. The experimental results are presented in Table \ref{supervision_comparison}. Notably, when employing 2D supervision, PCN exhibits a significant performance drop of nearly 50\%, falling far behind our proposed method. In contrast, the performance gap between our method using 2D and 3D supervision is merely 2.4/2.4. These findings highlight two important observations: Firstly, 3D supervision yields better results than 2D supervision because 2D images cannot fully cover the complete 3D shape. Nevertheless, acquiring complete and clean 3D point clouds proves more challenging compared to 2D images. Secondly, our method not only achieves remarkable performance under 2D supervision but also demonstrates a minimal performance gap compared to using 3D supervision.

Additionally, we investigate the impact of using different numbers of 2D images as supervision. As Table \ref{view_num} shows, 
the difference in CD values between our model optimized with supervision from one single-view silhouette and eight-view silhouettes is less than 1. This suggests that even when supervised with only a single-view silhouette, our model can achieve competitive outcomes. Increasing the number of supervised images further improves the performance of the model.

\subsubsection{Effect of view angles of the input images}
Our Cross-PCC consumes a single-view image as input to aid the completion. Intuitively, the images captured from different angles may affect the completion results differently. To verify the effect of view angles, we first compute the standard deviations of CD values of the results predicted by eight view images for all objects. The formula of this metric is listed below.
\begin{equation}
	{\rm STD}=\frac{1}{J}\sum_{j=1}^J \sqrt{\frac{1}{I}\sum_{i=1}^I({\rm CD}_{j,i}-\overline{{\rm CD}_j})^2},
\end{equation}
where $J$ denotes the number of objects, and $\overline{{\rm CD}_j}$ means the average CD of the results from $I$ view images of the $j$-th object.
Table \ref{angle_var} lists the standard deviation of each category. From Table \ref{angle_var}, we can find that the view angles have a larger impact on the results of the VSR stage than the CSR stage, which demonstrates that the view calibrator more requires a suitable view angle. The standard deviations of most categories are less than 0.6, which means the results generated by images from different view angles are similar.
However, the standard deviation of the ``lamp" is much larger than those of others. To further study the effect of view angles of images on our results, we visualize some typical outputs with large CD standard deviations in Figure \ref{view_angle_visual}. Figure \ref{view_angle_visual} (\textcolor{red}{c}) shows that the predicted outliers often populate around the thin bars and holes, such as the legs of the table and the line of the lamp. From Figures \ref{view_angle_visual} (\textcolor{red}{d})$\sim$(\textcolor{red}{k}), we can see that the view images whose results have smaller CD values always capture the parts where outliers populate. Besides, the two sub-figures in the first line of Figures \ref{view_angle_visual} (\textcolor{red}{j}) and (\textcolor{red}{k}) illustrate that the view images cannot yield good results if the outliers occlude the body of the object. This is because the outliers in the 3D space will lie in the foregrounds when projected into the image plane in this case. That means they are not outliers and thus will not be calibrated.
Generally, for those objects with many thin bars, holes or gaps, we should capture these parts in view images to obtain better results.

\section{Conclusion}
\label{conclusion}
In this paper, we have proposed Cross-PCC, the first model to complete partial point clouds without any complete 3D ground truth. In contrast to previous methods that require a large complete and clean 3D dataset, we only leveraged 2D view images to guide the completion process.  Cross-PCC consumes a partial point cloud and a single-view image as inputs and uses different views of silhouette images to constrain the predicted shape.
Considering that the single modality of partial point clouds cannot always determine the unknown parts, we extracted image features and fuse them with 3D features to aid the prediction of the complete shapes. To reduce the outliers in the predicted results, we proposed a calibrator to move the outliers to correct positions and then employ an offset predictor to refine them. We conducted extensive experiments on both synthetic data and real-world data. The experimental results show that our method outperforms state-of-the-art unsupervised methods by a large margin and only falls behind some supervised methods by a tiny gap.

\ifCLASSOPTIONcaptionsoff
  \newpage
\fi


\end{document}